\documentclass[preprint,12pt]{elsarticle}

\usepackage{graphicx}
\usepackage{amssymb}

\usepackage{lineno,hyperref}
\usepackage{amsmath,amssymb,url,color}

\usepackage{color}
\usepackage{float}
\usepackage{tabularx}

\usepackage{jumoline}
\usepackage{ulem}
\usepackage{proofread} 
\noproofreadmark

\journal{Neural Networks}
  
\begin{document}

\begin{frontmatter}


\title{A Whole Brain Probabilistic Generative Model:\\ Toward Realizing Cognitive Architectures for Developmental Robots}


\author[1]{Tadahiro Taniguchi}
\author[2,7]{Hiroshi Yamakawa}
\author[3]{Takayuki Nagai}
\author[4]{Kenji Doya}
\author[5]{Masamichi Sakagami}
\author[2]{Masahiro Suzuki}
\author[6]{Tomoaki Nakamura}
\author[1]{Akira Taniguchi}

\address[1]{Ritsumeikan University, 1-1-1 Noji-higashi, Kusatsu, Japan}
\address[2]{The University of Tokyo, 7-3-1, Hongo, Bunkyo-ku, Tokyo, Japan}
\address[3]{Osaka University, 1-3 Machikane-yama, Toyonaka, Osaka, Japan}
\address[4]{Okinawa Institute of Science and Technology Graduate University, 1919-1 Tancha, Onna-son, Kunigami, Okinawa, Japan}
\address[5]{Tamagawa University, 6-1-1 Tamagawa Gakuen, Machida, Tokyo, Japan}
\address[6]{The University of Electro-Communications, 1-5-1 Chofugaoka, Chofu, Tokyo, Japan}
\address[7]{The Whole Brain Architecture Initiative, Nishikoiwa 2-19-21, Edogawa-ku, Tokyo, Japan}



\begin{abstract}
Building a humanlike integrative artificial cognitive system, that is, an artificial general intelligence (AGI), is the holy grail of the artificial intelligence (AI) field. Furthermore, a computational model that enables an artificial system to achieve cognitive development will be an excellent reference for brain and cognitive science. 
This paper describes an approach to develop a cognitive architecture by integrating elemental cognitive modules \addspan{to enable the training of the modules as a whole}. This approach is based on two ideas: (1) brain-inspired AI, learning human brain architecture to build human-level intelligence, and (2) \addspan{a probabilistic generative model}(PGM)-based cognitive system to develop a cognitive system for developmental robots by integrating PGMs. The development framework is called a whole brain PGM (WB-PGM)\addspan{, which differs fundamentally from existing cognitive architectures in that it can learn continuously through a system based on sensory-motor information.
}
In this study, we describe the rationale of WB-PGM, the current status of PGM-based elemental cognitive modules, their relationship with the human brain, the approach to the integration of the cognitive modules, and future challenges.
\addspan{Our findings can serve as a reference for brain studies. As PGMs describe explicit informational relationships between variables, this description provides interpretable guidance from computational sciences to brain science. By providing such information, researchers in neuroscience can provide feedback to researchers in AI and robotics on what the current models lack with reference to the brain. Further, it can facilitate collaboration among researchers in neuro-cognitive sciences as well as AI and robotics.}
\end{abstract}


\end{frontmatter}


\section{Introduction}\label{sec:1}
Infants acquire a wide range of cognitive capabilities through daily physical and social interactions with their environment. Through this developmental process, they acquire basic physical skills (e.g., reaching and grasping), perceptual skills (e.g., object recognition and phoneme recognition), and social skills (e.g., linguistic communication and intention estimation)~\citep{Taniguchi2018symbol}.
This open-ended learning process involving many types of modalities, tasks, and interactions is often referred to as {\it lifelong learning}~\citep{Najeeb1988,parisi2019continual}.
The central question in next-generation artificial intelligence (AI) and developmental robotics is how to build an {\it integrative cognitive system} capable of lifelong learning and human-like behavior in various environments such as homes, offices, and outdoors.  In this paper, inspired by the whole brain architecture (WBA) approach, using a whole brain probabilistic generative model (WB-PGM), we introduce the idea of building an integrative cognitive system that can alternatively be referred to as artificial general intelligence (AGI) \citep{Yamakawa2021-qa}.


\delspan{Against this backdrop, we explore the process of establishing a cognitive architecture for developmental robots.} 
{\it A cognitive architecture} is a hypothesis about the mechanisms of human intelligence underlying our behaviors~\citep{Rosenbloom2011}.
The study of cognitive architecture involves developing a presumably standard model of the {\it human mind}~\citep{Laird2017}.  
It integrates a wide range of cognitive capabilities: representation and memory, problem-solving and planning, learning, reflection, interaction, and the social aspects of cognition.
Notably, interaction includes perception, motor control, and the use of language. 
Social aspects of cognition include intention sharing, emotional expression, collaborative control, and language use for communication and collaboration.
In the literature on cognitive and developmental robotics, a robot is required to integrate a wide range of sensor information and perform a variety of cognitive tasks to explore environments, grasp and handle objects, and interact with people~\citep{Vernon2007,Vernon2016,Doncieux2020,tanevska2019cognitive}. In such contexts, a cognitive architecture is required to integrate various elemental cognitive modules. 

To build a cognitive architecture that addresses the functions of the entire brain, it is desirable to describe the computation of the entire brain with as few types of theoretical and computational elements (primitive structures, circuits, computational nodes, and so on) as possible (ideally, one type). The \addspan{probabilistic generative model (PGM)} is a strong candidate for a computational model for this purpose\delspan{, which is discussed later}.

\delspan{In this paper, the emphasis is on creating cognitive architectures using PGMs.} A PGM is a probabilistic description of how causes generate sensations, that is, observed data. That is, PGM is a statistical model of the joint probability distribution on observable data~\citep{bishop2006pattern}. 
PGMs learn to predict these observations. 
This is also often referred to as {\it predictive coding} and the {\it free energy principle} (FEP). FEP is a powerful idea for explaining the human brain.
It is a normative framework for Bayesian inference and learning of the human brain and is based on a PGM~\citep{hohwy2013predictive,friston2019free}.
 According to the FEP, perception and action can be modeled as self-evidencing (as described in Section~\ref{subsec:World Model}).
Many types of elemental cognitive modules have been developed based on PGMs; including hidden Markov models (HMMs), latent Dirichlet allocation (LDA), variational autoencoder (VAE), and partially observable Markov decision process (POMDP) ~\citep{rabiner1986introduction,Blei:2003:LDA:944919.944937,kingma2013auto,thrun2005probabilistic} as described in Section~\ref{sec:3}. Most PGMs can be represented using probabilistic graphical models. 
For a human-like developmental cognitive system, unsupervised (or self-supervised) learning is required. PGMs are known to be suitable for unsupervised learning.
Supervised learning using human-annotated training data, on which most current AI systems, such as image recognition and machine translation systems (i.e, single-purpose cognitive modules), depend is not a suitable approach to achieve lifelong learning~\citep{lecun2015deeplearning,Krizhevsky2012,Luong2015}. 
Using human-annotated training data is impractical for lifelong learning conducted by autonomous agents. 
In addition to being capable of unsupervised learning, PGMs representing elemental cognitive functions can be integrated to learn together (as described in Section 4). Owing to these features, we construct a cognitive architecture based on PGMs.


\delspan{Developing a cognitive architecture by integrating elemental cognitive modules provides us with a large degree of freedom with combinatorial complexity.
However, this raises questions such as what set of cognitive modules should be implemented? How can they be integrated to be able to work together? 
We attempt to answer these questions by combining two approaches: (1) {\bf brain-inspired AI}: learning human brain architecture to build human-level intelligence and (2) {\bf a PGM-based cognitive system}: developing an integrative cognitive system for developmental robots by integrating PGMs. Based on both arguments, we propose a promising approach called WB-PGM.}

\addspan{Developing a cognitive architecture by integrating elemental cognitive modules provide us with a large degree of freedom with combinatorial complexity.}
 We can reduce the design space of the integrative cognitive system using the human or animal brain architectures as a reference model.
%
\addspan{In the rapidly advancing field of neuroscience, researchers are beginning to have some comprehensive knowledge of the human and animal brain architectures and their anatomy, as well as the various neural activities that take place in it.
However, such knowledge is not organized in a way that is suitable for effectively constraining the design space of cognitive architectures.
Therefore, using the WBA approach, we define a data format called the brain reference architecture (BRA), which is a reference model of the brain, and describe knowledge in the field of neuroscience in a standardized way suitable for that application \citep{Yamakawa2021-qa}.}
\delspan{
Neuroscience studies have already revealed many connections between human and animal brains and their functions ~\citep{Hassabis2017-by,Tsotsos2014-yg}. 
Integrating such distributed knowledge about the brain via architectural description, we can use the human brain as an effective reference to build human-like intelligence.}
\delspan{We refer to the reference model as a brain reference architecture (BRA).}

\addspan{Adjacent research areas include biologically inspired cognitive architectures ~\citep{Samsonovich2016-tn, Goertzel2010-hy} and cognitive computational neuroscience~\citep{ Kriegeskorte2018-xw}, which is an interdisciplinary field of cognitive science and computational neuroscience. However, compared to the BRA, these fields have not made progress in accumulating design data in a standardized manner. In neuroinformatics~\citep{Amari2002-rp,Pradeep2013-gb,Crasto2007-yb}, which develops data and knowledge bases for neuroscience, progress has been made in experimental data on anatomical structures~\citep{Kuan2015-tl} and physiological phenomena~\citep{Poldrack2017-ni}. However, no progress has been made in accumulating data such that it can be used to design cognitive and behavioral functions, as in the BRA-driven development.}

However, even when we seek to build an AGI based on human-brain architectures, that is, using the WBA approach and having a reference model for the architecture, two challenges must be overcome.
 First, there are numerous alternative machine learning approaches to choose from to develop and integrate elemental models into a cognitive architecture. Nevertheless, if each elemental cognitive module is developed based on random machine learning approach, it will become difficult to integrate them in a coherent framework. Second, developmental cognitive architectures should be able to make all cognitive modules learn together based on the real-world sensory-motor information obtained by a robot.
This is crucial for achieving lifelong learning. However, most elemental cognitive modules, that is, AI functions, have been developed independently under different design principles.
To overcome this problem, a PGM-based approach is proposed in this paper to create cognitive architectures for developmental robots.
We propose that developing elemental cognitive modules and integrating them based on the theory of PGMs, specifically, the SERKET framework, can solve the above-mentioned problems ~\citep{nakamura2017serket,taniguchi2020neuro} (as described in Section~\ref{subsec:serket}).

Inspired by the human brain architecture, we propose an approach wherein a cognitive architecture is built by integrating PGM-based cognitive modules to fully mimic the human cognitive system. The framework for the integrative model is called WB-PGM.
This perspective paper describes the approach and the rationale, the current status, and the future challenges of WB-PGM.
The remainder of this paper is organized as follows.
Section~\ref{sec:2} describes the rationale \addspan{and design principle} for our approach \addspan{including its background}. Section~\ref{sec:3} reviews the relevant literature on elemental cognitive modules that can constitute the WB-PGM.
Section~\ref{sec:4} describes the PGM-based cognitive system integrating multimodal information (i.e., the world model) and the SERKET framework, which enables the efficient integration of PGM-based elemental cognitive modules.
Section~\ref{sec:5} describes a possible example of a WB-PGM and the future path of the development, interdisciplinary communication, and collaboration, specifically the fusion of AI and brain science. Section~\ref{sec:6} summarizes the conclusions.

\section{\addspan{Rationale and Design Principle for WB-PGM}}\label{sec:2}
To develop a cognitive architecture for realizing embodied AGI, we require a theoretical and practical development framework that enables researchers to efficiently design integrative artificial cognitive systems capable of lifelong learning, similar to the human brain. Such systems must integrate multimodal information and developmentally learn numerous cognitive skills. The proposed WB-PGM is a development framework that can contribute to overcoming these problems. This section describes the perspective of the WB-PGM and the rationale behind it.

\subsection{WB-PGM} \label{subsec:WB-PGM}
To develop a multimodal cognitive system, beyond the development of machine learning algorithms and signal processing modules that emphasize performance of a specific task, it is necessary to consider the following points: 
1) machine learning algorithms (theory) considering the correspondence with the structure of the brain, 2) realization of the whole structure rather than only a part, 3) interaction between the body, including the sensors and the real environment, 4) developmental learning considering timeline, and 5) consideration of computational and energy efficiency.
\delspan{Currently, it is challenging to build a system that considers all of these factors. However, it is possible, in principle, to generate complex structures by combining PGMs. This approach has the potential to build functions and information structures that are equivalent to those of the brain.}
Here, we describe the ideas, current trials, and challenges of the WB-PGM for developing a holistic approach.

Several related studies have investigated function-based modelling of the entire brain. 
For example, {\cite{Eliasmith1202}} proposed a neural architecture, ``Spaun,'' which models the entire brain function. This study aimed to elucidate the mechanism by bridging the gap between the neuron response and function as a whole. {\cite{sagar2016}} developed a system, ``BabyX,'' that simulates the entire brain functions.
A sensor-motor system using a functional model of the brain could generate realistic facial expressions.
These studies aimed to realize a holistic model, similar to our study. 
However, simulating the complexity of the brain functions as a whole has not been realized. 
This is because these systems are difficult to extend. In addition, they have focused on limited tasks. 
\delspan{We believe that it is important to realize interactions with the physical environment and long-term learning by integrating it with robots. 
Furthermore, we believe that the ease of analysis of the learning process is also important; thus, we consider the interpretability of PGMs as an important factor in our study.}
\addspan{
Further, an appropriate cognitive architecture for developmental robotics should enable a robot to perform a wide range of tasks in the real environment with its body through interactions and long-term learning capabilities.}

The idea behind the WB-PGM is to combine the recent developments in generative models from the field of machine learning with the latest knowledge of the structure and functional level of the entire brain. 
Thus, it aims to reproduce the flexible cognitive functions of humans, which cannot be achieved by the current single-purpose-oriented AI, which is often built on ``discriminative'' models. 
Moreover, neuroscience, which tends to study partial regions and specific functions of the brain (with a so-called worm's-eye view), may not efficiently grasp the whole structure at once. 
\addspan{Therefore, adopting the bird's-eye approach that can be operated/verified as a whole using our proposed WB-PGM model may help realize AGI.}\delspan{Therefore, it is necessary to create a bird's-eye view that can be operated/verified as a whole using WB-PGM.}

\begin{figure*}[t]
  \centering
  \includegraphics[width = 0.9\linewidth]{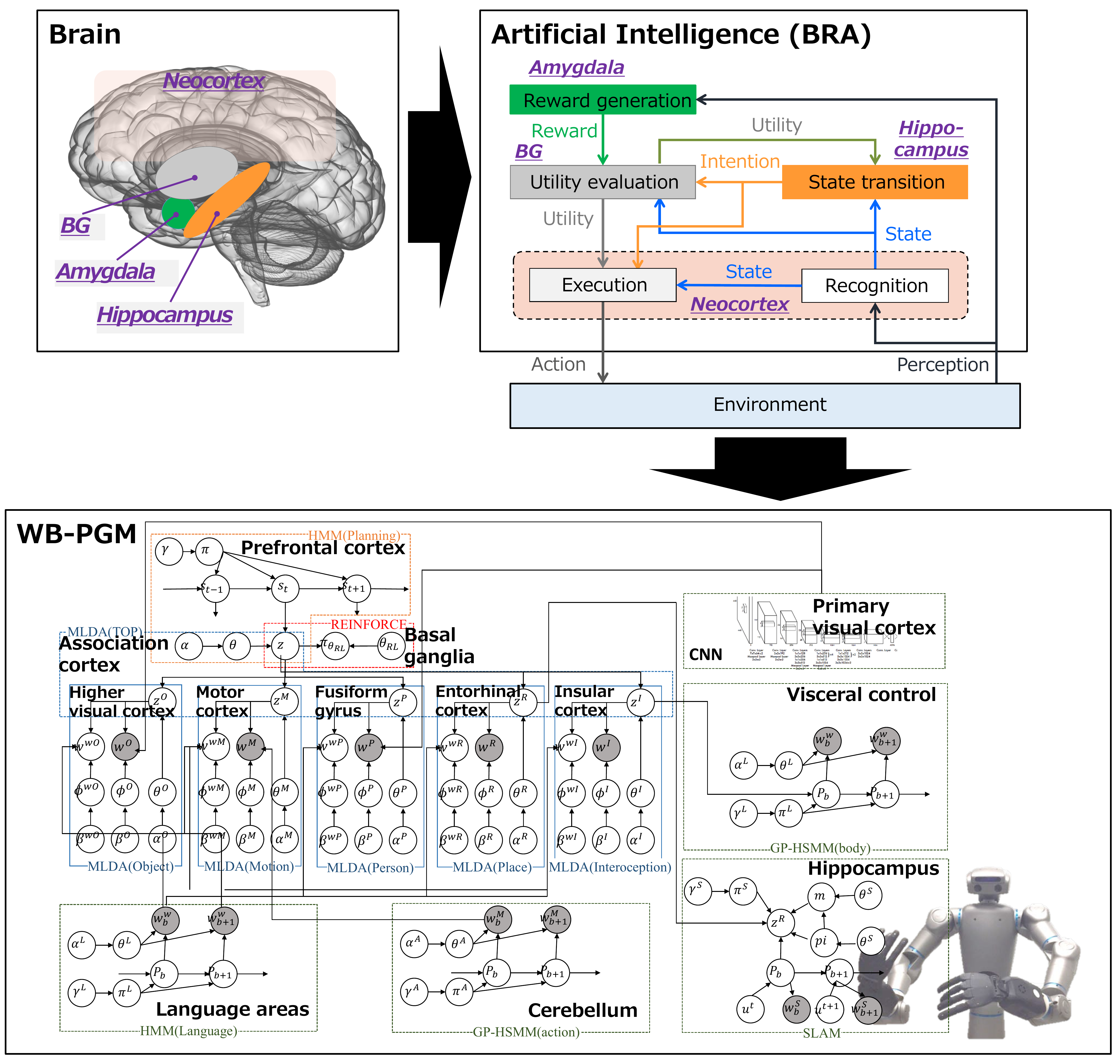}
  \caption{\addspan{Constructing WB-PGM based on the functional structure of the brain via BRA. Please note that this figure is intended to provide an overview of the WB-PGA development process. Model refinement using evaluation process (see Section~\ref{subsec:eval}) is also considered.}}
  \label{fig:wbr2wbpgm}
\end{figure*}

\delspan{It is not easy to match the structure of the brain with a machine learning model. Therefore, it is necessary to determine the level of thought. 
In this respect, the structure of the brain should refer to the brain reference architecture (BRA), as explained in Section~{\ref{subsec:BRA}. }}
\addspan{However, mapping the structure of the entire brain onto a machine learning model is not an easy task. This is because the neuroscience knowledge required to build it is vast and complex. Further, it is extremely difficult for a particular individual to design software for the entire brain. To solve this problem, we propose to use BRA-driven development that has been cultivated in the WBA literature (see Section~{\ref{subsec:overviewOfDevelopment}}).} 
\addspan{One of the most significant challenges facing the WB-PGM approach is matching the WB-PGM and BRA. Currently, efforts to match them are underway \citep{Taniguchi2021hpf-pgm} (see Section 2.4.3 as an example of this effort). 
Fig.~{\ref{fig:wbr2wbpgm}} illustrates an overview of the WB-PGM. }

\delspan{In contrast, in machine learning, the world model can mimic the human brain structure at the functional level. Currently, efforts to match the WB-PGM and BRA are underway . We developed a prototype of an integrated cognitive model based on the structure proposed by Doya . We are working on coordination with the BRA, as shown in Fig.}

\addspan{Miyazawa et al. have developed a representative prototype of an integrated cognitive model~\citep{miyazawaIROS2019,miyazawaFRONT2019} based on the structure proposed by Doya \citep{doya1999} instead of using the BRA. }
Hence, the prototype is based on the hypothesis that the cerebellum, basal ganglia, and cerebral cortex are specialized in supervised, reinforcement, and unsupervised learning paradigms, respectively~\citep{doya1999}. Our first focus is on the unsupervised learning paradigm in the cerebral cortex, which is realized by PGMs that map observations to latent \delspan{valuables}\addspan{variables}. This paradigm is considered as a basic module, while the reinforcement learning (RL) module corresponding to the basal ganglia and the motor control part of the robot that can be considered as supervised learning corresponding to the cerebellum are connected~\citep{miyazawaFRONT2019}. Furthermore, in the prototype, HMM is connected as the temporal learning mechanism to enable the robots to perform planning for longer periods of time using dynamic programming such as the Viterbi algorithm~\citep{miyazawaIROS2019}. 
\addspan{Various prototypes of WB-PGM, inspired by the above basic hypothesis of the brain, were implemented in an actual robot to enable basic learning ~\citep{araki2013,nishihara2017,miyazawaFRONT2019}. }

\addspan{For example, \citet{nishihara2017} showed that a robot can actually learn object concepts and associations among concepts and words through interactions with humans using many real objects. 
\citet{miyazawaFRONT2019} extended this model further and showed that the robot can learn to use objects through its own experiences.
Moreover, \citet{taniguchi2020improved} showed that the robot can acquire the concept and name of a place through interactions with its human counterpart.
}

To deal more closely with the anatomical structure of the brain, we need to proceed with coordination between the PGM-based cognitive architecture and BRA\delspan{, which will be described later}\addspan{ (see Sections~\ref{subsec:overviewOfDevelopment} and \ref{subsec:construct})}. 
\addspan{This process also leads to the challenge of scaling-up the combination of PGMs to achieve a large-scale WB-PGM (see Sections~\ref{sec:3}) and \ref{sec:4}}

\delspan{
Based on these, the roadmap for the WB-PGM was realized as follows.
The directions can be broadly divided into two categories: model refinement and communication modeling. 
This is because PGMs are highly compatible with the interarea communication of the cerebral cortex (see Table \ref{tab:counterstream}) and are currently being implemented mainly in the cerebral regions.
}

\delspan{The model refinement of the individual brain can be realized as follows:}
\delspan{
1)(Rough) connection of the entire brain: Implement a framework to connect the whole brain, and build the whole system by the current correspondence between BRA and probabilistic graphical models through generation-inference process allocation (GIPA) (see Section). Experiments were performed by connecting and implementing the robots. 
At present, this is at a very rudimentary level; however, the simplest function is constructed using PGMs; furthermore, robotics experiments, such as, are being conducted.\\
2)Refinement of each region such as hippocampus, cerebellar, and amygdala: Detailed correspondence regarding the hippocampus is in progress . 
}
\delspan{It is necessary to promote the development by repeating 1) and 2).}

\subsection{PGMs for Cognitive Systems} \label{subsec:PGM}

The idea that the human brain functions can be described using various machine learning modules is unprecedentedly popular. 
In particular, PGM-based machine learning modules learn to predict observations. 
Improving this prediction capability is a general criterion for mathematical models of cognitive systems. 
PGMs have been used to explain brain functions in many contexts from a Bayesian perspective.

For example, it is believed that the hippocampus performs simultaneous localization and mapping (SLAM)~~\citep{Ball2013-hi,Tolman1948, Okeefe1978placecells}. From a theoretical viewpoint, to perform SLAM, it is often assumed that actions, states, and observations follow a POMDP, and a map is defined as a global parameter of an observation model. Meanwhile, the Bayesian inference on the POMDP, a type of PGM, is regarded as a function of localization and mapping. 

The hypothesis that the basal ganglia are responsible for reinforcement learning is also widely accepted~\citep{Barto1995,Montague1996,Doya2007}. Conventionally, RL is formulated as a problem in which an agent optimizes its policy function to maximize the expected cumulative future rewards in an environment modeled as an MDP, a type of PGM~\citep{sutton2018reinforcement}. However, the concept of {\it control as probabilistic inference} clarifies that RL can be converted into an inference problem on an extended PGM for MDP~\citep{Levine2018} (see Section 3.3). This also inspired us to model brain functions using PGMs.

In a series of studies on symbol emergence in robotics, many unsupervised learning systems for robots have been developed based on PGMs to help observe multimodal sensory signals, model environments, acquire languages, and adopt behaviors~\citep{Taniguchi2016symbol}.
Multimodal latent Dirichlet allocation (MLDA) is a basic example of such a process ~\citep{nakamura_grounding}. 
The MLDA was able to integrate visual, auditory, haptic, and linguistic sensor information and form object categories without supervision. Many variants and extensions of the MLDA, including the nonparametric Bayesian extension, have been proposed~\citep{nakamura2011multimodal,nakamura2011bag}. 
A series of studies have suggested that a PGM-based approach has the potential for building an integrative cognitive system for developmental robots.

Advancements in deep learning have extended the capability of PGMs by inducing deep PGMs (DPGMs)~\citep{kingma2013auto,goodfellow2014generative,suzuki2016joint}.
DPGMs use deep neural networks as part of the generative process. They can extract features, that is, they are capable of representation learning.
In the 2010s, most PGMs proposed in conventional studies to develop integrative cognitive systems in symbol emergence in robotics were based on classical models of probabilistic distributions such as categorical, Gaussian, Wishart, and Dirichlet distributions~\citep{Taniguchi2016symbol}. However, these conventional integrative PGMs were not capable of feature extraction.
To tackle this problem, DPGMs can enable us to develop more flexible cognitive systems, maximizing the representation learning capability of deep neural networks.

Based on this evidence, we argue that a PGM is a reasonable approach to describe the whole brain cognitive system.

\subsection{ Overview of BRA-driven development for WB-PGM}
\label{subsec:overviewOfDevelopment}
\addspan{As already mentioned in Section \ref{subsec:WB-PGM}, to build a versatile cognitive system similar to a human, the entire system needs to be designed on a large scale. However, since that design space is vast, it is advantageous to constrain the design space by mimicking the architecture of the brain. As a way to embody this, we use BRA-driven development extended for PGM.}

\begin{figure*}[t]
  \centering
  \includegraphics[width = 0.75\linewidth]{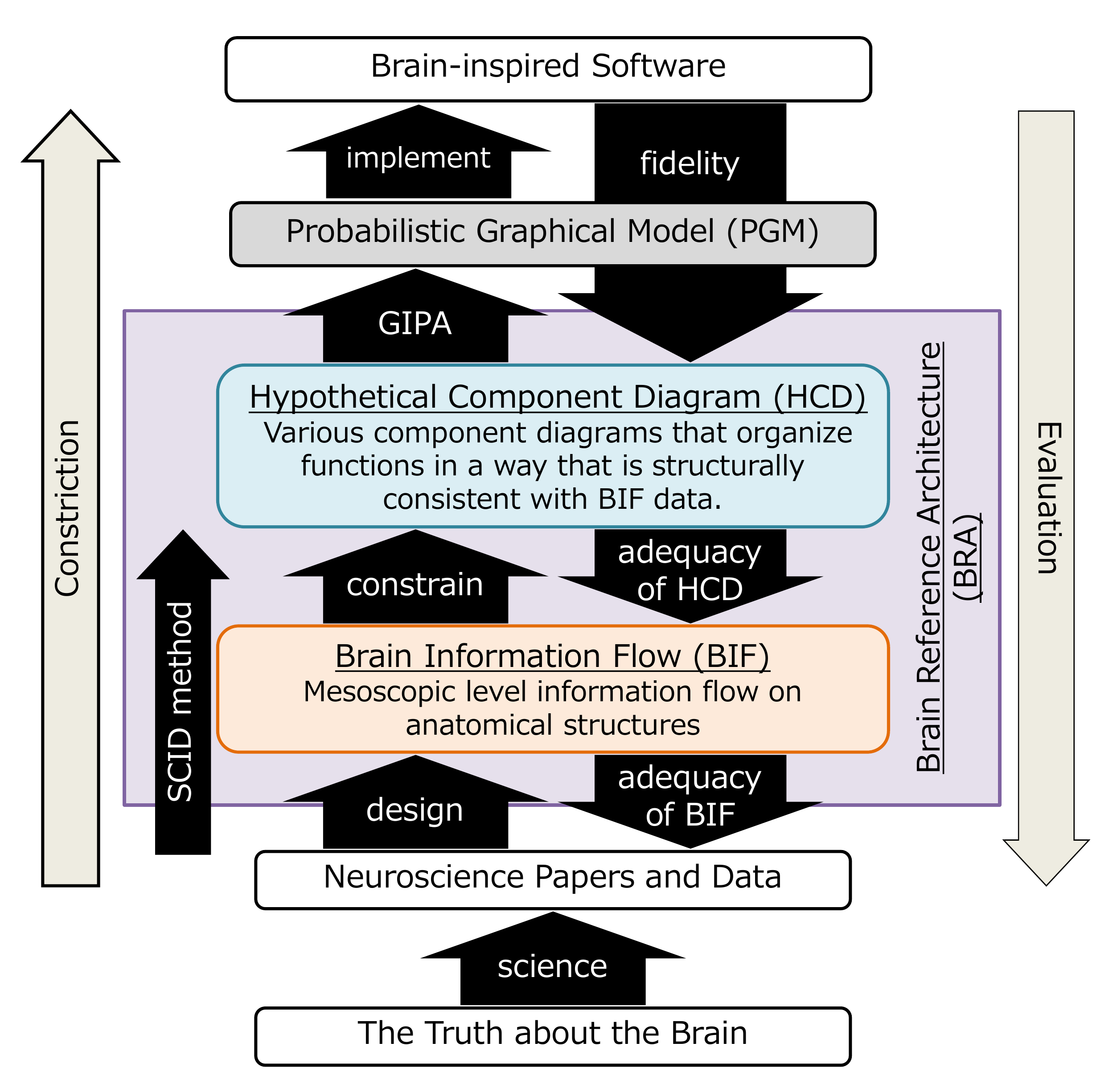}
  \caption{WB-PGM development as extended BRA-driven development.  
  A development method that extends BRA-driven development, which utilizes BRA data such as BIF and HCD, by adding GIPA, thus creating PGMs. The upward direction is the construction process, while the downward direction is the evaluation process.}
  \label{fig:BRA}
\end{figure*}

\addspan{BRA-driven development is a method of constructing software that reproduces human-like cognitive functions by referring to the neural circuits of the entire brain \citep{Yamakawa2021-qa}. BRA data, which play a central role in BRA-driven development, form a reference model consisting of a brain information flow (BIF) , which extracts mesoscopic-level anatomical knowledge related to human cognitive behavior, and a hypothetical component diagram (HCD), which shows the structure of functional components organized consistently with respect to the BIF, as shown in Fig. \ref{fig:BRA}.}

\addspan{The BRA-driven development process consists of the construction and evaluation processes, mediated by BRA data.
In the conventional construction process, BIF is first designed by the structure-constrained interface decomposition (SCID) method described below, HCD is created under the constraints of BIF, and then, brain-type software is implemented using the HCD as specification information. In the current WB-PGM development, a process called GIPA is added to convert HCD into PGM before implementation (see Section \ref{subsec:construct}).
In contrast, the evaluation process consists of ``adequacy evaluation,'' which confirms that the BRA is consistent with existing brain science findings, and ``fidelity evaluation,'' which evaluates whether the brain-type software is implemented in a manner consistent with the BRA.}

\addspan{As shown in Fig.~\ref{fig:BRA}, BRA-driven development separates the design of the BRA from the implementation of the software based on it, thus enabling multiple brain scientists and multiple software developers to collaborate on large-scale development.}

\addspan{BRA data \citep{Sasaki2020-zp} essentially consist of BIF and HCD. More precisely, findings from neuroscience describe not only BIF, which is anatomical structural information, but also neural activity and the processes that constitute it, which are omitted in this description.}

\noindent
\textbf{BIF:}
A BIF is an information flow diagram that describes the anatomical structure of the entire brain at the mesoscopic level, without assuming any specific task \citep {Arakawa2020-ep}.
It is constructed by analyzing findings from neuroscience, such as data and connectome \citep{Negishi2019-pt}.
It consists of nodes, called circuits, and directed links, called connections (see Fig.~\ref{fig:BIF-HCD-PGM} (A)).
Consequently, BIFs can provide the basis for an architecture that combines numerous computational mechanisms.
To prevent too fine granularity, the lower bound of granularity is defined as a population of (sub-)types of neurons that are considered almost homogeneous and belong to the same brain organ.
This population of neurons is called a uniform circuit (see Fig. \ref{fig:BIF-HCD-PGM} (A)).

\noindent
\textbf{HCD:}
An HCD is a diagram that breaks down the task or function performed by a particular region of interest (ROI) in the brain into its functional components, where the structure of the functional decomposition must be consistent with the mesoscopic level anatomical structure (see Fig. \ref{fig:BIF-HCD-PGM} (B)).
A component diagram is a major type of diagram in the unified modeling language for modeling the structure of object-oriented software \citep{Bell2004-xg}. It illustrates the static aspects of the operating principles of software through a network of components that perform computational functions and the semantics of the dependencies among those components in any complex system. As demonstrated subsequently, the PGM developed in this study is based on the component diagram.

\subsection{\addspan{Construction of WB-PGM}} \label{subsec:construct}

\addspan{The construction process of the WB-PGM consists of the design process of BRA by the SCID method and the conversion process from HCD to PGM by GIPA, as described below.}

\subsubsection{SCID method for designing BRA} \label{subsubsec:scid}

\addspan{The SCID method is a protocol for designing HCD, which is brain-inspired software specification information, based on knowledge from the neuroscience field 
\citep{Yamakawa2021-qa, Fukawa2020-hl,Yamakawa2020-xa}.
The method constructs a BIF for a specific ROI based on anatomical knowledge and then designs an HCD that can realize the top-level function (TLF) of the ROI consistent with the BIF.}

\addspan{The SCID method consists of the following three steps. However, Step.1 and Step.2 are often executed back and forth.}

\begin{itemize}
    \item \addspan{Step 1: BIF construction: Surveying anatomical knowledge in ROI}
\item \addspan{Step 2: Determining ROI and TLF consistently}  \par
  \quad\quad \addspan{Including the creation of provisionally component diagrams}
\item \addspan{Step 3: HCD creation}
    \begin{itemize}
        \item \addspan{Step 3-A: Enumeration of candidate component diagrams}
        \item \addspan{Step 3-B: Rejection of HCDs that are inconsistent with scientific knowledge}
    \end{itemize}
\end{itemize}

\addspan{Conventional computational neuroscience models neural activity in the brain by interpreting it based on changes in the external world. However, the range of neural activities that can be interpreted in the brain is limited. However, the SCID method can organize the structure of functions to achieve its goal, consistent with anatomical knowledge of the brain, at the mesoscopic level.
Since this anatomical knowledge is now being accumulated over a relatively wide region of the brain, the SCID method can potentially be used to create HCD over a wide region of the brain.}

\subsubsection{\addspan{GIPA for mapping HCD to PGM}} \label{subsubsec:gipa}
This section provides a foothold for building a probabilistic graphical model representation of a PGM that is consistent with the \addspan{BRA data} \delspan{brain neural circuits}.
As mentioned in Sections~\ref{subsec:overviewOfDevelopment} and \ref{subsubsec:scid}, the HCD designed using the SCID method represents a dependency structure between components corresponding to each specific brain region. 
Thus, to construct a probabilistic graphical model corresponding to a brain neural circuit, the dependency interfaces in the HCD should be classified such that they correspond either to the generative or inference processes. 
This task is called GIPA~\citep{Taniguchi2021hpf-pgm}.
Here, the links in both the generative and inference processes can be considered to be specialized dependencies. 
In other words, GIPA should be performed for every interface~\citep{Taniguchi2021hpf-pgm}.

\textbf{Problems:}
In preparation, the dynamic recurrent property should be discussed with respect to PGMs and brain structures.
First, there are many loops in the brain's anatomical circuits, whereas a PGM needs to be a directed acyclic graph.
In most cases, it is difficult to assign acyclic PGMs to brain circuits\footnote{However, the occurrence of apparent loops in the static structure, which is obtained by degenerating the temporal evolution of an acyclic PGM, is not a problem.}.

The PGM is a model that provides a consistent link structure in the data generation process using directed links that represent the signal transfers between random variables\footnote{Each of these variables is, in principle, associated with a `uniform circuit,' (see \ref{subsec:overviewOfDevelopment}), which is the minimum descriptive unit of the BIF.}. 
When inferring latent variables, an inference model is used to calculate the posterior probability distribution of the latent variables conditioned by the observed values. 
In ordinary PGMs, signal propagation in the inference process causes signals to propagate in the opposite direction to the links used in the generation process.
In contrast, in brain neural circuits, signal propagation between the regions by electrical spikes that propagate terminally on axons is essentially unidirectional.
Therefore, to realize a PGM in its normal form, the condition, ``whenever there is a connection between two regions, it is a mutual connection," must be satisfied in the brain. 
However, in most regions of actual brain neural circuits, satisfying this condition is difficult.

\begin{table}[!b]
  \begin{center}
    \caption{Counter stream pathways in the neocortex}
    \begin{tabularx}{\linewidth}{p{40mm} p{40mm} p{40mm} } 
    \hline
    Pathways & Feedforward & Feedback \\ \hline
    Direction & From the outside world to the internal state (from lower to higher areas) & From the internal state to the outside world (from higher to lower areas) \\ \hline 
   Laminar on cortical microcircuits \citep{Markov2014-ez,Markov2013-zd} & Layer 3, and Layer 4 & Layer 2\\ \hline
    Meaning of signals \citep{Yamakawa2020-xa} & Observation & Prediction \\ \hline
    Graphical model representation of PGM & inference process & generative process \\ \hline
    \end{tabularx}
    \label{tab:counterstream}
  \end{center}
\end{table}

\begin{figure*}[t]
  \centering
  \includegraphics[width = 0.7\linewidth]{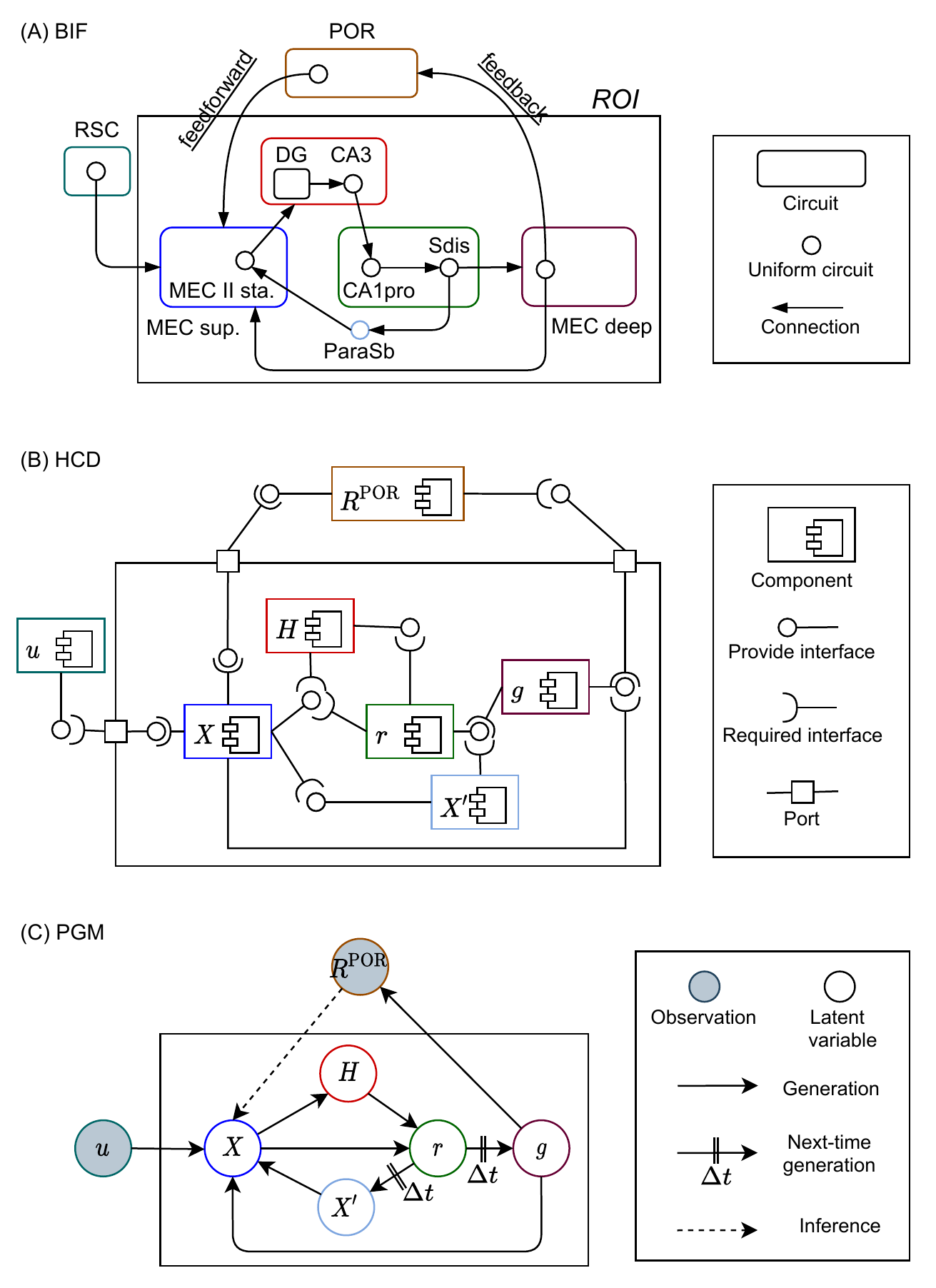}
  \caption{
    BIF, HCD and graphical model representation of PGMs.
    Corresponding areas in the three figures are indicated by same-colored borders.
    (A) BIF with the hippocampus and medial entorhinal cortex (MEC) as the ROI. This figure is obtained by organizing the anatomy at the mesoscopic level of the brain.
    Projection from POR to the MEC sup represents feedforward coupling, whereas the projection from the MEC deep to the POR is feedback coupling.
    CA1 pro: Cornu ammonis 1 proximal;
    CA3: Cornu ammonis 3;
    DG: Dentate gyrus;
    MEC deep: MEC, deep layers;
    MEC II sta: MEC, Layer 2, stellate cell;
    MEC sup: MEC, superficial layers;
    ParaSb: parasubiculum;
    POR: Postrhinal cortex;
    RSC: Retrosplenial cortex;
    ROI: Region of Interest;
    Sdis: Subiculum distal.
    (B) HCD associated with the above BIF.
    The diagram is constructed using the SCID method to assign functions to the components that correspond to the BIF and perform SLAM functions.
    $r$:              Cluster information regarding positions; 
    $H$:              Pattern separation/completion, information integration; 
    $X$ and $X^{\prime}$: Self-posture; 
    $g$:              Prediction at the future time regarding movement/speed amount or posture; 
    $R^{\text{POR}}$: Allocentric visual information;
    $u$:              Rotational speed movement.  
    (C) A probabilistic graphical model created from the above HCD.
    Probabilistic graphical models are converted from an HCD by GIPA, that is, assigning all interfaces on that HCD to either generative or inference processes. 
    The flat arrow with $\Delta t $ indicates the generation of the variable in the next time step.
  }
  \label{fig:BIF-HCD-PGM}
\end{figure*}

\textbf{Solution strategy:}
To avoid this problem, we adopt an amortized inference to define the link structure of the inference process independently of the link structure of the generative process~\citep{Gershman2014}. 
The amortized inference is a type of variational inference, an approach that introduces functions for efficient approximate inferencing of latent variables.
A typical example of this is the VAE, which is an auto-encoding variational Bayesian model~\citep{kingma2013auto}.
Modeling using amortized inference is illustrated in Fig.~\ref{fig:BIF-HCD-PGM} (C). 
The model used for the amortized inference can be designed with a high degree of freedom, as long as it is consistent with the link structure of the generative process. 
Therefore, in this type of probabilistic graphical model, it is easy to relate the link structure to the actual structure of the brain neural circuits.

The major interarea connections of the neocortex can be allocated to either of the generation or inference processes.
In the neocortex, there is a feedforward pathway that transmits signals from lower to higher areas while processing signals received by sensors, and a feedback pathway that transmits signals in the opposite direction \citep{Markov2014-ez,Markov2013-zd} (see Table \ref{tab:counterstream}).
In computational neuroscience theories, such as the Bayesian brain \citep{Doya2007-of, Friston2012-cz} and predictive coding \citep{Rao1999-ym}, inference and generation are assumed to be processed by the feedforward and feedback pathways, respectively. 
The HCD for the neocortical interarea connections was designed based on these findings \citep{Yamakawa2020-xa}.
During GIPA in the neural circuits adjacent to the cortex, care must be taken to avoid inconsistencies in GIPA at the interface with the cortex.

\clearpage
\clearpage

\subsubsection{\addspan{Construction example: PGM for hippocampal formation}}

\delspan{Taniguchi et al. (2021a) illustrated this process using a portion of the results of a probabilistic graphical model constructed while performing GIPA on regions of the hippocampal formation (HPF), which includes the entorhinal cortex and hippocampus.}

\addspan{An example of PGM by applying GIPA to the hippocampal formation (HPF)\footnote{HPF is a brain organ containing the entorhinal cortex and hippocampus.} \citep{Taniguchi2021hpf-pgm} is illustrated in this section.}
As shown in Fig.~\ref{fig:BIF-HCD-PGM}, first the BIF of the hippocampus and entorhinal cortex (Fig.~\ref{fig:BIF-HCD-PGM} (A)) is constructed. Then, the HCD (Fig.~\ref{fig:BIF-HCD-PGM} (B)) is designed using the SCID method.
The PGM on HPF (Fig.~\ref{fig:BIF-HCD-PGM} (C)) is constructed by performing a GIPA.
The dotted arrows represent the inference process, while the lined arrows represent the generative process.


The connection from POR to MEC II superficial on the BIF can be regarded as a feed-forward pathway.
According to Table~\ref{tab:counterstream}, an inference process can be allocated to the connection from variable $R^{\rm POR}$ to variable $ X $ in the probabilistic graphical model representation.
Similarly, the connection from the deep MEC to the POR on the BIF can be regarded as a feedback pathway. 
According to Table~\ref{tab:counterstream}, a generation process can be allocated to the connection from variable $ g $ to variable $R^{\rm POR}$ in the probabilistic graphical model representation.
Nevertheless, there exist limitations to performing GIPA in terms of the inside of the hippocampus when considering only the connectivity with the neocortex.
Therefore, the engineering formulation of the SLAM modeled as PGM is used as a reference.

The graphical model representation of the PGM for HPF in Fig.~\ref{fig:BIF-HCD-PGM} (C) is constructed to be consistent with SLAM's PGM using the GIPA procedure based on the above discussion. 
For engineering SLAM, we consider a PGM that estimates the future self-posture $X(t+1)$ directly from the current self-posture $X(t)$.
In contrast, the self-posture $X$ at the next time in HPF is generated via variables such as $H$, $r$, $X'$, and $g$.
Since the probabilistic graphical model representation of the HPF in Fig.~\ref{fig:BIF-HCD-PGM} (C) degenerates over time, there is a circulation in the generation process, at a glance.
To make it clear that circulations with time advancing, (e.g., POMDP and state-space models such as the Kalman filter) are acceptable for PGMs, the notation ``next time generation process'' is introduced.
Generation with one-time step progress is represented by a double line orthogonal to the generation arrow, plus the symbol $\Delta t$.
Note that there is arbitrariness of position at which the time progress can be allocated in the loop of PGMs\footnote{It is arbitrary in the sense that the International Date Line can be placed at any longitude on Earth.}.

\subsection{\addspan{Evaluation of WB-PGM}}\label{subsec:eval}
To ensure that the developed software mirrors the brain efficiently, the evaluation method proposed for BRA-driven development \citep{Yamakawa2021-qa} can be used. It comprises the two evaluations described below.

The first method entails evaluating the software adequacy by estimating the consistency between existing neuroscientific findings and BRA. The second method is the fidelity evaluation, wherein the reproducibility of the BRA in the brain-inspired software is evaluated. 
\delspan{Fidelity was evaluated based on structural similarity, functional similarity, activity reproducibility, and performance for normal systems. 
Changes due to dysfunction are often also subject to evaluation.}

\subsubsection{Evaluating adequacy}
\addspan{Adequacy evaluation can be divided into that for BIF and that for HCD.}

\noindent
{\bf 1) Adequacy evaluation of BIF}

\addspan{The consistency of the anatomical structures and neural activity described in the BIF with those described in neuroscientific papers and data is evaluated.}

\addspan{Two main inspection criteria are used to verify that the description of BIF is sufficient. The first criterion is ensuring that the description element of the structure or phenomenon that is provided in the data submitted for registration is not already registered in the BRA database (novelty). The other criterion is that the element must be directly or indirectly supported by current neuroscientific findings (authenticity). As a rule, the authenticity of facts is guaranteed by their inclusion in one or more peer-reviewed articles.}

\noindent
{\bf 2) Adequacy evaluation of HCD}

\addspan{The functionality of the HCD and its consistency with the BIF are evaluated to determine whether the process generated by the behavior of the structured components in the HCD can achieve the goals of the ROI.}

\addspan{The consistency evaluation determines whether the HCD corresponds to the description of the BIF according to two aspects:
\begin{enumerate}
    \item The dependency structure of the HCD corresponds to the anatomical structure contained in the ROI of the BIF. 
    \item The behavior of the components within the HCD is consistent with the physiological findings described in the BIF.
\end{enumerate}}

\subsubsection{Evaluating fidelity}
\addspan{
The biological plausibility of brain-inspired software is evaluated by comparing it with BIF and HCD in the BRA data. The estimated degree of consistency between the software and BRA is referred to as the fidelity.}

\addspan{To date, four methods have been explored for the evaluation of fidelity.
\begin{itemize}
  \item {\bf Structural similarity:} An evaluation of how strongly the static structure of the software matches the BIF in the BRA.
  \item {\bf Functional similarity:} An evaluation of how strongly the behavior of a particular component that is implemented during the execution of a specific task matches the behavior (e.g., behavior timing) that is designed in the HCD in the BRA.
  \item {\bf Activity reproducibility:} An evaluation of how effectively the behavior of a certain variable in the internal components of the software implemented according to the BRA reproduces the characteristics of neural activity (e.g., activity timing and patterns in the corresponding brain region during the execution of a specific task).
  \item {\bf Performance:} An evaluation of the performance and ability of the software as a whole (integrative testing).
\end{itemize}}

\addspan{Among these evaluation methods, structural similarity and performance are easy to use for the evaluation of the whole software. Nevertheless, functional similarity and activity reproducibility are useful for unit tests for each component as well as for integrative development. Furthermore, it is possible to consider an evaluation method wherein dysfunction states are induced by intentionally destroying/ablating parts of the software and comparing them with the brain functioning under conditions such as mental illness or brain injury.}

\subsection{WB-PGM for Lifelong-learning Robots (AGI)}

At some point in the future, BRA may be able to cover the entire brain, the PGM-based design that is being developed based on BRA may also cover entire brain, and the construction of WB-PGM may help in realizing lifelong-learning robots, that is, AGI.
A PGM-based approach is suitable for developing an integrative cognitive architecture for lifelong-learning robots because it enables integrative cognitive systems to perform unsupervised learning that does not require human-annotated data for training. Since PGM-based cognitive systems learn internal models, representations, and behaviors by inferring latent variables of the system, they can maximize the marginal likelihood of sensory-motor observations, with the learning process requiring only sensory-motor information (i.e., performing unsupervised learning). This process is also called predictive coding. In particular, the WB-PGM is expected to be able to adapt the entire cognitive system for lifelong learning by exploiting findings from the fields of neuroscience and cognitive science in its development.

Many scientific fields attempt to reach a better understanding of the human mind, including cognitive science, neuroscience, robotics, with each field adopting a unique approach. In particular, cognitive architectures are often regarded as the standard model of a {\it human-like mind}~\citep{Laird2017}.  

The study of cognitive architectures has a long history, with architectures proposed, including Soar, ACT-R, Sigma, and their variants~\citep{laird2012soar,rosenbloom2016sigma,anderson2009can}. Traditional models were based on symbolic AI, whereas more recent approaches such as neural networks and PGMs have also been considered for modeling cognitive architectures.
However, although most studies on cognitive architecture adopted a top-down theoretical approach, they were rarely implemented and tested on embodied artifacts in real-world physical and social scenarios such as cognitive and social robotics. That is, they were not based on real-world multimodal information, unlike modern AI (e.g., deep learning-based pattern recognition and synthesis as well as concept formation and representation learning~\citep{lecun2015deeplearning,mmlda}.

Therefore, the proposed WB-PGM should be validated by evaluating it on real-world tasks. As human intelligence has evolved throughout the history of environmental adaptation, our cognitive systems were developed to survive in a real-world environment, involving physical and social interactions. Consequently, a human-like cognitive system, that is, AGI, should be evaluated in a real-world environment where the human cognitive system evolved. 
In other words, if some tasks cannot be achieved in a real-world environment by a developed AI system, then the corresponding aspects of human intelligence are missing. Thereby, a cognitive system should learn various skills by organizing multimodal sensory-motor information observed by the system itself. For this purpose, the cognitive system ought to have a body to actively and autonomously explore the physical and social environment. 
That is, it should be tested using robots that have bodies to act in a real environment. \addspan{This is part of the performance test for evaluation fidelity mentioned in Section~\ref{subsec:eval}). }

From the viewpoint of brain science, where researchers seek to understand how human minds and brains work in greater detail, a standard model, such as the WB-PGM, can provide top-down guidance to interpret actual experimental results, which is always obtained from a partial cognitive process and under limited conditions. Furthermore, such a model can suggest new experiments for efficiently uncovering the mystery of the human brain and cognition. Therefore, the cognitive architecture in this study is located at the intersection of neuroscience, AI, and robotics. 
There is strong evidence that the proposed WB-PGM can also extend neuroscience studies.

\section{PGM-based Cognitive modules}\label{sec:3}
The WB-PGM is being developed by integrating cognitive modules into a single PGM.
A wide range of elemental cognitive modules has been developed with regard to their respective cognitive capabilities. This section provides a survey of the PGM-based cognitive modules to further integrate them to realize the WB-PGM.

As mentioned in the Introduction, if the entire brain-like cognitive structure can be constructed as a unified model called PGM, its development will be more efficient. However, the question remains as to whether each part in the brain can be treated as a PGM. \delspan{The details are discussed later.} Here, we outline that the computational elements of each area of the brain involved in higher-order cognitive functions can be represented generally as a module of the PGM.
\\
Historically, information processing, primarily in the visual cortex, has been modelled as a PGM. However, all neocortical areas are composed of canonical microcircuits with homogeneity to some extent \citep{Douglas1989-ql, Bastos2012-wv, Beul2014-kj}, and the network between neocortical areas constitutes a counter stream of feedforward and feedback systems \citep{Markov2013-zd}. As this mechanism allows us to regard the flow of observational and predictive signals as opposing \citep{Yamakawa2020-xa}, it acts as the basis for the PGM mechanism for the whole brain. Recently, it has also been pointed out that the PGM mechanism in the neocortex has the duality of cognition and control \citep{Doya2021-dk} (See also Section \ref{subsec:World Model}).
Additionally, it is reasonable to consider the hippocampus, which is connected to the neocortex via the entorhinal cortex, as a PGM because it performs computations similar to SLAM, which is essentially described as a PGM \citep{Taniguchi2021hpf-pgm}.
Further, the basal ganglia and amygdala can estimate the desirability of a certain state from the system's viewpoint, and the neocortex can use this information to perform optimal control as a PGM (see Section \ref{subsec:value}).
Moreover, the cerebellum can also be thought of as a mechanism that accelerates the PGM by partially extracting the computations performed by the neocortex and basal ganglia and quickly performing alternative computations (see Section \ref{subsec:action}).


\subsection{Visual perception and representation learning}
\label{subsec:visual}
The mammalian brain has a minimum of two processing modules for each sensory modality, one
for action and the other for recognition or consciousness \citep{ungerleider1982two, goodale1992separate, sakagami2007functional}. For example, the visual information in the retina is sent to the primary visual cortex (V1), which includes two cortical pathways. The dorsal pathway from V1 to the parietal cortex determines the spatial layout of objects and computes their disposition for actions. The ventral pathway from V1 to the inferotemporal cortex mediates object recognition and contributes to the formation of our cognitive world. Anatomical data further reveal that the parietal cortex projects primarily to the premotor and dorsolateral prefrontal cortices (DLPFC)~\citep{pandya1982intrinsic}. Specifically, the reciprocal connection between the parietal cortex and DLPFC contributes to spatial attention and spatial working memory~\citep{constantinidis2016neuroscience}. In contrast, the inferotemporal cortex has many efferents to the prefrontal cortex, especially the ventrolateral prefrontal cortex (VLPFC)~\citep{ungerleider1989projections}. In particular, the projection from the inferotemporal cortex to the VLPFC appears to be critical for concept formation (e.g., categorization) and generation of new information (deductive inference)~\citep{pan2008reward, tanaka2015dissociable}. Recently, ~\citet{bengio2013representation} have demonstrated that their machine learning algorithm could successfully simulate representation learning in the ventral pathway.



In representation learning, a good representation entails generalizability to arbitrary tasks, with various hypotheses proposed as properties that such a representation should satisfy~\citep{bengio2013representation, goodfellow2016deep}. Inspired by the idea of human concept formation, one of the most important proposed hypotheses is disentanglement~\citep{higgins2016beta}, which holds that each element of a representation should be semantically meaningful.
For example, if we have a picture of a cat, we observe that the picture consists of various meaningful elements, such as the cat type, its orientation, and the position of the light source.



\begin{figure*}[t]
  \centering
  \includegraphics[width = 0.8\linewidth]{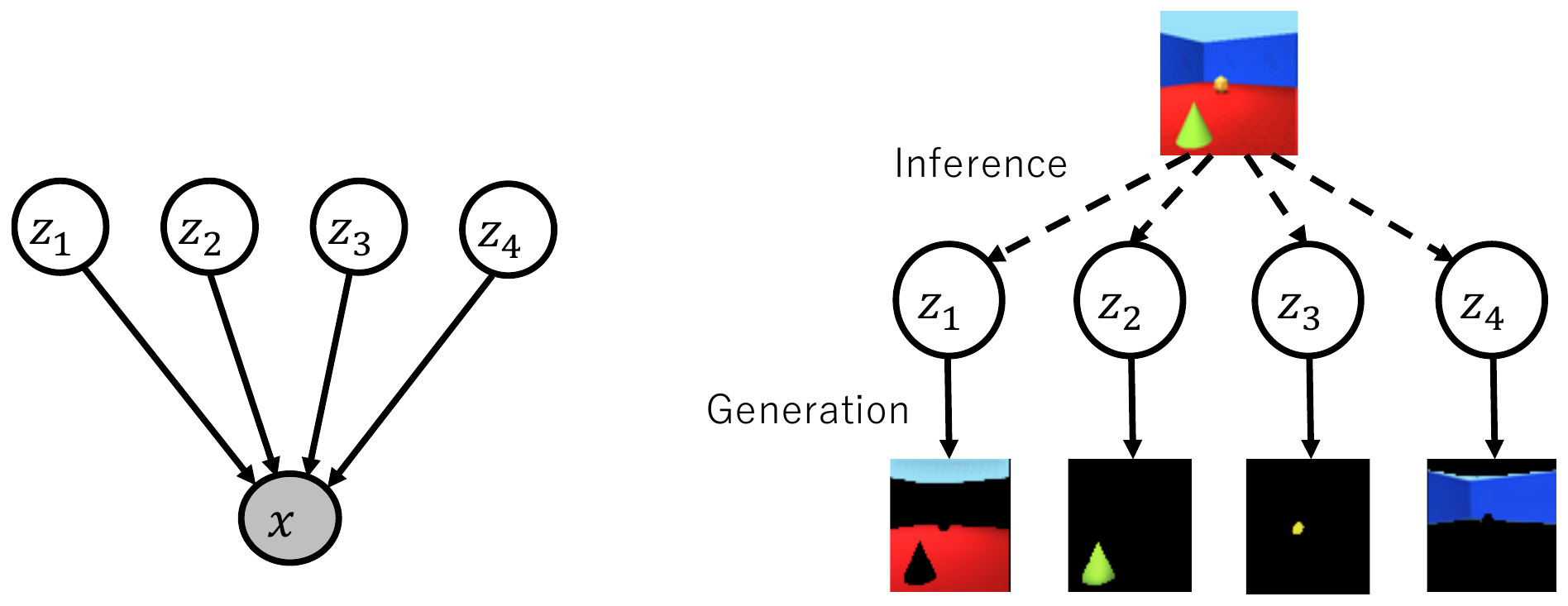}
  \caption{Graphical model of scene interpretation and its process of image decomposition. The decomposed images were generated using GENESIS~\citep{engelcke2019genesis}.}
  \label{fig:scene_interpretation}
\end{figure*}

Further, scene interpretation corresponds to the ventral pathway in the brain. It is the study of recognizing the images of multiple objects using VAEs in an unsupervised manner to decompose the disentangled representation corresponding to each object. In this study, the models are designed to assume several latent variables corresponding to the objects and generate decomposed images from them (Fig.~\ref{fig:scene_interpretation}).
\citet{eslami2016attend} proposed the attend-infer-repeat (AIR) approach, which decomposes the latent variables of each object into what the object is and its location in the image and infers the latent variables of the objects in the image.
The AIR approach can recognize and reconstruct each object end-to-end on images of multiple handwritten digits.
\citet{kosiorek2018sequential} extended the AIR approach temporally by introducing latent variables that correspond to objects that have been present since the previous step and those that only just appear in the current step.
Furthermore, to properly decompose images with multiple complex objects, an approach was introduced, whereby a mask is employed for each object as a latent variable, which is also inferred in the recognition process ~\citep{greff2019multi, burgess2019monet, engelcke2019genesis}.

\subsection{Value and Reinforcement}
\label{subsec:value}
The reward and value system is fundamental to the survival of a biological system in a given environment. 
In the RL theory, a value function is defined as the expected sum of future rewards. 
Neuroscience studies have revealed several brain areas that play major roles in RL, including the amygdala and basal ganglia.
These areas receive strong projections of dopaminergic neurons, which have been demonstrated to encode reward prediction errors \citep{Schultz1997}.
The projection from the cortex to the basal ganglia has distinct forms of plasticity, depending on the presynaptic input and postsynaptic spike output, followed by the dopamine input, that is, the dopamine-dependent plasticity \citep{Reynolds2001,Yagishita2014,Iino2020}.
Owing to these observations, RL models of the cortico-basal ganglia circuit have been proposed \citep{Barto1995,Montague1996,Doya2007}.
A specific hypothesis, which has been supported by neural recording experiments \citep{Samejima2005,Pasquereau2007,Lau2008,Ito2015}, is that the neurons in the basal ganglia learn state and action value functions \citep{Doya2000}.
The circuit of the basal ganglia is composed of multiple pathways starting from the striosome and matrix compartments in the striatum \citep{Yoshizawa2018} and the direct and indirect pathways downstream \citep{Hikida2010}. Incidentally, recent RL algorithms for robust and efficient performance use multiple types of value functions \citep{Haarnoja2018,Wang2020}, possibly hinting at the need for multiple pathways in the basal ganglia.


In addition, effective RL critically depends on the representation of states and actions.
The cerebral cortex provides multimodal, hierarchical representations of states and actions through unsupervised representation learning and inference of hidden variables \citep{doya1999}.
Although backpropagation in a deep Q-network solves the problem of value-oriented representation learning, it is known to be highly data-demanding \citep{Lake2017}.
Furthermore, although representation learning in the cortex appears to be unsupervised, experimental observations suggest that learning is modulated by reward or value signals \citep{Bao2001,Seitz2009}.
A recent study that applied variational recurrent neural networks to RL demonstrated that task-critical latent variables can be learned \citep{Han2020iclr}.

Moreover, to achieve fast learning and fine control, it is important to select the right level of abstraction.
The amygdala and the cortico-basal ganglia circuit appear to form a hierarchical RL system.
The evolutionarily old amygdala is crucial for immediate actions for primary reward and punishment. The amygdala is composed of a cortex-like lateral part and a basal ganglia-like central part \citep{Cassel1999}, which may be seen as a prototype cortico-basal ganglia circuit.
The cortico-basal ganglia circuit is composed of multiple parallel loops: the limbic loop through the ventral striatum, the prefrontal loop through the dorsomedial striatum, and the motor loop through the dorsolateral striatum \citep{Voorn2004,Ito2015,Balleine2015}.
These parallel loops appear to form a hierarchical RL system, spanning different levels of abstraction \citep{Haber2000,Voorn2004,Ito2015,Balleine2015}.
Determining the right set of action options and their combinations is an active area of research \citep{Bacon2017,Han2020nn}.

\subsection{Action planning and control}
\label{subsec:action}
Although model-free RL provides a generic solution to control problems, learning requires many trials, and evidence suggests that humans deploy model-based strategies using action-dependent state transition models or forward models \citep{Wolpert1998}.
The classic framework for model-based optimal control is dynamic programming based on the Bellman equation \citep{Bellman1952}.
The similarity between the equations for optimal state inference and optimal control is known as the Kalman duality \citep{Kalman1960}, which indicate the similarity between the computation of the log posterior in dynamic Bayesian inference and the state value function in optimal control \citep{Todorov2008,Levine2018,Doya2021-dk}.
The framework is recognized as ``planning as inference'' \citep{Botvinick2012} or ``control as inference''(CaI) \citep{Levine2018}.

\begin{figure*}[t]
  \centering
  \includegraphics[width = 0.4\linewidth]{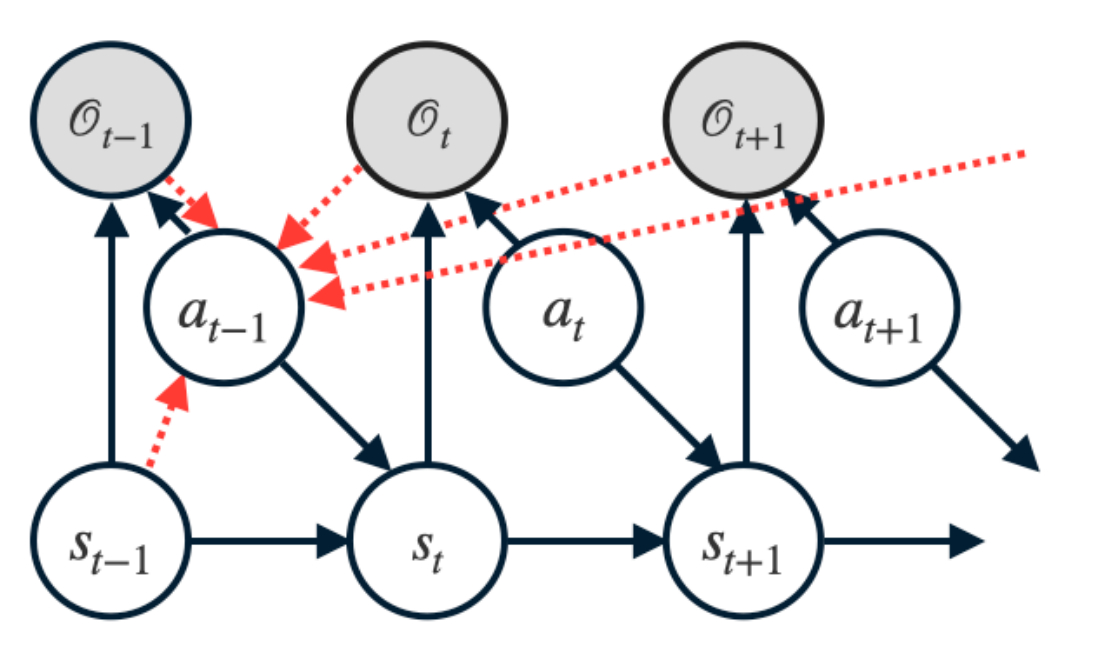}
  \caption{MDP model featuring optimality variable $\mathcal{O}$ \citep{Levine2018}. The dotted line represents the inference to action from the optimality variable over the future and the current state, which is the optimal policy.}
  \label{fig:CaI}
\end{figure*}

\citet{Levine2018} introduced a binary optimality variable that indicates whether the state and action at each time in the MDP are optimal and formulated the reward function as the probability for the optimality variable to take one (Fig.~\ref{fig:CaI}).
In the CaI framework, we can derive the entropy regularized expected reward objective by performing variational inference for the optimality variable at all times, from which we can derive the soft actor-critic (SAC)~\citep{haarnoja2018soft}. In addition, we can derive an iterative planning method based on the inference of the plan, that is, a series of actions, instead of policy optimization, as in the SAC. ~\citet{okada2020variational} demonstrated that the difference between various planning methods can be generalized as the choice of the posterior distribution in the inference of optimality variables.
This idea was extended to POMDP settings as well~\citep{planet_b}. 

In the cerebral cortex, while the posterior half is mostly involved in sensory inference, the anterior half is mostly involved in control and planning. Thereby, the CaI framework can provide an answer to the basic question of why common circuit architectures can be used for both inference and control \citep{Doya2021-dk}.
An important difference between the sensory cortex and the motor or frontal cortex, in addition to the thickness of different layers, is that the latter receives inputs from the cerebellum and the basal ganglia through the thalamus.
Further, while probabilistic dynamic models and value computation are realized in the cortical circuit, these subcortical circuits may provide useful shortcuts.
The cerebellum has been proposed to provide deterministic forward models learned by supervised learning \citep{Wolpert1998,doya1999}, which can supplement probabilistic models in the cortex.
Additionally, the basal ganglia can provide learned value functions \citep{doya1999,Daw2005} to complement online computations of value functions.
The network linking the cortex, basal ganglia, and cerebellum is involved in motor learning \cite{Tanaka2018} and action planning \citep{Fermin2016}, and their exact roles are a topic of active research.

\subsection{Spatial cognition and mapping}
\label{subsec:spatial}



In neuroscience, it has long been assumed that the HPF, consisting of the hippocampus and entorhinal cortex is responsible for functions such as episodic memory, spatial cognition, and response inhibition. Interestingly, memories are transferred to the neocortex through the phenomenon of memory replay and consolidation during sleep.

Thus, HPF has various functional aspects. However, for the following reasons, it would be useful to learn from the brain by focusing on spatial cognitive functions that play an important role in the navigation of mobile robots.
This is because the SLAM technology~\citep{thrun2005probabilistic,ref:taketomi2017visual}, which combines the functions of self-position estimation and map formation, has been formulated as PGM. Thereby, this function can be naturally incorporated as part of the WB-PGM.
Furthermore, as described below, there has been extensive neuroscientific research on spatial cognition related to HPF, mainly in rodents.

Further, involvement in spatial cognitive abilities in the hippocampus has long been considered responsible for cognitive maps~\citep{Tolman1948, Okeefe1978placecells}.
There have also been epoch-making discoveries of space-encoding cells such as place, border \citep{Solstad2008-ko,Savelli2008-dl}, head-direction \citep{Taube1990-vm, Taube1990-oi}, and grid cells.
Place cells are neurons that are active in specific locations within the hippocampus~\citep{Okeefe1978placecells}, while grid cells are neurons in the MEC that are cyclically active as rodents and other animals move through space~\citep {Hafting2005gridcells}.

Essentially, spatial cognitive abilities involve the transformation of self-centered (egocentric) information obtained directly from sensors into a representation of world-centered (allocentric) information. In particular, grid cells contribute to the representation of this world-centered coordinate system. Since the beginning of the 21st century, research on the relationship between the HPF and the posterior parietal lobe, which represents self-centered and world-centered information, has been conducted \citep{Whitlock2008-zq,Wilber2014-ob,Wilber2017-dp}.

Subsequently, we discuss the evolution of the SLAM technology from a practical perspective. The SLAM-related mathematical theory and implementation have made rapid progress in the last decade.
SLAM models can be represented by PGMs based on the POMDP.
PGM-based SLAM models are estimated based on Bayes filters such as landmark-based~\citep{montemerlo2002fastslam} and grid-based~\citep{gridbasedfastslam2007} FastSLAM.

\begin{figure*}[t]
  \centering
  \includegraphics[width = 0.7\linewidth]{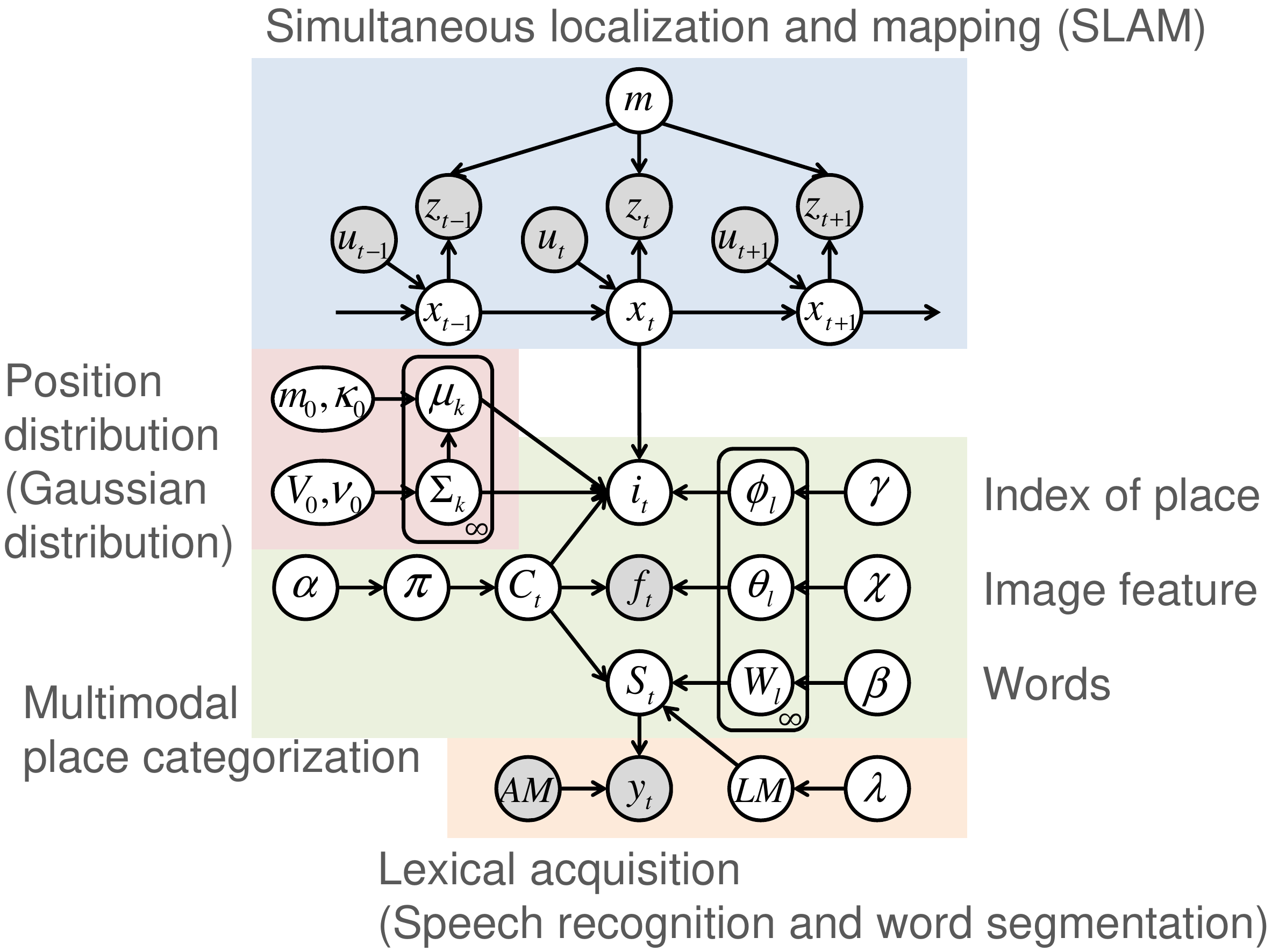}
  \caption{Graphical model of SpCoSLAM \citep{taniguchi2017online}, which has self-position, environmental map, position distribution, multimodal place categories, word sequences, and language models as latent variables. 
  The blue, red, green, and orange areas represent SLAM, the position distribution, the multimodal place categorization, and speech recognition and word segmentation, respectively. 
  Using a Rao-Blackwellized particle filter procedure~\citep{doucet2000rao}, SpCoSLAM can infer these model parameters and latent variables.}
  \label{fig:graphical_model_SpCoSLAM}
\end{figure*}

Furthermore, the semantic mapping approach, which includes the meaning of places and objects, has been actively developed as the next direction of SLAM \citep{kostavelis2015semantic} due to its effectiveness in performing human-robot interaction tasks.
Specifically, it is important to appropriately generalize and form place categories while dealing with observation uncertainties.
To address these issues, PGMs for spatial concept formation have been constructed~\citep{taniguchi2016spatial,taniguchi2017online,hagiwara2018hierarchical,Katsumata2020SpCoMapGAN}.
\citet{taniguchi2017online} proposed the spatial concept formation using SLAM (SpCoSLAM), that is, place categorization and mapping through unsupervised online learning from multimodal observation.
SpCoSLAM is an integrated PGM composed of SLAM, a Gaussian mixture model, a multimodal Dirichlet process mixture model (MDPM), and speech recognition, as shown in Fig.~\ref{fig:graphical_model_SpCoSLAM}.
\citet{Katsumata2020SpCoMapGAN} successfully transferred global spatial knowledge related to multiple environments to a new environment by integrating the spatial concept model with generative adversarial networks (GANs).
This approach to spatial concept formation was also adopted for tasks in the World Robot Summit~\citep{lotfi2018WRS,ataniguchi2020TidyUpHere}.
We consider the above models as candidates for a cognitive module with functions similar to the HPF.
Details are discussed in \citet{Taniguchi2021hpf-pgm}. 


\subsection{Social interactions and inference}
\label{subsec:social}


Social behavior is an individual’s behavior that arises from the interaction with others under certain circumstances. Specifically, humans have a special characteristic in their prosocial behavior, that is, the tendency to benefit others by sacrificing their own profit. It is assumed that human prosocial behavior is calculated in the cortical model-based process~\citep{knoch2006diminishing}. However, \cite{rand2012spontaneous} demonstrated that human subjects tend to exhibit more prosocial behavior intuitively while being more selfish in deliberation. Furthermore, \cite{fermin2016representation} indicated that the subcortical areas, particularly the amygdala, play an important role in prosocial behavior, whereas cortical areas, specifically the prefrontal cortex, are critical for pro-self behavior. \cite{yamagishi2017response} suggested that the prosocial bias, which was related to the dependency of model-free or model-based processes, varied across individuals. Thus, it can be conjectured that certain social habits are acquired under stereotyped stimulus-response circumstances through RL of model-free systems, whereas others are mediated by goal-directed calculation of model-based systems. However, many neural mechanisms underlying social behavior remain unclear.

Except for some aspects that will be subsequently explored, social communication is a broad concept that has not been sufficiently investigated from the angle of PGMs. 

Nevertheless, as social communication involves estimating and understanding the intention of others, intent estimation can be modeled as an inference of the latent variables of others to predict their behavior. 
In other words, since if we estimate a person's intention, we can predict their behavior more accurately, intention estimation can be modeled as a prediction problem to some extent.
In the early days following the postulation of this idea, \cite{wolpert2003unifying} argued that intention estimation can be modeled using multiple forward-inverse models. However, the discussion can be reinterpreted with PGMs in a more sophisticated manner. Intention is regarded as a latent variable within a system, that is, the inference of the latent variable can be regarded as intention estimation. 

However, if a person attempts to model each person's behavior by assuming latent variables, the complexity of the model increases. For example, when we play football, we apparently do not predict the players' behavior one by one. Thus, PGMs that enable a cognitive system to predict the behaviors of a group of people and conduct a cooperative task are required.

\subsection{Speech recognition and language}
\label{subsec:speech}
Language is a representative fruit of the evolution of human species. It has enabled us to transfer knowledge and form communities and societies through communicating complex information by combining linguistic symbols using our vocalization system. An important aspect of the human language is that the symbolic system is not directly encoded in the biological gene (genome); rather, it is encoded in a cultural gene (meme). Knowledge is transferred using language. Further, language has non-biological and non-physiological effects on the development and function of the brain ~\citep{deacon1998symbolic}.

In particular, speech recognition and generation are fundamental parts of our spoken language. The motor theory of speech perception, which is well-known and widely debated in cognitive and brain sciences, argues that we use generative models of speech signals per utterance ~\citep{liberman1967perception,galantucci2006motor,laurent2017complementary}. In other words, it claims that the objects of speech perception are the speakers’ vocal tract gestures. According to this brain theory, a PGM-based approach is suitable for speech recognition and language. 

\begin{figure*}[t]
  \centering
  \includegraphics[width = 0.5\linewidth]{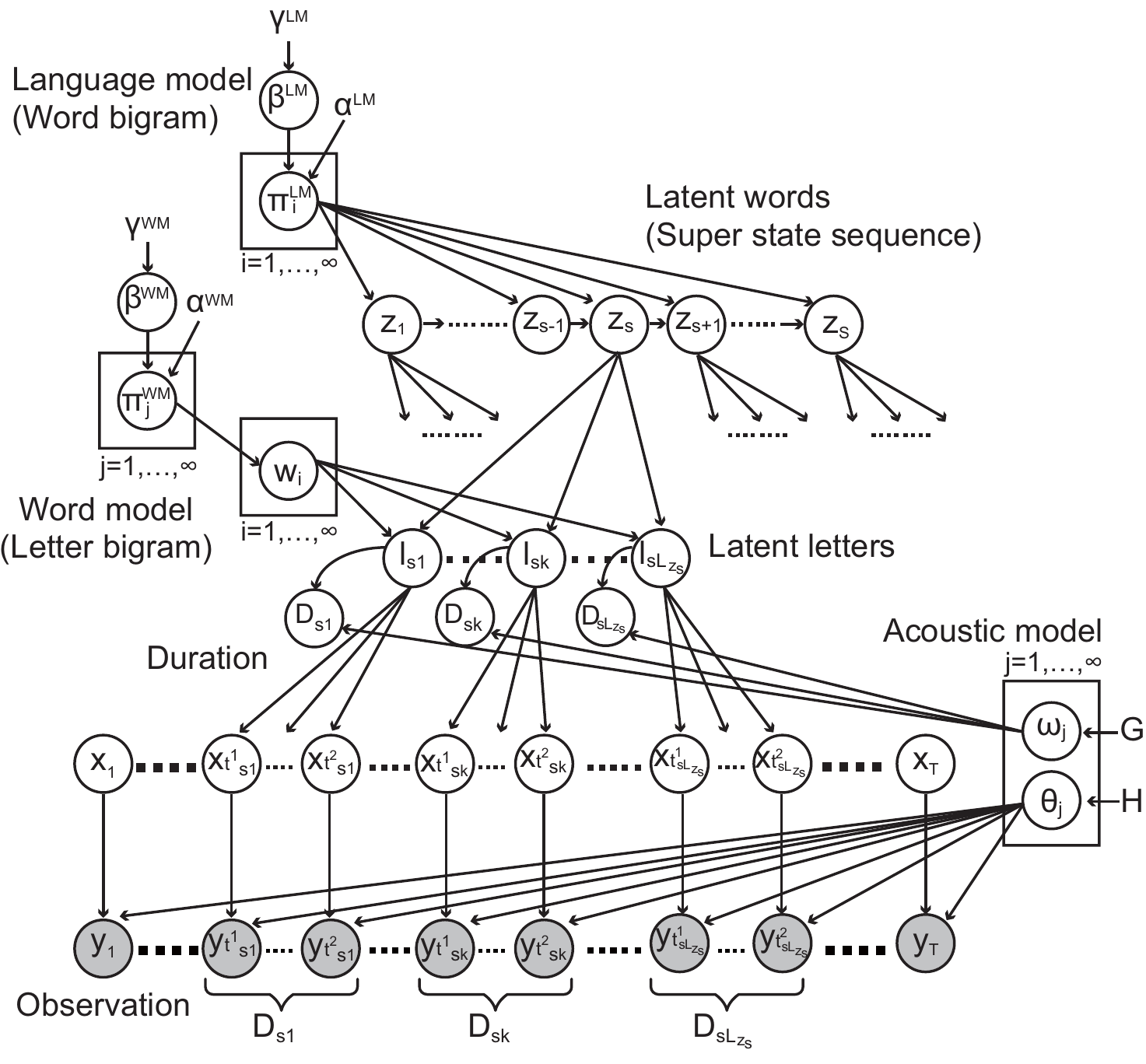}
  \caption{Graphical model of HDP-HLM \citep{taniguchi2016nonparametric}, which has language, word, and acoustic models as latent variables. Using a blocked Gibbs sampling procedure, HDP-HLM can infer these models and latent sequences of words (i.e., latent words) and phonemes (i.e., latent letters).}
  \label{fig:graphical_model}
\end{figure*}

Before deep learning-based approaches became dominant, speech recognition and synthesis were conventionally studied using PGMs ~\citep{verydeep}. The HMM, a type of PGM for time-series data, has been widely used for speech recognition systems~\citep{rabiner1986introduction,lee2009recent}. Generally, a speech recognition system is composed of an acoustic model and a language model. The acoustic model mimics the acoustic features of speech signals for phonemes. It is a generative model of acoustic features.
In contrast, the language model represents the generative process of word sequences. A word corresponds to a sequence of phonemes. This two-layer hierarchy for speech generation is typically applicable to spoken language and is termed ``double articulation.''
In particular, the hierarchical Dirichlet process-hidden language model (HDP-HLM) is a total generative model that involves language and acoustic models within a unified PGM (Fig.~\ref{fig:graphical_model}). \cite{taniguchi2016nonparametric} proposed HDP-HLM and derived a blocked Gibbs sampler for the PGM. It was demonstrated that the machine learning-based method can perform simultaneous phoneme and word discovery from only speech signals.

The syntactic nature of language has also been studied from the viewpoint of generative models for a long time. 
Probabilistic models for generative grammar, such as probabilistic context-free grammar and combinatory categorial grammar, assume that there exists a latent tree structure behind a sentence, that is, a word sequence~\citep{liang2007infinite,bisk2013hdp}. Thereby, inferring the latent structure corresponds to parsing in syntactic analysis.

Recently, a neural network-based approach to natural language processing has become dominant, with many GAN- and VAE-based methods for speech signal processing developed~\citep{van2020vector,kameoka2018stargan}.
Such DPGMs involving neural networks can exploit the advantages of deep learning for speech signal processing. Although HMM-like PGMs have become less popular in the late 2010s, this does not mean that a PGM-based approach is not valid, as the mathematical framework of PGMs involves DPGMs as well. 
Specifically, to leverage non-annotated data, unsupervised learning methods based on PGMs and self-supervised learning methods are promising. Self-supervised learning methods for language, for example, BERT and GPT-3, exhibit remarkable performance ~\citep{devlin2018bert,brown2020language}.    
In addition, when considering the integrative cognitive model involving not only speech recognition but also spoken language acquisition, (e.g., phoneme and word discovery), the PGM-based approach is still promising. 

Nevertheless, developing a cognitive architecture to achieve such linguistic communication remains a difficult challenge in brain and cognitive science, AI, and robotics~\citep{Taniguchi2019langrobo}.
For example, a service robot is required to understand human utterances whose meaning is not always explicitly determined. 
If speaker suggests or implies with an utterance using an implicature, such as, ``it is too hot,'' he or she actually implies ``please turn on the air conditioner'' or ``please wait a while until the coffee becomes less hot.'' In pragmatics, people believe that the meaning of an utterance is context-based. The context could involve a speaker, place, situation, and habit. Such information is not encoded by the utterance itself. Therefore, an assumption in which an utterance and meaning are regarded as the input of a function and output, respectively, is not plausible, as contextual information is provided by the multimodal sensor information and a history of interactions. This suggests that language understanding inevitably involves latent variables, and PGMs can model the cognitive system that facilitates linguistic communication.
Nevertheless, the social aspects of language are not limited to what we have described above. Although language is a social phenomenon, it is clearly realized by human cognitive systems.

In summary, the cognitive architecture that enables social interaction (see Section \ref{subsec:social}) and language processing (see Section \ref{subsec:speech}), which are described in this section, require a complex coordination of cognition and behavior. Such higher abilities are beneficial to the brain for learning. Given this situation, if all computational elements can be unified into PGMs, the efficiency of design and implementation is expected to improve.


\section{Integration of cognitive modules}\label{sec:4}
The WB-PGM is developed by integrating a wide range of elemental cognitive modules. This section describes the PGM-based approach to integrating cognitive modules. First, we revisit machine learning-based methods for world models that involve representation learning and the integration of multimodal information, including action and sensation, using PGMs.
The FEP, which provides a unified view of biological perception and behavior based on this PGM-based world model, is also introduced.
Finally, we introduce the SERKET framework, which allows us to develop an integrated PGM-based cognitive system, that is, WB-PGM, by combining elemental PGM-based cognitive modules.  

\subsection{World Models and FEP}
\label{subsec:World Model}

Humans can construct a mental model of the world by recognizing and learning from information obtained from the external world in a self-supervised manner.
This model, which represents our hypothesis about the world, can be used as a simulator to predict the unknown or the future based on current observations.
Furthermore, by incorporating our own behavior into this model, we can predict the future in the long term. A framework that realizes these human functions in machine learning is called a world model~\citep{ha2018world,hafner2019planet, hafner2019dream,hafner2020mastering}.

The key to realizing a world model is to compress the vast and multimodal information of the external world in a spatio-temporal manner to obtain their latent representations.
If such representations can be appropriately acquired, future predictions can be made easily by transitioning through representations on the time direction.
In addition, by assuming the structure of the world as an inductive bias for the model before learning, we can acquire a world model with higher generalizability more efficiently.
PGMs are good tools for designing a world model that meets these requirements.
Specifically, in the case of a time-evolving environment, POMDP models are often assumed because input stimuli are considered to be partial observations of the environment.
The generative process of observations designed by PGMs corresponds to predictions, with the inference of representations from those observations regarded as perceptions from the external world.
In the interarea signal transmission of the neocortex, the pathway responsible for prediction is called feedback. It is involved in the generative process in the PGM. In contrast, the perceptual pathway is called feed-forward and is involved in the amortized inference process in PGMs (see Table \ref{tab:counterstream}).

The world model has been studied as a model of the environment in model-based RL. 
However, early attempts to ``make the world differentiable'' using RNNs~ \citet{schmidhuber1990making} were not applied to large-scale environments because they were not equipped with representation learning techniques.
\citet{ha2018world} introduced a recurrent neural network (RNN)-based model that combined VAEs to enable spatial abstraction, and trained it on the trajectories, that is, the time series of images and actions, of complex game environments.
They proved that agents that were reinforcement-trained only on the world model behaved appropriately in real-game environments.
Therefore, we can obtain a model of the world that has sufficient predictive power for good representation learning, similar to mental imagery training in humans.

Recently, Planet~\citep{hafner2019planet} and Dreamer~\citep{hafner2019dream,hafner2020mastering}, which are model-based RL methods with more sophisticated deep generative models based on POMDPs, have shown high sample efficiency and performance in long-term control tasks based on a series of images in the environment.
Moreover, the uncertainty in the latent state representation of these models conveys the model's beliefs about the world; thus, the amount of information gleaned by the model from the environment is directly correlated to its level of certainty.
\citet{gregor2019shaping} introduced a decoder based on this world model on a two-dimensional map and demonstrated that although the generated map is blurry in the early stages of exploration in an unknown environment, it becomes more accurate with experience.

Further, the FEP is a promising theory that provides a unified view of biological perception and behavior based on a PGM-based world model.
\citet{von1867treatise} hypothesized that perception is the inference in an internal model. Subsequently, the Bayesian brain hypothesis and predictive coding were considered as works to minimize prediction errors or ``surprises.'' 
\citet{friston2005theory, friston2006free} extended these models with insights from statistical machine learning and thermodynamics, arguing that decision-making, as well as perception, is unified within a framework of variational inference or free energy minimization.

In addition, ~\citet{friston2010action, friston2015active} proposed active inference, in which the organism selects a sequence of actions to minimize the expected free energy, that is, \delspan{the expected value of free energy from the present to the future.}
\addspan{the expected value of the free energy with respect to an observation, considering the observation as a latent variable, i.e., an unknown to be obtained in the future.}
This framework is similar to that of CaI~\citep{Levine2018}, in that it considers decision-making as inference, However, they differ in the way they consider ``preferences'' for states and actions~\citep{millidge2020relationship}. CaI introduces preference as a new random variable (i.e., optimality) in the model, while CaI introduces preference as a new random variable (i.e., optimality) into PGMs, active inference considers that preference is encoded as the bias of PGMs.
Under this assumption, the terms corresponding to the extrinsic and intrinsic values are derived from the expected free energy. The extrinsic value refers to the value of exploitation, which encourages agents to be goal-oriented, whereas the intrinsic value refers to the value of exploration, which encourages agents to act to obtain novel observations. Therefore, based on only the expected free energy of active inference, we can derive the term corresponding to the trade-off between exploitation and exploration in RL. 

\subsection{SERKET and Multimodal Integration}\label{subsec:serket}
Similar to the human brain, a robot's cognitive system must integrate many types of sensory-motor information, find relationships between them, and utilize the organized internal representations. That is, the cognitive system must be very complex. MMLDA was proposed by combining many PGMs of MLDA~\citep{mmlda}.
Specifically, MLDA was combined with the nested Pitman-Yor language model (NPYLM), thus obtaining MLDA+NPYLM\footnote{The abbreviation, MLDA+NPYLM, used in this paper was not used in the original paper. It was however used in later articles.} to achieve unsupervised word segmentation using multimodal object category information formed by the MLDA~\citep{TomoakiNakamura2014}.
SpCoSLAM integrates PGMs for multimodal categorization, SLAM, and automatic speech recognition systems, which involves the word discovery capability. Thus, it achieves multimodal categorization and lexical acquisition about places~\citep{taniguchi2017online}.
By maximizing the trajectory probability based on the CaI framework, path planning based on semantic information can be conducted using PGMs, that is, SpCoSLAM. This method is called SpCoNavi~\citep{ataniguchi2020spconavi}.
These results clearly demonstrate that the PGM-based approach has the flexibility of integrating a broad range of cognitive modules and the capability to make them learn together.

The benefit of cognitive architectures integrating multimodal sensory-motor information is evident. First, when a unimodal signal does not have sufficient information and suffers from noise or uncertainty, the system can find latent structures (e.g., object categories and word units) using multimodal sensory-motor data~\citep{taniguchi2018unsupervised,taniguchi2020improved,nakamura_grounding,TomoakiNakamura2014}. Second, many cognitive functions, including localization using images and utterances, language understanding, and action planning, can be realized as a cross-modal inference within such multimodal cognitive architecture~\citep{mmlda,taniguchi2017online,ataniguchi2020spconavi}. Finally, owing to the second feature, we can construct multi-purpose and system-oriented cognitive architectures rather than a single-purpose or goal-oriented function.

Nevertheless, when attempting to develop a large-scale cognitive system involving a wide range of cognitive functions of the human brain, developing efficient large-scale computational models based on PGMs becomes a critical problem. 
The use of a probabilistic programming language (PPL) is a possible solution. Many types of PPLs that enable the efficient development of PGMs have been proposed ~\citep{goodman2012church,tran2017deep,bingham2019pyro,sato1997prism}. However, when developing a WB-PGM using only a PPL, it is necessary to re-implement each cognitive module using the PPL. However, this may cause a reusability problem from the viewpoint of software development. Thus, we need a development framework that allows us to use elemental PGMs developed in a heterogeneous and distributed manner.

SERKET is a framework for integrating PGM-based cognitive modules and developing cognitive architectures involving multiple cognitive functions~\citep{nakamura2017serket}.
Although building integrative cognitive systems using PGMs is a promising approach, designing an inference procedure for the developed PGMs sequentially is highly tasking. 
Further, modern integrative cognitive models for multimodal concept formation and language acquisition by robots (e.g., MMLDA and SpCoSLAM) involve many nodes; thus, variables and inference procedures are proposed for each model. However, we found that most of such integrative PGMs can be regarded as composites of several elemental cognitive modules, that is, PGMs, whose inference procedures have already been developed. SERKET provides a protocol with which the inference procedure of an integrative PGM is divided into inference procedures for each elemental PGM and communication among them.
In addition, Neuro-SERKET is a natural extension of SERKET~ {\citep{taniguchi2020neuro}} in that although SERKET does not support neural network-based PGMs, that is, DPGMs, Neuro-SERKET supports them.
Further, Neuro-SERKET allows elemental cognitive modules to learn in a heterogeneous manner. Each module uses different learning methods. For example, two modules trained using Gibbs sampling and variational inference can be integrated. Practically, this flexibility helps people develop an integrative cognitive system using pre-existing and distributed cognitive modules.
In the deep learning-based approach, differentiability is required throughout the system. In this sense, each cognitive module must be homogeneous from the viewpoint of optimization. In contrast, the SERKET framework allows us to use heterogeneous cognitive modules, that is, the learning and inference processes of each module can be encapsulated. This characteristic can improve the reusability of the elemental cognitive modules in a practical sense.

\begin{figure}[tb]
  \begin{minipage}[b]{0.55\linewidth}
    \centering
     \includegraphics[keepaspectratio, scale=0.6]{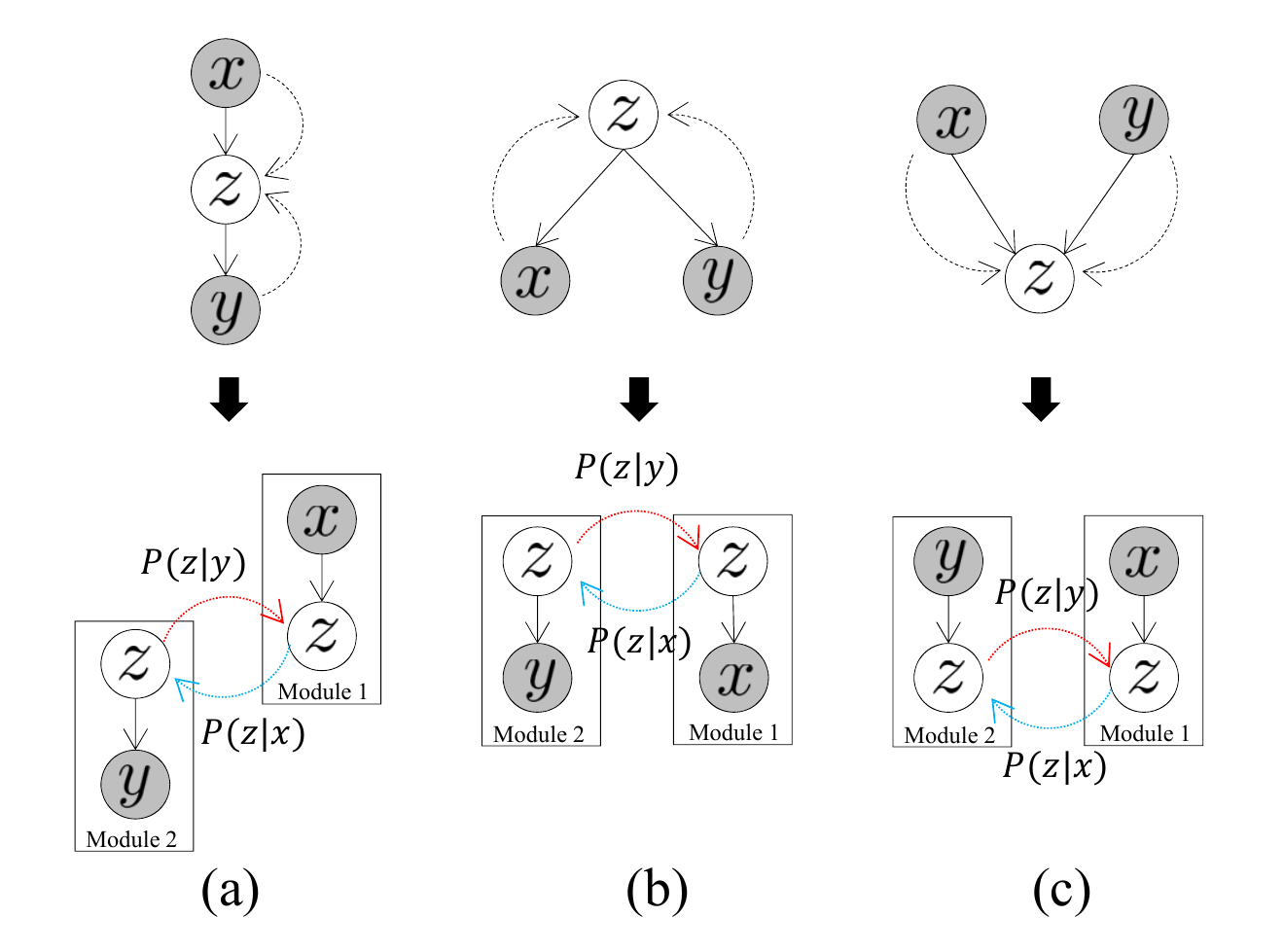}   
  \end{minipage}
  \begin{minipage}[b]{0.35\linewidth}
    \centering
    \includegraphics[keepaspectratio, scale=0.8]{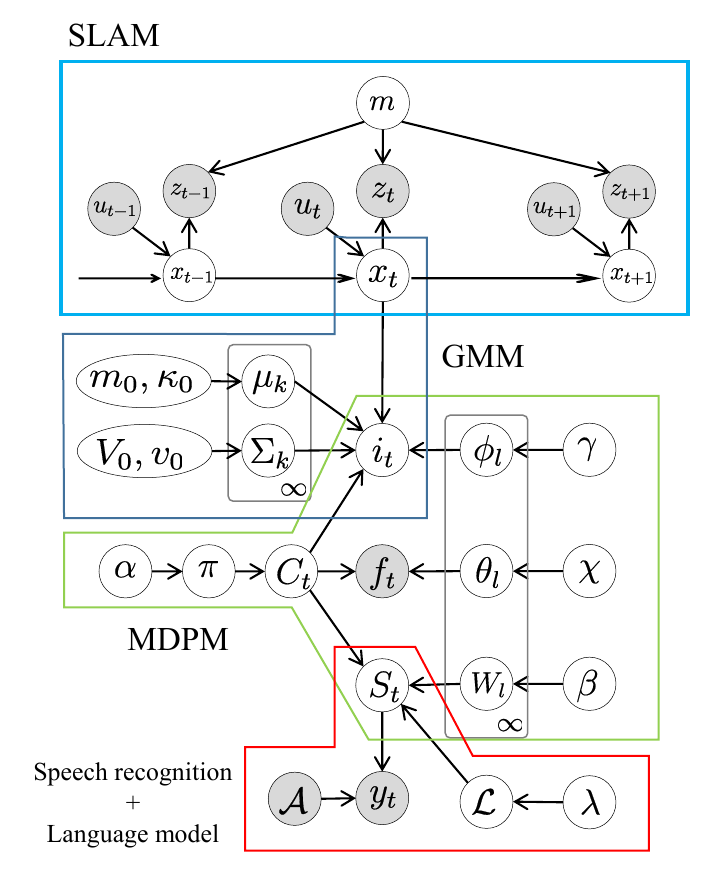}
  \end{minipage}
  \caption{Three types of connection of PGMs and their decomposition in SERKET framework (left). In the development phase, each PGM was developed in a distributed manner. In the inference phase, the modules work together by exchanging probabilistic information as an integrated cognitive system. For example, SpCoSLAM (Section~{\ref{subsec:spatial}}) can be decomposed into four elemental modules (right).}
  \label{fig:serket}
\end{figure}

Fig.~{\ref{fig:serket}} shows an overview of the SERKET framework. Generally, PGMs have three types of connections: (a) head-to-tail, (b) tail-to-tail, and (c) head-to-head). A complex PGM can be decomposed into two modules at the shared node (i.e., $z$ in Fig.~{\ref{fig:serket}}). 
In the inference phase, the internal variables in each elemental PGM can be updated independently, and the shared node can be updated by exchanging probabilistic information (i.e., posterior distributions conditioned by observations).
The right side of Fig.~{\ref{fig:serket}} shows the possible decomposition of SpCoSLAM (Section~{\ref{subsec:spatial}}). The total PGM of SpCoSLAM can be decomposed into SLAM, GMM, MDPM, and an automatic speech recognition system.
Communication—message passing—between elemental cognitive modules enables the whole cognitive model composed of many elemental modules, that is, the proposed WB-PGM composed through the SERKET framework, to be trained throughout the integrative system. 
For more details, please refer to the original papers~{\citep{nakamura2017serket,taniguchi2020neuro}}.

From the viewpoint of the practical implementation of the SERKET framework, there is the question of how to realize the communication protocol. Using a PPL or a library, for example, Pixyz~\citep{suzuki2021pixyz}, is a promising approach.
In addition, SERKET can reduce developmental efforts to create an integrative cognitive system. For example, {\cite{taniguchi2020neuro}} showed that an unsupervised machine learning system that categorizes \delspan{row}\addspan{raw} image data and speech signals simultaneously can be developed quickly and efficiently.
In particular, using the Neuro-SERKET framework, a complex learning system can be developed by connecting pre-existing modules (e.g., VAE, Gaussian mixture model, LDA, and automatic speech recognition system).

However, the possible integrative cognitive architecture has a huge degree of freedom to pick up elemental cognitive modules, connect them, and enable them to work and learn together. This is a design problem of cognitive architectures.
. Since the human brain is an excellent example of an integrative cognitive system that can work in a real-world environment and perform a wide range of complex tasks, learning the human brain architecture is a good approach to reduce the complexity of the design problem 

Thus, the WB-PGM should be interpreted from the viewpoint of the brain architecture. Such interpretability allows researchers to identify what is missing and what is implemented. It also facilitates communication between AI and robotics researchers and brain and cognitive science researchers.

\section{Current status of WB-PGM}\label{sec:5}
This section describes the current status of the development of WB-PGM including elemental modules and integration. Future issues are also discussed.

As mentioned in Section \ref{subsec:overviewOfDevelopment} and \ref{subsec:construct}, to create a PGM of the entire brain, it is crucial to construct a BIF of the entire brain, design the corresponding HCDs, and then run GIPA to create PGMs.
For hippocampal formation, data have been created for the PGM \citep{Taniguchi2021hpf-pgm}\footnote{
 https://wba-initiative.org/wiki/en/brain\_reference\_architecture, accessed: 2021-7-2.}.
For the interconnections between the neocortex, thalamus, and basal ganglia, we plan to proceed with BIF and HCD data creation based on the existing research results \citep{Yamakawa2020-xa}.
For eye movement, the claustrum, basal ganglia, and cerebellum, we are currently constructing a BIF.
To cover the entire brain, other brain regions, including the amygdala, midbrain, and pons, need to be designed as well.
Owing to the physical structure of BIF, it is expected to converge to a stable content once sufficient knowledge of neuroscience is accumulated.
Depending on the diversity of human cognitive functions, multiple HCDs will be described on a specific region of BIF data; thus, these HCDs should be integrated prior to creating PGMs as possible.

\subsection{Primitive structure in WB-PGM}\label{subsec:primitive}
The WB-PGM has a primitive function structure, similar to the basic layered structure of the cerebral cortex. 
As mentioned above, the WB-PGM uses a PGM as a primitive structure (i.e., a cognitive module), which corresponds to the ``circuit'' in the BIF.
From a computational viewpoint, we consider that PGMs can be classified into three generations.

The first generation is a PGM described in the probabilistic generative process, which is the most basic structure.
Model learning is a parameter estimation problem for probability distributions and is realized using Gibbs sampling or variational inference \citep{nakamura2007}. 

The second generation is a DPGM represented by VAE, which replaces the parameter estimation of the probability distributions with learning through neural networks. 
Learning is realized by \delspan{minimizing}\addspan{maximizing} the evidence lower bound using a neural network~\citep{kingma2013auto,suzuki2016joint}. 

As for the third generation, there is a structure that uses a self-attention mechanism and self-supervised learning. The self-attention mechanism has attracted attention owing to its extremely high performance in natural language processing (NLP) techniques, such as transformer, BERT, and GPT~\citep{vaswani2017attention,devlin2018bert,brown2020language}.
The application of BERT to multimodal expansion and RL is advancing \citep{miyazawa2020lambert}. In particular, multimodal BERT may become the basic module of the third generation. 
Contrastive learning, which is a representative approach of self-supervised learning, enables neural networks to perform representation learning without supervision or explicit definition of the generative process. This approach is used in a variety of tasks, including visual recognition and RL~{\citep{chen2020simple,srinivas2020curl,okada2020dreaming}}.

\delspan{As described in section~\ref{sec:3}, a wide range of cognitive modules have been developed for functions corresponding to brain regions. They can be used to develop an integrative cognitive architecture. The exploration of the selection and integration of modules is a future challenge.}

\delspan{Furthermore, the development of brain-inspired cognitive modules using GIPA is an important challenge. Although the HPF-PGM was developed and insightful PGMs were obtained following this approach, there is still considerable scope for exploration.}

\subsection{Connection of the building blocks} \label{subsubsec:building-block}
How is the overall structure constructed by combining primitives (i.e., cognitive modules)? \delspan{By combining them, a large-scale representation learning system corresponding to the cerebral cortex is realized. Modules for vision, speech, and language can be incorporated. }
\addspan{By integrating the modules, a path is opened toward the construction of a large-scale representation learning system that imitates the cerebral cortex. In fact, an integrated system is achieved by \cite{miyazawaFRONT2019}; however, it is not yet sufficiently large. Thus, modules for vision, speech, and language are incorporated into the system.
}
Furthermore, an RL module and a temporal module for long-term temporal planning are connected via the latent space held by the representation learning module. 
To connect building blocks, the SERKET framework \delspan{can be} \addspan{is} used as described in Section~{\ref{sec:4}}~{\citep{nakamura2017serket,taniguchi2020neuro}}.

\delspan{However, the connection of the modules is still a challenge.} \addspan{When connecting primitives,} time management poses a challenge in the course of connecting building blocks to configure the entire structure. To handle information flexibly, it is important to consider the physical characteristics of each signal and the time scale, according to how it is used. 
Since the brain processes information on a multi-scale and asynchronous basis, it is necessary to realize such an integration mechanism.
Recently, \cite{Taniguchi2021hpf-pgm} abstracted a neural activity phenomenon called phase precession into a mechanism called discrete-event queues when constructing the PGM of the HPF. This type of brain-inspired research could help us handle time in PGMs.

In the current prototype of WB-PGM {\citep{miyazawaFRONT2019}}, the latent variable for perception level (i.e., lower level) is integrated by a latent variable at the higher level. 
Furthermore, the model is a simple mechanism for executing long-term action plans by modeling the temporal relationship on latent variables at the higher level~\footnote{This type of model, in which latent variables of the higher-level layer are regarded as a downsampled versions of the lower-level ones, is well known as a ``coarse-to-fine'' model in generative models of images. }. 

\delspan{On the other hand, it is important to study the technology through which various modules process information at different time intervals and asynchronously/non-linearly and integrate them. 
One approach would be to incorporate the idea of an integrated execution platform such as BriCA, which factors the asynchronous nature of the brain, into the PGM. }

\delspan{Furthermore, the current Neuro-SERKET framework does not support the third-generation PGMs (i.e., cognitive modules based on self-supervised learning). Extending the neuro-SERKET to a framework that can integrate all types of PGMs is a future challenge.}

\subsection{Implementation}
\begin{figure*}[!t]
  \centering
  \includegraphics[width = \linewidth]{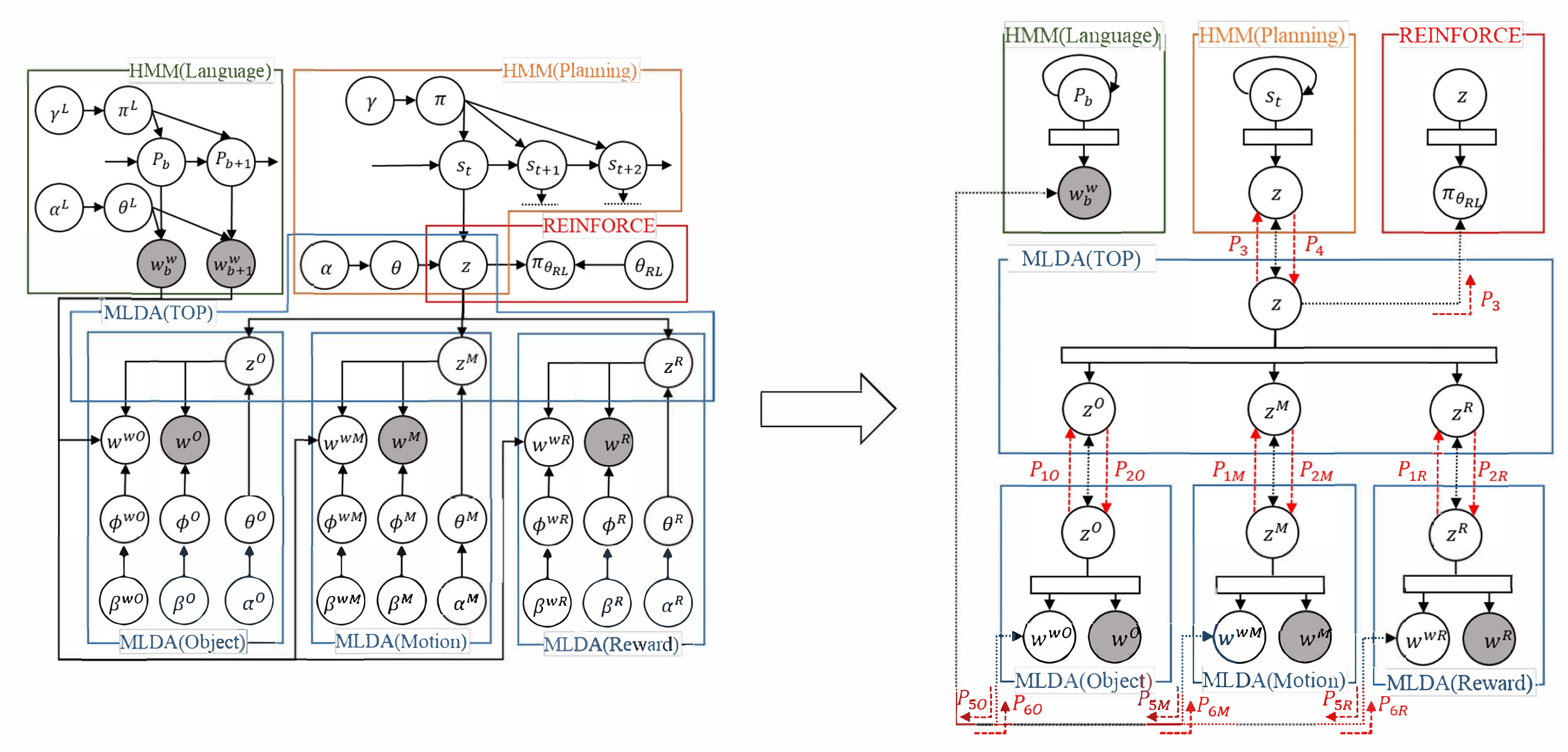}
  \caption{Our current prototype of the WB-PGM; probabilistic graphical model representation of the WB-PGM prototype (left), and SERKET implementation (right) \citep{miyazawaIROS2019}.}
  \label{fig:prototypeWBPGM}
\end{figure*}

Implementation of the WB-PGM is ongoing. 
{\cite{miyazawaFRONT2019}} integrated a representation learning module, a language learning module, and an RL module using Neuro-SERKET, and demonstrated the possibility of a real robot learning language and behavior simultaneously (Fig.~{\ref{fig:prototypeWBPGM}}). 
The example was developed using the Neuro-SERKET framework~{\citep{taniguchi2020neuro}}. 
As mentioned above, Neuro-SERKET can easily implement complex structures and optimize the entire structure by combining modules. 
\addspan{Note that (Neuro-)SERKET allows an integrated cognitive system to be trained as a whole throughout the system (as mentioned in Section~\ref{subsec:serket}). This is because SERKET decomposes the Bayesian inference of every latent variable of the integrated cognitive system into intra-module probabilistic inference ( i.e., training) and inter-module communication (i.e., message passing).}
In addition, it can incorporate an existing speech recognition module and deep learning models.

Despite the success of the implementation, we are at a preliminary stage of the development of the WB-PGM.
Ideally, the WB-PGM should involve whole-brain cognitive modules and functions; however, current examples of PGM-based integrative cognitive architecture only involve very limited brain functions.

When considering the realization of the WB-PGM, there are problems to be considered in relation to brain functions, such as the range of influence and scheduling of inference based on submodules and hierarchy. Thus, learning human brain architecture can be beneficial in solving these problems as well. 

\delspan{Once the PGM is implemented in this way, the activity reprehensibility can be assessed by evaluating the correspondence between neural activity data in a given circuit on the BIF and the behavior of the appropriate variables contained in software components at that location.
This is part of the ``fidelity assessment,'' which evaluates the biological plausibility of software by mapping it to BRA (see Section ).}

\subsection{Future perspectives}  
We have described the current status of the WB-PGM in the previous subsection. 
However, numerous aspects must be considered. \delspan{ as we have already addressed in each subsection of this section and throughout this paper. In this subsection, we point out two additional aspects.}

\addspan{As described in Section~{\ref{sec:3}}, a wide range of cognitive modules have been developed for functions corresponding to brain regions, which can be used to develop an integrative cognitive architecture. However, the exploration of the selection and integration of modules remains a future challenge. Furthermore, the current Neuro-SERKET framework does not support third-generation PGMs (i.e., cognitive modules based on self-supervised learning explained in Section~\ref{subsec:primitive}) . Thus, future research should extend neuro-SERKET to a framework that can integrate all types of PGMs.}

\addspan{Regarding time management discussed in Section~\ref{subsubsec:building-block}, it is important to study the technology through which various modules process information at different time intervals and asynchronously/non-linearly and integrate them. 
One approach would be to incorporate an integrated execution platform (e.g., BriCA) \citep{Takahashi2015-fd}, which factors the asynchronous nature of the brain, into PGMs. }

Trough fidelity evaluation (see \ref{subsec:eval}), the precision of the correspondence between the WB-PGM and BRA should be increased because the resolution of the correspondence between the current brain structure and WB-PGM is still coarse. 
It is crucial to consider the problem of dividing the area, such as how to manage time in the model and how to infer at a particular time. Further, considering missing elemental modules are also crucial.

\addspan{In the future, it is expected that BRA-driven development specialized for PGMs, will bring about refinements in the evaluation of the implemented brain-inspired software. The GIPA process was introduced to convert and build HCD into PGM, as shown in Figure 2 in Section 2.5. However, PGMs are not considered in fidelity evaluation, which estimates the conformance of the software to HCD. Therefore, it would be useful to add software to evaluate fidelity to PGM, PGM fidelity to HCD, and the fidelity of software implemented in PGM to HCD.}

It is also important to introduce a developmental perspective. For that purpose, it will be necessary to construct a physical body that can constantly work and learn and even grow. 
In this respect, the use of soft robots can be promising and may lead to realizing lifelong learning. 
This lead us to the suggestions on endowing robots with the following capabilities: 
\begin{enumerate}
\item learning of causality: generalizability, sample efficiency
\item emotions (maintaining its own body): 
from self/other to sociality
\item creativity: social value, imagination
\item explainability: communication
\item consciousness: global workspace, meta cognition, qualia 
\end{enumerate}

In our future work, we will develop the WB-PGM by addressing these issues

\section{Conclusions}\label{sec:6}
In this study, we proposed WB-PGM, an approach to develop an integrative cognitive architecture for developmental robots based on brain-inspired PGMs. PGMs and their inferences can learn knowledge from sensory-motor observations without manually crafted annotation/label data. Unlike most modern AI systems, biological cognitive systems, especially the human brain, can acquire a wide range of cognitive capabilities without supervision. We argue that a PGM-based approach is promising for the development of an integrative cognitive architecture. 
Although previous PGM-based integrative cognitive systems for developmental robots have been proposed, their cognitive capabilities were limited. Furthermore, elemental cognitive modules were introduced or discussed in relation to PGMs. Subsequently, we hypothetically described a prototype of integrative cognitive architecture. 

Building a WB-PGM has two advantages. First, it can serve as a reference for brain studies. The PGM describes explicit informational relationships between variables, that is, internal representations. This description provides interpretable guidance from computational sciences to brain science. By providing such information, researchers in neuroscience can provide feedback to researchers in AI and robotics on what the current models lack with reference to the brain. Our WB-PGM approach can facilitate discussion and collaboration among researchers in neuro-cognitive sciences as well as AI and robotics.


We admit that world models and the FEP are general and critical ideas for developing next-generation AI, which has to be integrative, autonomous, and developmental. However, current studies related to these theoretical ideas are mostly limited to simple problems and experiments in a simulation environment. However, to study cognition, it is crucial to test hypothetical ideas in a real-world environment. Developing robots and making them perform practical tasks in real-world environments are important processes for the exploration of next-generation AI. To this end, top-down engineering of cognitive architecture is required. A theory-based practical implementation of artificial cognitive systems in robots is crucial.


We argue that referring to the WBA, that is, the structure of the human brain, is beneficial for developing a cognitive architecture.
However, naturally, it may be argued that such brain structures should emerge from data learned by a large neural network. Current success in very large language models, for example, GPT-3 and BERT, seems to corroborate this idea.
We do not have an answer to this question; however, such a general approach is not feasible for realizing AGI. The WB-PGM is a more promising approach to develop a cognitive architecture for a developmental robot that admits the current technological background.
In addition, even if we can develop a meta-cognitive system that can facilitate the emergence of whole brain structures, the BRA and WB-PGM will be a good reference for evaluating the emerged structure.

The emphasis in this paper was on PGMs, which are basically directed graphs. Another type of Bayesian network is Markov networks, which are undirected graphs. We selected PGMs for two reasons. First, latent variable models based on PGMs, for example, LDA, HMM, and VAE, have been successful at modeling cognitive systems and exhibiting good properties from engineering and practical viewpoints. Second, in generative models, arrows represent predictions, and the characteristics fit the idea of predictive coding. However, this is not a strict argument. It is acceptable to partially introduce a Markov network into the proposed WB-PGM. 

In addition, it may be argued that the human brain has a recursive structure and bilateral connections. However, in a probabilistic graphical model representing a PGM, a directed arrow connecting two variables does not imply the absence of an inverse connection between two nodes, as the arrows represent the generative process and not the physical connections. In the inference process, an inverse information stream should be considered. This point is clearly underscored by amortized inferences connecting nodes inversely in inference networks.
In this sense, when we compare probabilistic graphical models and the neural connections in the human brain, it is necessary to differentiate generative processes from the inference ones.

\section*{Acknowledgements}

This research was partially supported by Grant-in-Aid for Scientific Research (B) (18H03308), for Scientific Research on Innovative Areas (16K21738, 16H06561, 16H06562, 16H06563, 16H06569, 16H06571) funded by the Ministry of Education, Culture, Sports, Science, and Technology, Japan and by CREST (JPMJCR15E3), JST
and by the Japan Ministry of Education, Culture, Sports, Science and Technology (KAKENHI Grant Number 17H06315, Grant-in-Aid for Scientific Research on Innovative Areas, Brain information dynamics underlying multi-area interconnectivity and parallel processing), and for a project, JPNP16007, subsidized by the New Energy and Industrial Technology Development Organization (NEDO).
We would like to thank Yuya Kobayashi for providing with the experimental results for Fig.~\ref{fig:scene_interpretation}.






\bibliographystyle{elsarticle-harv}\biboptions{authoryear}
\bibliography{main.bib}

\begin{thebibliography}{188}
\expandafter\ifx\csname natexlab\endcsname\relax\def\natexlab#1{#1}\fi
\providecommand{\url}[1]{\texttt{#1}}
\providecommand{\href}[2]{#2}
\providecommand{\path}[1]{#1}
\providecommand{\DOIprefix}{doi:}
\providecommand{\ArXivprefix}{arXiv:}
\providecommand{\URLprefix}{URL: }
\providecommand{\Pubmedprefix}{pmid:}
\providecommand{\doi}[1]{\href{http://dx.doi.org/#1}{\path{#1}}}
\providecommand{\Pubmed}[1]{\href{pmid:#1}{\path{#1}}}
\providecommand{\bibinfo}[2]{#2}
\ifx\xfnm\relax \def\xfnm[#1]{\unskip,\space#1}\fi
\bibitem[{Amari et~al.(2002)Amari, Beltrame, Bjaalie, Dalkara, De~Schutter,
  Egan, Goddard, Gonzalez, Grillner, Herz, Hoffmann, Jaaskelainen, Koslow, Lee,
  Matthiessen, Miller, Da~Silva, Novak, Ravindranath, Ritz, Ruotsalainen,
  Sebestra, Subramaniam, Tang, Toga, Usui, Van~Pelt, Verschure, Willshaw,
  Wrobel and {OECD Neuroinformatics Working Group}}]{Amari2002-rp}
\bibinfo{author}{Amari, S.I.}, \bibinfo{author}{Beltrame, F.},
  \bibinfo{author}{Bjaalie, J.G.}, \bibinfo{author}{Dalkara, T.},
  \bibinfo{author}{De~Schutter, E.}, \bibinfo{author}{Egan, G.F.},
  \bibinfo{author}{Goddard, N.H.}, \bibinfo{author}{Gonzalez, C.},
  \bibinfo{author}{Grillner, S.}, \bibinfo{author}{Herz, A.},
  \bibinfo{author}{Hoffmann, K.P.}, \bibinfo{author}{Jaaskelainen, I.},
  \bibinfo{author}{Koslow, S.H.}, \bibinfo{author}{Lee, S.Y.},
  \bibinfo{author}{Matthiessen, L.}, \bibinfo{author}{Miller, P.L.},
  \bibinfo{author}{Da~Silva, F.M.}, \bibinfo{author}{Novak, M.},
  \bibinfo{author}{Ravindranath, V.}, \bibinfo{author}{Ritz, R.},
  \bibinfo{author}{Ruotsalainen, U.}, \bibinfo{author}{Sebestra, V.},
  \bibinfo{author}{Subramaniam, S.}, \bibinfo{author}{Tang, Y.},
  \bibinfo{author}{Toga, A.W.}, \bibinfo{author}{Usui, S.},
  \bibinfo{author}{Van~Pelt, J.}, \bibinfo{author}{Verschure, P.},
  \bibinfo{author}{Willshaw, D.}, \bibinfo{author}{Wrobel, A.},
  \bibinfo{author}{{OECD Neuroinformatics Working Group}},
  \bibinfo{year}{2002}.
\newblock \bibinfo{title}{Neuroinformatics: the integration of shared databases
  and tools towards integrative neuroscience}.
\newblock \bibinfo{journal}{J. Integr. Neurosci.} \bibinfo{volume}{1},
  \bibinfo{pages}{117--128}.
\bibitem[{Anderson(2009)}]{anderson2009can}
\bibinfo{author}{Anderson, J.R.}, \bibinfo{year}{2009}.
\newblock \bibinfo{title}{How can the human mind occur in the physical
  universe?}
\newblock \bibinfo{publisher}{Oxford University Press}.
\bibitem[{Arakawa and Yamakawa(2020)}]{Arakawa2020-ep}
\bibinfo{author}{Arakawa, N.}, \bibinfo{author}{Yamakawa, H.},
  \bibinfo{year}{2020}.
\newblock \bibinfo{title}{The brain information flow format}, in:
  \bibinfo{booktitle}{The 1st {Asia-Pacific} Computational and Cognitive
  Neuroscience ({AP-CCN}) Conference}, p. \bibinfo{pages}{0029}.
\bibitem[{{Araki} et~al.(2013){Araki}, {Nakamura} and {Nagai}}]{araki2013}
\bibinfo{author}{{Araki}, T.}, \bibinfo{author}{{Nakamura}, T.},
  \bibinfo{author}{{Nagai}, T.}, \bibinfo{year}{2013}.
\newblock \bibinfo{title}{Long-term learning of concept and word by robots:
  Interactive learning framework and preliminary results}, in:
  \bibinfo{booktitle}{2013 IEEE/RSJ International Conference on Intelligent
  Robots and Systems}, pp. \bibinfo{pages}{2280--2287}.
\newblock \DOIprefix\doi{10.1109/IROS.2013.6696675}.
\bibitem[{Bacon et~al.()Bacon, Harb and Precup}]{Bacon2017}
\bibinfo{author}{Bacon, P.L.}, \bibinfo{author}{Harb, J.},
  \bibinfo{author}{Precup, D.}, .
\newblock \bibinfo{title}{The option-critic architecture}, in:
  \bibinfo{booktitle}{Proceedings of the Thirty-First AAAI Conference on
  Artificial Intelligence (AAAI-17)}.
\bibitem[{Ball et~al.(2013)Ball, Heath, Wiles, Wyeth, Corke and
  Milford}]{Ball2013-hi}
\bibinfo{author}{Ball, D.}, \bibinfo{author}{Heath, S.},
  \bibinfo{author}{Wiles, J.}, \bibinfo{author}{Wyeth, G.},
  \bibinfo{author}{Corke, P.}, \bibinfo{author}{Milford, M.},
  \bibinfo{year}{2013}.
\newblock \bibinfo{title}{{OpenRatSLAM}: an open source brain-based {SLAM}
  system}.
\newblock \bibinfo{journal}{Auton. Robots} \bibinfo{volume}{34},
  \bibinfo{pages}{149--176}.
\bibitem[{Balleine et~al.(2015)Balleine, Dezfouli, Ito and Doya}]{Balleine2015}
\bibinfo{author}{Balleine, B.W.}, \bibinfo{author}{Dezfouli, A.},
  \bibinfo{author}{Ito, M.}, \bibinfo{author}{Doya, K.}, \bibinfo{year}{2015}.
\newblock \bibinfo{title}{Hierarchical control of goal-directed action in the
  cortical–basal ganglia network}.
\newblock \bibinfo{journal}{Current Opinion in Behavioral Sciences}
  \bibinfo{volume}{5}, \bibinfo{pages}{1--7}.
\newblock \DOIprefix\doi{10.1016/j.cobeha.2015.06.001}.
\bibitem[{Bao et~al.(2001)Bao, Chan and Merzenich}]{Bao2001}
\bibinfo{author}{Bao, S.}, \bibinfo{author}{Chan, V.T.},
  \bibinfo{author}{Merzenich, M.M.}, \bibinfo{year}{2001}.
\newblock \bibinfo{title}{Cortical remodelling induced by activity of ventral
  tegmental dopamine neurons}.
\newblock \bibinfo{journal}{Nature} \bibinfo{volume}{412},
  \bibinfo{pages}{79--83}.
\newblock \DOIprefix\doi{10.1038/35083586}.
\bibitem[{Barto(1995)}]{Barto1995}
\bibinfo{author}{Barto, A.G.}, \bibinfo{year}{1995}.
\newblock \bibinfo{title}{Adaptive critics and the basal ganglia}.
  \bibinfo{address}{Cambridge, MA, USA}.
\newblock pp. \bibinfo{pages}{215--232}.
\bibitem[{Bastos et~al.(2012)Bastos, Usrey, Adams, Mangun, Fries and
  Friston}]{Bastos2012-wv}
\bibinfo{author}{Bastos, A.M.}, \bibinfo{author}{Usrey, W.M.},
  \bibinfo{author}{Adams, R.A.}, \bibinfo{author}{Mangun, G.R.},
  \bibinfo{author}{Fries, P.}, \bibinfo{author}{Friston, K.J.},
  \bibinfo{year}{2012}.
\newblock \bibinfo{title}{Canonical microcircuits for predictive coding}.
\newblock \bibinfo{journal}{Neuron} \bibinfo{volume}{76},
  \bibinfo{pages}{695--711}.
\bibitem[{Bell(2004)}]{Bell2004-xg}
\bibinfo{author}{Bell, D.}, \bibinfo{year}{2004}.
\newblock \bibinfo{title}{The component diagram}.
\newblock
  \bibinfo{howpublished}{\url{https://developer.ibm.com/articles/the-component-diagram/}}.
\newblock \bibinfo{note}{Accessed: 2021-1-21}.
\bibitem[{Bellman(1952)}]{Bellman1952}
\bibinfo{author}{Bellman, R.}, \bibinfo{year}{1952}.
\newblock \bibinfo{title}{On the theory of dynamic programming}.
\newblock \bibinfo{journal}{Proceedings of the National Academy of Science,
  USA} \bibinfo{volume}{38}, \bibinfo{pages}{716--719}.
\bibitem[{Bengio et~al.(2013)Bengio, Courville and
  Vincent}]{bengio2013representation}
\bibinfo{author}{Bengio, Y.}, \bibinfo{author}{Courville, A.},
  \bibinfo{author}{Vincent, P.}, \bibinfo{year}{2013}.
\newblock \bibinfo{title}{Representation learning: A review and new
  perspectives}.
\newblock \bibinfo{journal}{IEEE transactions on pattern analysis and machine
  intelligence} \bibinfo{volume}{35}, \bibinfo{pages}{1798--1828}.
\bibitem[{Beul and Hilgetag(2014)}]{Beul2014-kj}
\bibinfo{author}{Beul, S.F.}, \bibinfo{author}{Hilgetag, C.C.},
  \bibinfo{year}{2014}.
\newblock \bibinfo{title}{Towards a ``canonical'' agranular cortical
  microcircuit}.
\newblock \bibinfo{journal}{Front. Neuroanat.} \bibinfo{volume}{8},
  \bibinfo{pages}{165}.
\bibitem[{Bingham et~al.(2019)Bingham, Chen, Jankowiak, Obermeyer, Pradhan,
  Karaletsos, Singh, Szerlip, Horsfall and Goodman}]{bingham2019pyro}
\bibinfo{author}{Bingham, E.}, \bibinfo{author}{Chen, J.P.},
  \bibinfo{author}{Jankowiak, M.}, \bibinfo{author}{Obermeyer, F.},
  \bibinfo{author}{Pradhan, N.}, \bibinfo{author}{Karaletsos, T.},
  \bibinfo{author}{Singh, R.}, \bibinfo{author}{Szerlip, P.},
  \bibinfo{author}{Horsfall, P.}, \bibinfo{author}{Goodman, N.D.},
  \bibinfo{year}{2019}.
\newblock \bibinfo{title}{Pyro: Deep universal probabilistic programming}.
\newblock \bibinfo{journal}{The Journal of Machine Learning Research}
  \bibinfo{volume}{20}, \bibinfo{pages}{973--978}.
\bibitem[{Bishop(2006)}]{bishop2006pattern}
\bibinfo{author}{Bishop, C.M.}, \bibinfo{year}{2006}.
\newblock \bibinfo{title}{Pattern recognition and machine learning}.
\newblock \bibinfo{publisher}{springer}.
\bibitem[{Bisk and Hockenmaier(2013)}]{bisk2013hdp}
\bibinfo{author}{Bisk, Y.}, \bibinfo{author}{Hockenmaier, J.},
  \bibinfo{year}{2013}.
\newblock \bibinfo{title}{An hdp model for inducing combinatory categorial
  grammars}.
\newblock \bibinfo{journal}{Transactions of the Association for Computational
  Linguistics} \bibinfo{volume}{1}, \bibinfo{pages}{75--88}.
\bibitem[{Blei et~al.(2003)Blei, Ng and Jordan}]{Blei:2003:LDA:944919.944937}
\bibinfo{author}{Blei, D.M.}, \bibinfo{author}{Ng, A.Y.},
  \bibinfo{author}{Jordan, M.I.}, \bibinfo{year}{2003}.
\newblock \bibinfo{title}{{Latent Dirichlet Allocation}}.
\newblock \bibinfo{journal}{Journal of Machine Learning Research}
  \bibinfo{volume}{3}, \bibinfo{pages}{993--1022}.
\bibitem[{Botvinick and Toussaint(2012)}]{Botvinick2012}
\bibinfo{author}{Botvinick, M.}, \bibinfo{author}{Toussaint, M.},
  \bibinfo{year}{2012}.
\newblock \bibinfo{title}{Planning as inference}.
\newblock \bibinfo{journal}{Trends Cogn Sci} \bibinfo{volume}{16},
  \bibinfo{pages}{485--8}.
\newblock \DOIprefix\doi{10.1016/j.tics.2012.08.006}.
\bibitem[{Brown et~al.(2020)Brown, Mann, Ryder, Subbiah, Kaplan, Dhariwal,
  Neelakantan, Shyam, Sastry, Askell et~al.}]{brown2020language}
\bibinfo{author}{Brown, T.B.}, \bibinfo{author}{Mann, B.},
  \bibinfo{author}{Ryder, N.}, \bibinfo{author}{Subbiah, M.},
  \bibinfo{author}{Kaplan, J.}, \bibinfo{author}{Dhariwal, P.},
  \bibinfo{author}{Neelakantan, A.}, \bibinfo{author}{Shyam, P.},
  \bibinfo{author}{Sastry, G.}, \bibinfo{author}{Askell, A.}, et~al.,
  \bibinfo{year}{2020}.
\newblock \bibinfo{title}{Language models are few-shot learners}.
\newblock \bibinfo{journal}{arXiv preprint arXiv:2005.14165} .
\bibitem[{Burgess et~al.(2019)Burgess, Matthey, Watters, Kabra, Higgins,
  Botvinick and Lerchner}]{burgess2019monet}
\bibinfo{author}{Burgess, C.P.}, \bibinfo{author}{Matthey, L.},
  \bibinfo{author}{Watters, N.}, \bibinfo{author}{Kabra, R.},
  \bibinfo{author}{Higgins, I.}, \bibinfo{author}{Botvinick, M.},
  \bibinfo{author}{Lerchner, A.}, \bibinfo{year}{2019}.
\newblock \bibinfo{title}{Monet: Unsupervised scene decomposition and
  representation}.
\newblock \bibinfo{journal}{arXiv preprint arXiv:1901.11390} .
\bibitem[{Cassell et~al.(1999)Cassell, Freedman and Shi}]{Cassel1999}
\bibinfo{author}{Cassell, M.D.}, \bibinfo{author}{Freedman, L.J.},
  \bibinfo{author}{Shi, C.}, \bibinfo{year}{1999}.
\newblock \bibinfo{title}{The intrinsic organization of the central extended
  amygdala}.
\newblock \bibinfo{journal}{Annals of New York Academy of Sciences} .
\bibitem[{Chen et~al.(2020)Chen, Kornblith, Norouzi and
  Hinton}]{chen2020simple}
\bibinfo{author}{Chen, T.}, \bibinfo{author}{Kornblith, S.},
  \bibinfo{author}{Norouzi, M.}, \bibinfo{author}{Hinton, G.},
  \bibinfo{year}{2020}.
\newblock \bibinfo{title}{A simple framework for contrastive learning of visual
  representations}, in: \bibinfo{booktitle}{International conference on machine
  learning}, \bibinfo{organization}{PMLR}. pp. \bibinfo{pages}{1597--1607}.
\bibitem[{Constantinidis and Klingberg(2016)}]{constantinidis2016neuroscience}
\bibinfo{author}{Constantinidis, C.}, \bibinfo{author}{Klingberg, T.},
  \bibinfo{year}{2016}.
\newblock \bibinfo{title}{The neuroscience of working memory capacity and
  training}.
\newblock \bibinfo{journal}{Nature Reviews Neuroscience} \bibinfo{volume}{17},
  \bibinfo{pages}{438}.
\bibitem[{Crasto(2007)}]{Crasto2007-yb}
\bibinfo{author}{Crasto, C.J.}, \bibinfo{year}{2007}.
\newblock \bibinfo{title}{Neuroinformatics}.
\newblock \bibinfo{publisher}{Springer Science \& Business Media}.
\bibitem[{Daw et~al.(2005)Daw, Niv and Dayan}]{Daw2005}
\bibinfo{author}{Daw, N.D.}, \bibinfo{author}{Niv, Y.}, \bibinfo{author}{Dayan,
  P.}, \bibinfo{year}{2005}.
\newblock \bibinfo{title}{Uncertainty-based competition between prefrontal and
  dorsolateral striatal systems for behavioral control}.
\newblock \bibinfo{journal}{Nature Neuroscience} \bibinfo{volume}{8},
  \bibinfo{pages}{1704--11}.
\newblock \DOIprefix\doi{10.1038/nn1560}.
\bibitem[{Deacon(1998)}]{deacon1998symbolic}
\bibinfo{author}{Deacon, T.W.}, \bibinfo{year}{1998}.
\newblock \bibinfo{title}{The symbolic species: The co-evolution of language
  and the brain}.
\newblock \bibinfo{number}{202}, \bibinfo{publisher}{WW Norton \& Company}.
\bibitem[{Devlin et~al.(2018)Devlin, Chang, Lee and Toutanova}]{devlin2018bert}
\bibinfo{author}{Devlin, J.}, \bibinfo{author}{Chang, M.W.},
  \bibinfo{author}{Lee, K.}, \bibinfo{author}{Toutanova, K.},
  \bibinfo{year}{2018}.
\newblock \bibinfo{title}{Bert: Pre-training of deep bidirectional transformers
  for language understanding}.
\newblock \bibinfo{journal}{arXiv preprint arXiv:1810.04805} .
\bibitem[{Doncieux et~al.(2020)Doncieux, Bredeche, Goff, Girard, Coninx,
  Sigaud, Khamassi, D{\'{i}}az-Rodr{\'{i}}guez, Filliat, Hospedales, Eiben and
  Duro}]{Doncieux2020}
\bibinfo{author}{Doncieux, S.}, \bibinfo{author}{Bredeche, N.},
  \bibinfo{author}{Goff, L.L.}, \bibinfo{author}{Girard, B.},
  \bibinfo{author}{Coninx, A.}, \bibinfo{author}{Sigaud, O.},
  \bibinfo{author}{Khamassi, M.}, \bibinfo{author}{D{\'{i}}az-Rodr{\'{i}}guez,
  N.}, \bibinfo{author}{Filliat, D.}, \bibinfo{author}{Hospedales, T.},
  \bibinfo{author}{Eiben, A.}, \bibinfo{author}{Duro, R.},
  \bibinfo{year}{2020}.
\newblock \bibinfo{title}{{DREAM Architecture: a Developmental Approach to
  Open-Ended Learning in Robotics}} , \bibinfo{pages}{1--29}\URLprefix
  \url{http://arxiv.org/abs/2005.06223},
  \href{http://arxiv.org/abs/2005.06223}{{\tt arXiv:2005.06223}}.
\bibitem[{Doucet et~al.(2000)Doucet, Freitas, Murphy and
  Russell}]{doucet2000rao}
\bibinfo{author}{Doucet, A.}, \bibinfo{author}{Freitas, N.d.},
  \bibinfo{author}{Murphy, K.P.}, \bibinfo{author}{Russell, S.J.},
  \bibinfo{year}{2000}.
\newblock \bibinfo{title}{Rao-blackwellised particle filtering for dynamic
  bayesian networks}, in: \bibinfo{booktitle}{Proceedings of the 16th
  Conference on Uncertainty in Artificial Intelligence}, pp.
  \bibinfo{pages}{176--183}.
\bibitem[{Douglas et~al.(1989)Douglas, Martin and Whitteridge}]{Douglas1989-ql}
\bibinfo{author}{Douglas, R.J.}, \bibinfo{author}{Martin, K.A.C.},
  \bibinfo{author}{Whitteridge, D.}, \bibinfo{year}{1989}.
\newblock \bibinfo{title}{A canonical microcircuit for neocortex}.
\newblock \bibinfo{journal}{Neural Comput.} \bibinfo{volume}{1},
  \bibinfo{pages}{480--488}.
\bibitem[{Doya(1999)}]{doya1999}
\bibinfo{author}{Doya, K.}, \bibinfo{year}{1999}.
\newblock \bibinfo{title}{What are the computations of the cerebellum, the
  basal ganglia and the cerebral cortex?}
\newblock \bibinfo{journal}{Neural networks} \bibinfo{volume}{12},
  \bibinfo{pages}{961--974}.
\bibitem[{Doya(2000)}]{Doya2000}
\bibinfo{author}{Doya, K.}, \bibinfo{year}{2000}.
\newblock \bibinfo{title}{Complementary roles of basal ganglia and cerebellum
  in learning and motor control}.
\newblock \bibinfo{journal}{Current Opinion in Neurobiology}
  \bibinfo{volume}{10}, \bibinfo{pages}{732--739}.
\bibitem[{Doya(2007)}]{Doya2007}
\bibinfo{author}{Doya, K.}, \bibinfo{year}{2007}.
\newblock \bibinfo{title}{Reinforcement learning: Computational theory and
  biological mechanisms}.
\newblock \bibinfo{journal}{HFSP J} \bibinfo{volume}{1},
  \bibinfo{pages}{30--40}.
\newblock \DOIprefix\doi{10.2976/1.2732246/10.2976/1}.
\bibitem[{Doya(2021)}]{Doya2021-dk}
\bibinfo{author}{Doya, K.}, \bibinfo{year}{2021}.
\newblock \bibinfo{title}{Canonical cortical circuits and the duality of
  bayesian inference and optimal control}
  \href{http://arxiv.org/abs/2106.02785}{{\tt arXiv:2106.02785}}.
\bibitem[{Doya et~al.(2007)Doya, Ishii, Pouget and Rao}]{Doya2007-of}
\bibinfo{author}{Doya, K.}, \bibinfo{author}{Ishii, S.},
  \bibinfo{author}{Pouget, A.}, \bibinfo{author}{Rao, R.P.N.},
  \bibinfo{year}{2007}.
\newblock \bibinfo{title}{Bayesian Brain: Probabilistic Approaches to Neural
  Coding}.
\newblock \bibinfo{publisher}{MIT Press}.
\bibitem[{{El Hafi} et~al.(2020){El Hafi}, Isobe, Tabuchi, Katsumata, Nakamura,
  Fukui, Matsuo, Ricardez, Yamamoto, Taniguchi, Hagiwara and
  Taniguchi}]{lotfi2018WRS}
\bibinfo{author}{{El Hafi}, L.}, \bibinfo{author}{Isobe, S.},
  \bibinfo{author}{Tabuchi, Y.}, \bibinfo{author}{Katsumata, Y.},
  \bibinfo{author}{Nakamura, H.}, \bibinfo{author}{Fukui, T.},
  \bibinfo{author}{Matsuo, T.}, \bibinfo{author}{Ricardez, G.A.G.},
  \bibinfo{author}{Yamamoto, M.}, \bibinfo{author}{Taniguchi, A.},
  \bibinfo{author}{Hagiwara, Y.}, \bibinfo{author}{Taniguchi, T.},
  \bibinfo{year}{2020}.
\newblock \bibinfo{title}{{System for augmented human-robot interaction through
  mixed reality and robot training by non-experts in customer service
  environments}}.
\newblock \bibinfo{journal}{Advanced Robotics} \bibinfo{volume}{34},
  \bibinfo{pages}{157--172}.
\newblock \URLprefix \url{https://doi.org/10.1080/01691864.2019.1694068},
  \DOIprefix\doi{10.1080/01691864.2019.1694068}.
\bibitem[{Eliasmith et~al.(2012)Eliasmith, Stewart, Choo, Bekolay, DeWolf, Tang
  and Rasmussen}]{Eliasmith1202}
\bibinfo{author}{Eliasmith, C.}, \bibinfo{author}{Stewart, T.C.},
  \bibinfo{author}{Choo, X.}, \bibinfo{author}{Bekolay, T.},
  \bibinfo{author}{DeWolf, T.}, \bibinfo{author}{Tang, Y.},
  \bibinfo{author}{Rasmussen, D.}, \bibinfo{year}{2012}.
\newblock \bibinfo{title}{A large-scale model of the functioning brain}.
\newblock \bibinfo{journal}{Science} \bibinfo{volume}{338},
  \bibinfo{pages}{1202--1205}.
\newblock \URLprefix
  \url{https://science.sciencemag.org/content/338/6111/1202},
  \DOIprefix\doi{10.1126/science.1225266},
  \href{http://arxiv.org/abs/https://science.sciencemag.org/content/338/6111/1202.full.pdf}{{\tt
  arXiv:https://science.sciencemag.org/content/338/6111/1202.full.pdf}}.
\bibitem[{Engelcke et~al.(2019)Engelcke, Kosiorek, Jones and
  Posner}]{engelcke2019genesis}
\bibinfo{author}{Engelcke, M.}, \bibinfo{author}{Kosiorek, A.R.},
  \bibinfo{author}{Jones, O.P.}, \bibinfo{author}{Posner, I.},
  \bibinfo{year}{2019}.
\newblock \bibinfo{title}{Genesis: Generative scene inference and sampling with
  object-centric latent representations}.
\newblock \bibinfo{journal}{arXiv preprint arXiv:1907.13052} .
\bibitem[{Eslami et~al.(2016)Eslami, Heess, Weber, Tassa, Szepesvari,
  Kavukcuoglu and Hinton}]{eslami2016attend}
\bibinfo{author}{Eslami, S.}, \bibinfo{author}{Heess, N.},
  \bibinfo{author}{Weber, T.}, \bibinfo{author}{Tassa, Y.},
  \bibinfo{author}{Szepesvari, D.}, \bibinfo{author}{Kavukcuoglu, K.},
  \bibinfo{author}{Hinton, G.E.}, \bibinfo{year}{2016}.
\newblock \bibinfo{title}{Attend, infer, repeat: Fast scene understanding with
  generative models}.
\newblock \bibinfo{journal}{arXiv preprint arXiv:1603.08575} .
\bibitem[{Fadlil et~al.(2013)Fadlil, Ikeda, Abe, Nakamura and Nagai}]{mmlda}
\bibinfo{author}{Fadlil, M.}, \bibinfo{author}{Ikeda, K.},
  \bibinfo{author}{Abe, K.}, \bibinfo{author}{Nakamura, T.},
  \bibinfo{author}{Nagai, T.}, \bibinfo{year}{2013}.
\newblock \bibinfo{title}{Integrated concept of objects and human motions based
  on multi-layered multimodal lda}, in: \bibinfo{booktitle}{IEEE/RSJ
  International Conference on Intelligent Robots and Systems},
  \bibinfo{organization}{IEEE}. pp. \bibinfo{pages}{2256--2263}.
\bibitem[{Fermin et~al.(2016a)Fermin, Sakagami, Kiyonari, Li, Matsumoto and
  Yamagishi}]{fermin2016representation}
\bibinfo{author}{Fermin, A.S.}, \bibinfo{author}{Sakagami, M.},
  \bibinfo{author}{Kiyonari, T.}, \bibinfo{author}{Li, Y.},
  \bibinfo{author}{Matsumoto, Y.}, \bibinfo{author}{Yamagishi, T.},
  \bibinfo{year}{2016}a.
\newblock \bibinfo{title}{Representation of economic preferences in the
  structure and function of the amygdala and prefrontal cortex}.
\newblock \bibinfo{journal}{Scientific reports} \bibinfo{volume}{6},
  \bibinfo{pages}{1--11}.
\bibitem[{Fermin et~al.(2016b)Fermin, Yoshida, Yoshimoto, Ito, Tanaka and
  Doya}]{Fermin2016}
\bibinfo{author}{Fermin, A.S.}, \bibinfo{author}{Yoshida, T.},
  \bibinfo{author}{Yoshimoto, J.}, \bibinfo{author}{Ito, M.},
  \bibinfo{author}{Tanaka, S.C.}, \bibinfo{author}{Doya, K.},
  \bibinfo{year}{2016}b.
\newblock \bibinfo{title}{Model-based action planning involves
  cortico-cerebellar and basal ganglia networks}.
\newblock \bibinfo{journal}{Sci Rep} \bibinfo{volume}{6},
  \bibinfo{pages}{31378}.
\newblock \DOIprefix\doi{10.1038/srep31378}.
\bibitem[{Friston(2005)}]{friston2005theory}
\bibinfo{author}{Friston, K.}, \bibinfo{year}{2005}.
\newblock \bibinfo{title}{A theory of cortical responses}.
\newblock \bibinfo{journal}{Philosophical transactions of the Royal Society B:
  Biological sciences} \bibinfo{volume}{360}, \bibinfo{pages}{815--836}.
\bibitem[{Friston(2012)}]{Friston2012-cz}
\bibinfo{author}{Friston, K.}, \bibinfo{year}{2012}.
\newblock \bibinfo{title}{The history of the future of the bayesian brain}.
\newblock \bibinfo{journal}{Neuroimage} \bibinfo{volume}{62},
  \bibinfo{pages}{1230--1233}.
\bibitem[{Friston(2019)}]{friston2019free}
\bibinfo{author}{Friston, K.}, \bibinfo{year}{2019}.
\newblock \bibinfo{title}{A free energy principle for a particular physics}.
\newblock \bibinfo{journal}{arXiv preprint arXiv:1906.10184} .
\bibitem[{Friston et~al.(2006)Friston, Kilner and Harrison}]{friston2006free}
\bibinfo{author}{Friston, K.}, \bibinfo{author}{Kilner, J.},
  \bibinfo{author}{Harrison, L.}, \bibinfo{year}{2006}.
\newblock \bibinfo{title}{A free energy principle for the brain}.
\newblock \bibinfo{journal}{Journal of Physiology-Paris} \bibinfo{volume}{100},
  \bibinfo{pages}{70--87}.
\bibitem[{Friston et~al.(2015)Friston, Rigoli, Ognibene, Mathys, Fitzgerald and
  Pezzulo}]{friston2015active}
\bibinfo{author}{Friston, K.}, \bibinfo{author}{Rigoli, F.},
  \bibinfo{author}{Ognibene, D.}, \bibinfo{author}{Mathys, C.},
  \bibinfo{author}{Fitzgerald, T.}, \bibinfo{author}{Pezzulo, G.},
  \bibinfo{year}{2015}.
\newblock \bibinfo{title}{Active inference and epistemic value}.
\newblock \bibinfo{journal}{Cognitive neuroscience} \bibinfo{volume}{6},
  \bibinfo{pages}{187--214}.
\bibitem[{Friston et~al.(2010)Friston, Daunizeau, Kilner and
  Kiebel}]{friston2010action}
\bibinfo{author}{Friston, K.J.}, \bibinfo{author}{Daunizeau, J.},
  \bibinfo{author}{Kilner, J.}, \bibinfo{author}{Kiebel, S.J.},
  \bibinfo{year}{2010}.
\newblock \bibinfo{title}{Action and behavior: a free-energy formulation}.
\newblock \bibinfo{journal}{Biological cybernetics} \bibinfo{volume}{102},
  \bibinfo{pages}{227--260}.
\bibitem[{Fukawa et~al.(2020)Fukawa, Aizawa, Yamakawa and
  Yairi}]{Fukawa2020-hl}
\bibinfo{author}{Fukawa, A.}, \bibinfo{author}{Aizawa, T.},
  \bibinfo{author}{Yamakawa, H.}, \bibinfo{author}{Yairi, I.E.},
  \bibinfo{year}{2020}.
\newblock \bibinfo{title}{Identifying core regions for path integration on
  medial entorhinal cortex of hippocampal formation}.
\newblock \bibinfo{journal}{Brain Sci} \bibinfo{volume}{10}.
\bibitem[{Galantucci et~al.(2006)Galantucci, Fowler and
  Turvey}]{galantucci2006motor}
\bibinfo{author}{Galantucci, B.}, \bibinfo{author}{Fowler, C.A.},
  \bibinfo{author}{Turvey, M.T.}, \bibinfo{year}{2006}.
\newblock \bibinfo{title}{The motor theory of speech perception reviewed}.
\newblock \bibinfo{journal}{Psychonomic bulletin \& review}
  \bibinfo{volume}{13}, \bibinfo{pages}{361--377}.
\bibitem[{Gershman and Goodman(2014)}]{Gershman2014}
\bibinfo{author}{Gershman, S.J.}, \bibinfo{author}{Goodman, N.D.},
  \bibinfo{year}{2014}.
\newblock \bibinfo{title}{{Amortized Inference in Probabilistic Reasoning}},
  in: \bibinfo{booktitle}{Proceedings of the Annual Conference of the Cognitive
  Science Society (CogSci)}, pp. \bibinfo{pages}{517--522}.
\bibitem[{Goertzel et~al.(2010)Goertzel, Lian, Arel, de~Garis and
  Chen}]{Goertzel2010-hy}
\bibinfo{author}{Goertzel, B.}, \bibinfo{author}{Lian, R.},
  \bibinfo{author}{Arel, I.}, \bibinfo{author}{de~Garis, H.},
  \bibinfo{author}{Chen, S.}, \bibinfo{year}{2010}.
\newblock \bibinfo{title}{A world survey of artificial brain projects, part
  {II}: Biologically inspired cognitive architectures}.
\newblock \bibinfo{journal}{Neurocomputing} \bibinfo{volume}{74},
  \bibinfo{pages}{30--49}.
\bibitem[{Goodale and Milner(1992)}]{goodale1992separate}
\bibinfo{author}{Goodale, M.A.}, \bibinfo{author}{Milner, A.D.},
  \bibinfo{year}{1992}.
\newblock \bibinfo{title}{Separate visual pathways for perception and action}.
\newblock \bibinfo{journal}{Trends in neurosciences} \bibinfo{volume}{15},
  \bibinfo{pages}{20--25}.
\bibitem[{Goodfellow et~al.(2016)Goodfellow, Bengio, Courville and
  Bengio}]{goodfellow2016deep}
\bibinfo{author}{Goodfellow, I.}, \bibinfo{author}{Bengio, Y.},
  \bibinfo{author}{Courville, A.}, \bibinfo{author}{Bengio, Y.},
  \bibinfo{year}{2016}.
\newblock \bibinfo{title}{Deep learning}. volume~\bibinfo{volume}{1}.
\newblock \bibinfo{publisher}{MIT press Cambridge}.
\bibitem[{Goodfellow et~al.(2014)Goodfellow, Pouget-Abadie, Mirza, Xu,
  Warde-Farley, Ozair, Courville and Bengio}]{goodfellow2014generative}
\bibinfo{author}{Goodfellow, I.J.}, \bibinfo{author}{Pouget-Abadie, J.},
  \bibinfo{author}{Mirza, M.}, \bibinfo{author}{Xu, B.},
  \bibinfo{author}{Warde-Farley, D.}, \bibinfo{author}{Ozair, S.},
  \bibinfo{author}{Courville, A.}, \bibinfo{author}{Bengio, Y.},
  \bibinfo{year}{2014}.
\newblock \bibinfo{title}{Generative adversarial networks}.
\newblock \bibinfo{journal}{arXiv preprint arXiv:1406.2661} .
\bibitem[{Goodman et~al.(2012)Goodman, Mansinghka, Roy, Bonawitz and
  Tenenbaum}]{goodman2012church}
\bibinfo{author}{Goodman, N.}, \bibinfo{author}{Mansinghka, V.},
  \bibinfo{author}{Roy, D.M.}, \bibinfo{author}{Bonawitz, K.},
  \bibinfo{author}{Tenenbaum, J.B.}, \bibinfo{year}{2012}.
\newblock \bibinfo{title}{Church: a language for generative models}.
\newblock \bibinfo{journal}{arXiv preprint arXiv:1206.3255} .
\bibitem[{Greff et~al.(2019)Greff, Kaufman, Kabra, Watters, Burgess, Zoran,
  Matthey, Botvinick and Lerchner}]{greff2019multi}
\bibinfo{author}{Greff, K.}, \bibinfo{author}{Kaufman, R.L.},
  \bibinfo{author}{Kabra, R.}, \bibinfo{author}{Watters, N.},
  \bibinfo{author}{Burgess, C.}, \bibinfo{author}{Zoran, D.},
  \bibinfo{author}{Matthey, L.}, \bibinfo{author}{Botvinick, M.},
  \bibinfo{author}{Lerchner, A.}, \bibinfo{year}{2019}.
\newblock \bibinfo{title}{Multi-object representation learning with iterative
  variational inference}, in: \bibinfo{booktitle}{International Conference on
  Machine Learning}, \bibinfo{organization}{PMLR}. pp.
  \bibinfo{pages}{2424--2433}.
\bibitem[{Gregor et~al.(2019)Gregor, Rezende, Besse, Wu, Merzic and
  Oord}]{gregor2019shaping}
\bibinfo{author}{Gregor, K.}, \bibinfo{author}{Rezende, D.J.},
  \bibinfo{author}{Besse, F.}, \bibinfo{author}{Wu, Y.},
  \bibinfo{author}{Merzic, H.}, \bibinfo{author}{Oord, A.v.d.},
  \bibinfo{year}{2019}.
\newblock \bibinfo{title}{Shaping belief states with generative environment
  models for rl}.
\newblock \bibinfo{journal}{arXiv preprint arXiv:1906.09237} .
\bibitem[{Grisetti et~al.(2007)Grisetti, Stachniss and
  Burgard}]{gridbasedfastslam2007}
\bibinfo{author}{Grisetti, G.}, \bibinfo{author}{Stachniss, C.},
  \bibinfo{author}{Burgard, W.}, \bibinfo{year}{2007}.
\newblock \bibinfo{title}{{Improved Techniques for Grid Mapping with
  Rao-Blackwellized Particle Filters}}.
\newblock \bibinfo{journal}{IEEE Transactions on Robotics}
  \bibinfo{volume}{23}, \bibinfo{pages}{34--46}.
\bibitem[{Ha and Schmidhuber(2018)}]{ha2018world}
\bibinfo{author}{Ha, D.}, \bibinfo{author}{Schmidhuber, J.},
  \bibinfo{year}{2018}.
\newblock \bibinfo{title}{World models}.
\newblock \bibinfo{journal}{arXiv preprint arXiv:1803.10122} .
\bibitem[{Haarnoja et~al.(2018a)Haarnoja, Zhou, Abbeel and
  Levine}]{Haarnoja2018}
\bibinfo{author}{Haarnoja, T.}, \bibinfo{author}{Zhou, A.},
  \bibinfo{author}{Abbeel, P.}, \bibinfo{author}{Levine, S.},
  \bibinfo{year}{2018}a.
\newblock \bibinfo{title}{Soft actor-critic: Off-policy maximum entropy deep
  reinforcement learning with a stochastic actor}.
\newblock \bibinfo{journal}{arXiv:1801.01290} .
\bibitem[{Haarnoja et~al.(2018b)Haarnoja, Zhou, Abbeel and
  Levine}]{haarnoja2018soft}
\bibinfo{author}{Haarnoja, T.}, \bibinfo{author}{Zhou, A.},
  \bibinfo{author}{Abbeel, P.}, \bibinfo{author}{Levine, S.},
  \bibinfo{year}{2018}b.
\newblock \bibinfo{title}{Soft actor-critic: Off-policy maximum entropy deep
  reinforcement learning with a stochastic actor}, in:
  \bibinfo{booktitle}{International Conference on Machine Learning},
  \bibinfo{organization}{PMLR}. pp. \bibinfo{pages}{1861--1870}.
\bibitem[{Haber et~al.(2000)Haber, Fudge and McFarland}]{Haber2000}
\bibinfo{author}{Haber, S.N.}, \bibinfo{author}{Fudge, J.L.},
  \bibinfo{author}{McFarland, N.R.}, \bibinfo{year}{2000}.
\newblock \bibinfo{title}{Striatonigrostriatal pathways in primates form an
  ascending spiral from the shell to the dorsolateral striatum}.
\newblock \bibinfo{journal}{J Neurosci} \bibinfo{volume}{20},
  \bibinfo{pages}{2369--82}.
\newblock \URLprefix \url{http://www.jneurosci.org/content/20/6/2369.full.pdf}.
\bibitem[{Hafner et~al.(2019a)Hafner, Lillicrap, Ba and
  Norouzi}]{hafner2019dream}
\bibinfo{author}{Hafner, D.}, \bibinfo{author}{Lillicrap, T.},
  \bibinfo{author}{Ba, J.}, \bibinfo{author}{Norouzi, M.},
  \bibinfo{year}{2019}a.
\newblock \bibinfo{title}{Dream to control: Learning behaviors by latent
  imagination}.
\newblock \bibinfo{journal}{arXiv preprint arXiv:1912.01603} .
\bibitem[{Hafner et~al.(2019b)Hafner, Lillicrap, Fischer, Villegas, Ha, Lee and
  Davidson}]{hafner2019planet}
\bibinfo{author}{Hafner, D.}, \bibinfo{author}{Lillicrap, T.},
  \bibinfo{author}{Fischer, I.}, \bibinfo{author}{Villegas, R.},
  \bibinfo{author}{Ha, D.}, \bibinfo{author}{Lee, H.},
  \bibinfo{author}{Davidson, J.}, \bibinfo{year}{2019}b.
\newblock \bibinfo{title}{Learning latent dynamics for planning from pixels},
  in: \bibinfo{booktitle}{International Conference on Machine Learning}, pp.
  \bibinfo{pages}{2555--2565}.
\bibitem[{Hafner et~al.(2020)Hafner, Lillicrap, Norouzi and
  Ba}]{hafner2020mastering}
\bibinfo{author}{Hafner, D.}, \bibinfo{author}{Lillicrap, T.},
  \bibinfo{author}{Norouzi, M.}, \bibinfo{author}{Ba, J.},
  \bibinfo{year}{2020}.
\newblock \bibinfo{title}{Mastering atari with discrete world models}.
\newblock \bibinfo{journal}{arXiv preprint arXiv:2010.02193} .
\bibitem[{Hafting et~al.(2005)Hafting, Fyhn, Molden, Moser and
  Moser}]{Hafting2005gridcells}
\bibinfo{author}{Hafting, T.}, \bibinfo{author}{Fyhn, M.},
  \bibinfo{author}{Molden, S.}, \bibinfo{author}{Moser, M.B.},
  \bibinfo{author}{Moser, E.I.}, \bibinfo{year}{2005}.
\newblock \bibinfo{title}{{Microstructure of a spatial map in the entorhinal
  cortex}}.
\newblock \bibinfo{journal}{Nature} \bibinfo{volume}{436},
  \bibinfo{pages}{801--806}.
\newblock \DOIprefix\doi{10.1038/nature03721}.
\bibitem[{Hagiwara et~al.(2018)Hagiwara, Inoue, Kobayashi and
  Taniguchi}]{hagiwara2018hierarchical}
\bibinfo{author}{Hagiwara, Y.}, \bibinfo{author}{Inoue, M.},
  \bibinfo{author}{Kobayashi, H.}, \bibinfo{author}{Taniguchi, T.},
  \bibinfo{year}{2018}.
\newblock \bibinfo{title}{Hierarchical spatial concept formation based on
  multimodal information for human support robots}.
\newblock \bibinfo{journal}{Frontiers in neurorobotics} \bibinfo{volume}{12},
  \bibinfo{pages}{11}.
\bibitem[{Han et~al.(2020a)Han, Doya and Tani}]{Han2020nn}
\bibinfo{author}{Han, D.}, \bibinfo{author}{Doya, K.}, \bibinfo{author}{Tani,
  J.}, \bibinfo{year}{2020}a.
\newblock \bibinfo{title}{Self-organization of action hierarchy and
  compositionality by reinforcement learning with recurrent neural networks}.
\newblock \bibinfo{journal}{Neural Networks}
  \DOIprefix\doi{10.1016/j.neunet.2020.06.002}.
\bibitem[{Han et~al.(2020b)Han, Doya and Tani}]{Han2020iclr}
\bibinfo{author}{Han, D.}, \bibinfo{author}{Doya, K.}, \bibinfo{author}{Tani,
  J.}, \bibinfo{year}{2020}b.
\newblock \bibinfo{title}{Variational recurrent models for solving partially
  observable control tasks}.
\newblock \URLprefix \url{https://iclr.cc/virtual/poster_r1lL4a4tDB.html}.
\bibitem[{Higgins et~al.(2016)Higgins, Matthey, Pal, Burgess, Glorot,
  Botvinick, Mohamed and Lerchner}]{higgins2016beta}
\bibinfo{author}{Higgins, I.}, \bibinfo{author}{Matthey, L.},
  \bibinfo{author}{Pal, A.}, \bibinfo{author}{Burgess, C.},
  \bibinfo{author}{Glorot, X.}, \bibinfo{author}{Botvinick, M.},
  \bibinfo{author}{Mohamed, S.}, \bibinfo{author}{Lerchner, A.},
  \bibinfo{year}{2016}.
\newblock \bibinfo{title}{beta-vae: Learning basic visual concepts with a
  constrained variational framework} .
\bibitem[{Hikida et~al.(2010)Hikida, Kimura, Wada, Funabiki and
  Nakanishi}]{Hikida2010}
\bibinfo{author}{Hikida, T.}, \bibinfo{author}{Kimura, K.},
  \bibinfo{author}{Wada, N.}, \bibinfo{author}{Funabiki, K.},
  \bibinfo{author}{Nakanishi, S.}, \bibinfo{year}{2010}.
\newblock \bibinfo{title}{Distinct roles of synaptic transmission in direct and
  indirect striatal pathways to reward and aversive behavior}.
\newblock \bibinfo{journal}{Neuron} \bibinfo{volume}{66},
  \bibinfo{pages}{896--907}.
\newblock \DOIprefix\doi{10.1016/j.neuron.2010.05.011}.
\bibitem[{Hohwy(2013)}]{hohwy2013predictive}
\bibinfo{author}{Hohwy, J.}, \bibinfo{year}{2013}.
\newblock \bibinfo{title}{The predictive mind}.
\newblock \bibinfo{publisher}{Oxford University Press}.
\bibitem[{Iino et~al.(2020)Iino, Sawada, Yamaguchi, Tajiri, Ishii, Kasai and
  Yagishita}]{Iino2020}
\bibinfo{author}{Iino, Y.}, \bibinfo{author}{Sawada, T.},
  \bibinfo{author}{Yamaguchi, K.}, \bibinfo{author}{Tajiri, M.},
  \bibinfo{author}{Ishii, S.}, \bibinfo{author}{Kasai, H.},
  \bibinfo{author}{Yagishita, S.}, \bibinfo{year}{2020}.
\newblock \bibinfo{title}{Dopamine d2 receptors in discrimination learning and
  spine enlargement}.
\newblock \bibinfo{journal}{Nature} \DOIprefix\doi{10.1038/s41586-020-2115-1}.
\bibitem[{Ito and Doya(2015)}]{Ito2015}
\bibinfo{author}{Ito, M.}, \bibinfo{author}{Doya, K.}, \bibinfo{year}{2015}.
\newblock \bibinfo{title}{Distinct neural representation in the dorsolateral,
  dorsomedial, and ventral parts of the striatum during fixed- and free-choice
  tasks}.
\newblock \bibinfo{journal}{Journal of Neuroscience} \bibinfo{volume}{35},
  \bibinfo{pages}{3499--3514}.
\newblock \DOIprefix\doi{10.1523/JNEUROSCI.1962-14.2015}.
\bibitem[{Kalman(1960)}]{Kalman1960}
\bibinfo{author}{Kalman, R.E.}, \bibinfo{year}{1960}.
\newblock \bibinfo{title}{A new approach to linear filtering and prediction
  problems}.
\newblock \bibinfo{journal}{Transactions of ASME} \bibinfo{volume}{82-D},
  \bibinfo{pages}{35--45}.
\bibitem[{Kameoka et~al.(2018)Kameoka, Kaneko, Tanaka and
  Hojo}]{kameoka2018stargan}
\bibinfo{author}{Kameoka, H.}, \bibinfo{author}{Kaneko, T.},
  \bibinfo{author}{Tanaka, K.}, \bibinfo{author}{Hojo, N.},
  \bibinfo{year}{2018}.
\newblock \bibinfo{title}{Stargan-vc: Non-parallel many-to-many voice
  conversion using star generative adversarial networks}, in:
  \bibinfo{booktitle}{2018 IEEE Spoken Language Technology Workshop (SLT)},
  \bibinfo{organization}{IEEE}. pp. \bibinfo{pages}{266--273}.
\bibitem[{Katsumata et~al.(2020)Katsumata, Taniguchi, Hafi, Hagiwara and
  Taniguchi}]{Katsumata2020SpCoMapGAN}
\bibinfo{author}{Katsumata, Y.}, \bibinfo{author}{Taniguchi, A.},
  \bibinfo{author}{Hafi, L.E.}, \bibinfo{author}{Hagiwara, Y.},
  \bibinfo{author}{Taniguchi, T.}, \bibinfo{year}{2020}.
\newblock \bibinfo{title}{{SpCoMapGAN : Spatial Concept Formation-based
  Semantic Mapping with Generative Adversarial Networks}}.
\newblock \bibinfo{journal}{Proceedings of the IEEE/RSJ International
  Conference on Intelligent Robots and Systems (IROS)} ,
  \bibinfo{pages}{7927--7934}.
\bibitem[{Kingma and Welling(2014)}]{kingma2013auto}
\bibinfo{author}{Kingma, D.P.}, \bibinfo{author}{Welling, M.},
  \bibinfo{year}{2014}.
\newblock \bibinfo{title}{Auto-encoding variational bayes}.
\newblock \bibinfo{journal}{International Conference on Learning
  Representations} .
\bibitem[{Knoch et~al.(2006)Knoch, Pascual-Leone, Meyer, Treyer and
  Fehr}]{knoch2006diminishing}
\bibinfo{author}{Knoch, D.}, \bibinfo{author}{Pascual-Leone, A.},
  \bibinfo{author}{Meyer, K.}, \bibinfo{author}{Treyer, V.},
  \bibinfo{author}{Fehr, E.}, \bibinfo{year}{2006}.
\newblock \bibinfo{title}{Diminishing reciprocal fairness by disrupting the
  right prefrontal cortex}.
\newblock \bibinfo{journal}{science} \bibinfo{volume}{314},
  \bibinfo{pages}{829--832}.
\bibitem[{Kosiorek et~al.(2018)Kosiorek, Kim, Posner and
  Teh}]{kosiorek2018sequential}
\bibinfo{author}{Kosiorek, A.R.}, \bibinfo{author}{Kim, H.},
  \bibinfo{author}{Posner, I.}, \bibinfo{author}{Teh, Y.W.},
  \bibinfo{year}{2018}.
\newblock \bibinfo{title}{Sequential attend, infer, repeat: Generative
  modelling of moving objects}.
\newblock \bibinfo{journal}{arXiv preprint arXiv:1806.01794} .
\bibitem[{Kostavelis and Gasteratos(2015)}]{kostavelis2015semantic}
\bibinfo{author}{Kostavelis, I.}, \bibinfo{author}{Gasteratos, A.},
  \bibinfo{year}{2015}.
\newblock \bibinfo{title}{{Semantic mapping for mobile robotics tasks: A
  survey}}.
\newblock \bibinfo{journal}{Robotics and Autonomous Systems}
  \bibinfo{volume}{66}, \bibinfo{pages}{86--103}.
\newblock \URLprefix \url{http://dx.doi.org/10.1016/j.robot.2014.12.006},
  \DOIprefix\doi{10.1016/j.robot.2014.12.006}.
\bibitem[{Kriegeskorte and Douglas(2018)}]{Kriegeskorte2018-xw}
\bibinfo{author}{Kriegeskorte, N.}, \bibinfo{author}{Douglas, P.K.},
  \bibinfo{year}{2018}.
\newblock \bibinfo{title}{Cognitive computational neuroscience}.
\newblock \bibinfo{journal}{Nat. Neurosci.} \bibinfo{volume}{21},
  \bibinfo{pages}{1148--1160}.
\bibitem[{Krizhevsky et~al.(2012)Krizhevsky, Sutskever and
  Hinton}]{Krizhevsky2012}
\bibinfo{author}{Krizhevsky, A.}, \bibinfo{author}{Sutskever, I.},
  \bibinfo{author}{Hinton, G.E.}, \bibinfo{year}{2012}.
\newblock \bibinfo{title}{{ImageNet Classification with Deep Convolutional
  Neural Networks}}.
\newblock \bibinfo{journal}{Advances In Neural Information Processing Systems}
  , \bibinfo{pages}{1--9}\href{http://arxiv.org/abs/1102.0183}{{\tt
  arXiv:1102.0183}}.
\bibitem[{Kuan et~al.(2015)Kuan, Li, Lau, Feng, Bernard, Sunkin, Zeng, Dang,
  Hawrylycz and Ng}]{Kuan2015-tl}
\bibinfo{author}{Kuan, L.}, \bibinfo{author}{Li, Y.}, \bibinfo{author}{Lau,
  C.}, \bibinfo{author}{Feng, D.}, \bibinfo{author}{Bernard, A.},
  \bibinfo{author}{Sunkin, S.M.}, \bibinfo{author}{Zeng, H.},
  \bibinfo{author}{Dang, C.}, \bibinfo{author}{Hawrylycz, M.},
  \bibinfo{author}{Ng, L.}, \bibinfo{year}{2015}.
\newblock \bibinfo{title}{Neuroinformatics of the allen mouse brain
  connectivity atlas}.
\newblock \bibinfo{journal}{Methods} \bibinfo{volume}{73},
  \bibinfo{pages}{4--17}.
\bibitem[{Laird(2012)}]{laird2012soar}
\bibinfo{author}{Laird, J.E.}, \bibinfo{year}{2012}.
\newblock \bibinfo{title}{The Soar cognitive architecture}.
\newblock \bibinfo{publisher}{MIT press}.
\bibitem[{Laird et~al.(2017)Laird, Lebiere and Rosenbloom}]{Laird2017}
\bibinfo{author}{Laird, J.E.}, \bibinfo{author}{Lebiere, C.},
  \bibinfo{author}{Rosenbloom, P.S.}, \bibinfo{year}{2017}.
\newblock \bibinfo{title}{{A standard model of the mind: Toward a common
  computational framework across artificial intelligence, cognitive science,
  neuroscience, and robotics}}.
\newblock \bibinfo{journal}{AI Magazine} \bibinfo{volume}{38},
  \bibinfo{pages}{13--26}.
\newblock \DOIprefix\doi{10.1609/aimag.v38i4.2744}.
\bibitem[{Lake et~al.(2017)Lake, Ullman, Tenenbaum and Gershman}]{Lake2017}
\bibinfo{author}{Lake, B.M.}, \bibinfo{author}{Ullman, T.D.},
  \bibinfo{author}{Tenenbaum, J.B.}, \bibinfo{author}{Gershman, S.J.},
  \bibinfo{year}{2017}.
\newblock \bibinfo{title}{Building machines that learn and think like people}.
\newblock \bibinfo{journal}{Behav Brain Sci} \bibinfo{volume}{40},
  \bibinfo{pages}{e253}.
\newblock \DOIprefix\doi{10.1017/S0140525X16001837}.
\bibitem[{Lau and Glimcher(2008)}]{Lau2008}
\bibinfo{author}{Lau, B.}, \bibinfo{author}{Glimcher, P.W.},
  \bibinfo{year}{2008}.
\newblock \bibinfo{title}{Value representations in the primate striatum during
  matching behavior}.
\newblock \bibinfo{journal}{Neuron} \bibinfo{volume}{58},
  \bibinfo{pages}{451--63}.
\newblock \DOIprefix\doi{10.1016/j.neuron.2008.02.021}.
\bibitem[{Laurent et~al.(2017)Laurent, Barnaud, Schwartz, Bessi{\`e}re and
  Diard}]{laurent2017complementary}
\bibinfo{author}{Laurent, R.}, \bibinfo{author}{Barnaud, M.L.},
  \bibinfo{author}{Schwartz, J.L.}, \bibinfo{author}{Bessi{\`e}re, P.},
  \bibinfo{author}{Diard, J.}, \bibinfo{year}{2017}.
\newblock \bibinfo{title}{The complementary roles of auditory and motor
  information evaluated in a bayesian perceptuo-motor model of speech
  perception.}
\newblock \bibinfo{journal}{Psychological review} \bibinfo{volume}{124},
  \bibinfo{pages}{572}.
\bibitem[{LeCun et~al.(2015)LeCun, Bengio and Hinton}]{lecun2015deeplearning}
\bibinfo{author}{LeCun, Y.}, \bibinfo{author}{Bengio, Y.},
  \bibinfo{author}{Hinton, G.}, \bibinfo{year}{2015}.
\newblock \bibinfo{title}{Deep learning}.
\newblock \bibinfo{journal}{Nature} \bibinfo{volume}{521},
  \bibinfo{pages}{436--444}.
\newblock \URLprefix \url{https://doi.org/10.1038/nature14539},
  \DOIprefix\doi{10.1038/nature14539}.
\bibitem[{Lee and Kawahara(2009)}]{lee2009recent}
\bibinfo{author}{Lee, A.}, \bibinfo{author}{Kawahara, T.},
  \bibinfo{year}{2009}.
\newblock \bibinfo{title}{Recent development of open-source speech recognition
  engine julius}, in: \bibinfo{booktitle}{Proceedings: APSIPA ASC 2009:
  Asia-Pacific Signal and Information Processing Association, 2009 Annual
  Summit and Conference}, \bibinfo{organization}{Asia-Pacific Signal and
  Information Processing Association, 2009 Annual~…}. pp.
  \bibinfo{pages}{131--137}.
\bibitem[{Levine(2018)}]{Levine2018}
\bibinfo{author}{Levine, S.}, \bibinfo{year}{2018}.
\newblock \bibinfo{title}{Reinforcement learning and control as probabilistic
  inference: Tutorial and review}.
\newblock \bibinfo{type}{Report}.
\bibitem[{Liang et~al.(2007)Liang, Petrov, Jordan and
  Klein}]{liang2007infinite}
\bibinfo{author}{Liang, P.}, \bibinfo{author}{Petrov, S.},
  \bibinfo{author}{Jordan, M.I.}, \bibinfo{author}{Klein, D.},
  \bibinfo{year}{2007}.
\newblock \bibinfo{title}{The infinite pcfg using hierarchical dirichlet
  processes}, in: \bibinfo{booktitle}{Proceedings of the 2007 joint conference
  on empirical methods in natural language processing and computational natural
  language learning (EMNLP-CoNLL)}, pp. \bibinfo{pages}{688--697}.
\bibitem[{Liberman et~al.(1967)Liberman, Cooper, Shankweiler and
  Studdert-Kennedy}]{liberman1967perception}
\bibinfo{author}{Liberman, A.M.}, \bibinfo{author}{Cooper, F.S.},
  \bibinfo{author}{Shankweiler, D.P.}, \bibinfo{author}{Studdert-Kennedy, M.},
  \bibinfo{year}{1967}.
\newblock \bibinfo{title}{Perception of the speech code.}
\newblock \bibinfo{journal}{Psychological review} \bibinfo{volume}{74},
  \bibinfo{pages}{431}.
\bibitem[{Luong et~al.(2015)Luong, Pham and Manning}]{Luong2015}
\bibinfo{author}{Luong, M.T.}, \bibinfo{author}{Pham, H.},
  \bibinfo{author}{Manning, C.D.}, \bibinfo{year}{2015}.
\newblock \bibinfo{title}{{Effective Approaches to Attention-based Neural
  Machine Translation}}.
\newblock \bibinfo{journal}{Emnlp} , \bibinfo{pages}{11}\URLprefix
  \url{http://arxiv.org/abs/1508.04025}, \DOIprefix\doi{10.18653/v1/D15-1166},
  \href{http://arxiv.org/abs/1508.04025}{{\tt arXiv:1508.04025}}.
\bibitem[{Markov et~al.(2013)Markov, Ercsey-Ravasz, Van~Essen, Knoblauch,
  Toroczkai and Kennedy}]{Markov2013-zd}
\bibinfo{author}{Markov, N.T.}, \bibinfo{author}{Ercsey-Ravasz, M.},
  \bibinfo{author}{Van~Essen, D.C.}, \bibinfo{author}{Knoblauch, K.},
  \bibinfo{author}{Toroczkai, Z.}, \bibinfo{author}{Kennedy, H.},
  \bibinfo{year}{2013}.
\newblock \bibinfo{title}{Cortical high-density counterstream architectures}.
\newblock \bibinfo{journal}{Science} \bibinfo{volume}{342},
  \bibinfo{pages}{1238406}.
\bibitem[{Markov et~al.(2014)Markov, Vezoli, Chameau, Falchier, Quilodran,
  Huissoud, Lamy, Misery, Giroud, Ullman, Barone, Dehay, Knoblauch and
  Kennedy}]{Markov2014-ez}
\bibinfo{author}{Markov, N.T.}, \bibinfo{author}{Vezoli, J.},
  \bibinfo{author}{Chameau, P.}, \bibinfo{author}{Falchier, A.},
  \bibinfo{author}{Quilodran, R.}, \bibinfo{author}{Huissoud, C.},
  \bibinfo{author}{Lamy, C.}, \bibinfo{author}{Misery, P.},
  \bibinfo{author}{Giroud, P.}, \bibinfo{author}{Ullman, S.},
  \bibinfo{author}{Barone, P.}, \bibinfo{author}{Dehay, C.},
  \bibinfo{author}{Knoblauch, K.}, \bibinfo{author}{Kennedy, H.},
  \bibinfo{year}{2014}.
\newblock \bibinfo{title}{Anatomy of hierarchy: feedforward and feedback
  pathways in macaque visual cortex}.
\newblock \bibinfo{journal}{J. Comp. Neurol.} \bibinfo{volume}{522},
  \bibinfo{pages}{225--259}.
\bibitem[{Millidge et~al.(2020)Millidge, Tschantz, Seth and
  Buckley}]{millidge2020relationship}
\bibinfo{author}{Millidge, B.}, \bibinfo{author}{Tschantz, A.},
  \bibinfo{author}{Seth, A.K.}, \bibinfo{author}{Buckley, C.L.},
  \bibinfo{year}{2020}.
\newblock \bibinfo{title}{On the relationship between active inference and
  control as inference}, in: \bibinfo{booktitle}{International Workshop on
  Active Inference}, \bibinfo{organization}{Springer}. pp.
  \bibinfo{pages}{3--11}.
\bibitem[{Miyazawa et~al.(2019a)Miyazawa, Aoki, Horii and
  Nagai}]{miyazawaIROS2019}
\bibinfo{author}{Miyazawa, K.}, \bibinfo{author}{Aoki, T.},
  \bibinfo{author}{Horii, T.}, \bibinfo{author}{Nagai, T.},
  \bibinfo{year}{2019}a.
\newblock \bibinfo{title}{Integration of multiple generative modules for robot
  learning}.
\newblock \bibinfo{journal}{Workshop on Deep Probabilistic Generative Models
  for Cognitive Architecture in Robotics (in IROS2019)} .
\bibitem[{Miyazawa et~al.(2020)Miyazawa, Aoki, Horii and
  Nagai}]{miyazawa2020lambert}
\bibinfo{author}{Miyazawa, K.}, \bibinfo{author}{Aoki, T.},
  \bibinfo{author}{Horii, T.}, \bibinfo{author}{Nagai, T.},
  \bibinfo{year}{2020}.
\newblock \bibinfo{title}{lambert: Language and action learning using
  multimodal bert}.
\newblock \href{http://arxiv.org/abs/2004.07093}{{\tt arXiv:2004.07093}}.
\bibitem[{Miyazawa et~al.(2019b)Miyazawa, Horii, Aoki and
  Nagai}]{miyazawaFRONT2019}
\bibinfo{author}{Miyazawa, K.}, \bibinfo{author}{Horii, T.},
  \bibinfo{author}{Aoki, T.}, \bibinfo{author}{Nagai, T.},
  \bibinfo{year}{2019}b.
\newblock \bibinfo{title}{Integrated cognitive architecture for robot learning
  of action and language}.
\newblock \bibinfo{journal}{Frontiers in Robotics and AI} \bibinfo{volume}{6},
  \bibinfo{pages}{1--20}.
\newblock \URLprefix
  \url{https://www.frontiersin.org/article/10.3389/frobt.2019.00131},
  \DOIprefix\doi{10.3389/frobt.2019.00131}.
\bibitem[{Montague et~al.(1996)Montague, Dayan and Sejnowski}]{Montague1996}
\bibinfo{author}{Montague, P.R.}, \bibinfo{author}{Dayan, P.},
  \bibinfo{author}{Sejnowski, T.J.}, \bibinfo{year}{1996}.
\newblock \bibinfo{title}{A framework for mesencephalic dopamine systems based
  on predictive hebbian learning}.
\newblock \bibinfo{journal}{Journal of Neuroscience} \bibinfo{volume}{16},
  \bibinfo{pages}{1936--1947}.
\bibitem[{Montemerlo et~al.(2002)Montemerlo, Thrun, Koller and
  Wegbreit}]{montemerlo2002fastslam}
\bibinfo{author}{Montemerlo, M.}, \bibinfo{author}{Thrun, S.},
  \bibinfo{author}{Koller, D.}, \bibinfo{author}{Wegbreit, B.},
  \bibinfo{year}{2002}.
\newblock \bibinfo{title}{{FastSLAM: A factored solution to the simultaneous
  localization and mapping problem}}, in: \bibinfo{booktitle}{In Proceedings of
  the AAAI National Conference on Artificial Intelligence},
  \bibinfo{organization}{American Association for Artificial Intelligence}. pp.
  \bibinfo{pages}{593--598}.
\bibitem[{Nakamura et~al.(2012)Nakamura, Araki, Nagai and
  Iwahashi}]{nakamura_grounding}
\bibinfo{author}{Nakamura, T.}, \bibinfo{author}{Araki, T.},
  \bibinfo{author}{Nagai, T.}, \bibinfo{author}{Iwahashi, N.},
  \bibinfo{year}{2012}.
\newblock \bibinfo{title}{Grounding of word meanings in lda-based multimodal
  concepts}.
\newblock \bibinfo{journal}{Advanced Robotics} \bibinfo{volume}{25},
  \bibinfo{pages}{2189--2206}.
\bibitem[{Nakamura et~al.(2007)Nakamura, Nagai and Iwahashi}]{nakamura2007}
\bibinfo{author}{Nakamura, T.}, \bibinfo{author}{Nagai, T.},
  \bibinfo{author}{Iwahashi, N.}, \bibinfo{year}{2007}.
\newblock \bibinfo{title}{Multimodal object categorization by a robot}, in:
  \bibinfo{booktitle}{2007 IEEE/RSJ International Conference on Intelligent
  Robots and Systems}, pp. \bibinfo{pages}{2415--2420}.
\newblock \DOIprefix\doi{10.1109/IROS.2007.4399634}.
\bibitem[{Nakamura et~al.(2011a)Nakamura, Nagai and Iwahashi}]{nakamura2011bag}
\bibinfo{author}{Nakamura, T.}, \bibinfo{author}{Nagai, T.},
  \bibinfo{author}{Iwahashi, N.}, \bibinfo{year}{2011}a.
\newblock \bibinfo{title}{Bag of multimodal lda models for concept formation},
  in: \bibinfo{booktitle}{2011 IEEE International Conference on Robotics and
  Automation}, \bibinfo{organization}{IEEE}. pp. \bibinfo{pages}{6233--6238}.
\bibitem[{Nakamura et~al.(2011b)Nakamura, Nagai and
  Iwahashi}]{nakamura2011multimodal}
\bibinfo{author}{Nakamura, T.}, \bibinfo{author}{Nagai, T.},
  \bibinfo{author}{Iwahashi, N.}, \bibinfo{year}{2011}b.
\newblock \bibinfo{title}{Multimodal categorization by hierarchical dirichlet
  process}, in: \bibinfo{booktitle}{2011 IEEE/RSJ International Conference on
  Intelligent Robots and Systems}, \bibinfo{organization}{IEEE}. pp.
  \bibinfo{pages}{1520--1525}.
\bibitem[{Nakamura et~al.(2014)Nakamura, Nagai, nad Shogo~Nagasaka, Taniguchi
  and Iwahashi}]{TomoakiNakamura2014}
\bibinfo{author}{Nakamura, T.}, \bibinfo{author}{Nagai, T.},
  \bibinfo{author}{nad Shogo~Nagasaka, K.F.}, \bibinfo{author}{Taniguchi, T.},
  \bibinfo{author}{Iwahashi, N.}, \bibinfo{year}{2014}.
\newblock \bibinfo{title}{Mutual learning of an object concept and language
  model based on mlda and npylm}, in: \bibinfo{booktitle}{2014 IEEE/RSJ
  International Conference on Intelligent Robots and Systems (IROS'14)},
  \bibinfo{address}{Chicago, IL, USA}.
\bibitem[{Nakamura et~al.(2017)Nakamura, Nagai and
  Taniguchi}]{nakamura2017serket}
\bibinfo{author}{Nakamura, T.}, \bibinfo{author}{Nagai, T.},
  \bibinfo{author}{Taniguchi, T.}, \bibinfo{year}{2017}.
\newblock \bibinfo{title}{Serket: An architecture for connecting stochastic
  models to realize a large-scale cognitive model}.
\newblock \bibinfo{journal}{arXiv preprint arXiv:1712.00929} .
\bibitem[{Negishi et~al.(2019)Negishi, Hayami, Tamura, Mizutani and
  Yamakawa}]{Negishi2019-pt}
\bibinfo{author}{Negishi, S.}, \bibinfo{author}{Hayami, T.},
  \bibinfo{author}{Tamura, H.}, \bibinfo{author}{Mizutani, H.},
  \bibinfo{author}{Yamakawa, H.}, \bibinfo{year}{2019}.
\newblock \bibinfo{title}{Neocortical functional hierarchy estimated from
  connectomic morphology in the mouse brain}, in:
  \bibinfo{booktitle}{Biologically Inspired Cognitive Architectures 2018},
  \bibinfo{publisher}{Springer International Publishing}. pp.
  \bibinfo{pages}{234--238}.
\bibitem[{van Niekerk et~al.(2020)van Niekerk, Nortje and
  Kamper}]{van2020vector}
\bibinfo{author}{van Niekerk, B.}, \bibinfo{author}{Nortje, L.},
  \bibinfo{author}{Kamper, H.}, \bibinfo{year}{2020}.
\newblock \bibinfo{title}{Vector-quantized neural networks for acoustic unit
  discovery in the zerospeech 2020 challenge}.
\newblock \bibinfo{journal}{arXiv preprint arXiv:2005.09409} .
\bibitem[{{Nishihara} et~al.(2017){Nishihara}, {Nakamura} and
  {Nagai}}]{nishihara2017}
\bibinfo{author}{{Nishihara}, J.}, \bibinfo{author}{{Nakamura}, T.},
  \bibinfo{author}{{Nagai}, T.}, \bibinfo{year}{2017}.
\newblock \bibinfo{title}{Online algorithm for robots to learn object concepts
  and language model}.
\newblock \bibinfo{journal}{IEEE Transactions on Cognitive and Developmental
  Systems} \bibinfo{volume}{9}, \bibinfo{pages}{255--268}.
\newblock \DOIprefix\doi{10.1109/TCDS.2016.2552579}.
\bibitem[{Okada et~al.(2020)Okada, Kosaka and Taniguchi}]{planet_b}
\bibinfo{author}{Okada, M.}, \bibinfo{author}{Kosaka, N.},
  \bibinfo{author}{Taniguchi, T.}, \bibinfo{year}{2020}.
\newblock \bibinfo{title}{Planet of the bayesians: Reconsidering and improving
  deep planning network by incorporating bayesian inference}, in:
  \bibinfo{booktitle}{{IEEE/RSJ} International Conference on Intelligent Robots
  and Systems, {IROS} 2020, Las Vegas, NV, USA, October 24, 2020 - January 24,
  2021}, \bibinfo{publisher}{{IEEE}}. pp. \bibinfo{pages}{5611--5618}.
\newblock \URLprefix \url{https://doi.org/10.1109/IROS45743.2020.9340873},
  \DOIprefix\doi{10.1109/IROS45743.2020.9340873}.
\bibitem[{Okada and Taniguchi(2020a)}]{okada2020dreaming}
\bibinfo{author}{Okada, M.}, \bibinfo{author}{Taniguchi, T.},
  \bibinfo{year}{2020}a.
\newblock \bibinfo{title}{Dreaming: Model-based reinforcement learning by
  latent imagination without reconstruction}.
\newblock \bibinfo{journal}{arXiv preprint arXiv:2007.14535} .
\bibitem[{Okada and Taniguchi(2020b)}]{okada2020variational}
\bibinfo{author}{Okada, M.}, \bibinfo{author}{Taniguchi, T.},
  \bibinfo{year}{2020}b.
\newblock \bibinfo{title}{Variational inference mpc for bayesian model-based
  reinforcement learning}, in: \bibinfo{booktitle}{Conference on Robot
  Learning}, \bibinfo{organization}{PMLR}. pp. \bibinfo{pages}{258--272}.
\bibitem[{O'keefe and Nadel(1978)}]{Okeefe1978placecells}
\bibinfo{author}{O'keefe, J.}, \bibinfo{author}{Nadel, L.},
  \bibinfo{year}{1978}.
\newblock \bibinfo{title}{{The Hippocampus as a Cognitive Map}}.
\newblock \bibinfo{publisher}{Cambridge University Press}.
\newblock \DOIprefix\doi{10.5840/philstudies19802725}.
\bibitem[{Oudeyer et~al.(2007)Oudeyer, Kaplan and Hafner}]{Najeeb1988}
\bibinfo{author}{Oudeyer, P.Y.}, \bibinfo{author}{Kaplan, F.},
  \bibinfo{author}{Hafner, V.V.}, \bibinfo{year}{2007}.
\newblock \bibinfo{title}{{Intrinsic Motivation Systems for Autonomous Mental
  Development}}.
\newblock \bibinfo{journal}{IEEE TRANSACTIONS ON EVOLUTIONARY COMPUTATION,}
  \bibinfo{volume}{1}, \bibinfo{pages}{265--286}.
\newblock \DOIprefix\doi{10.1007/BF01252847}.
\bibitem[{Pan et~al.(2008)Pan, Sawa, Tsuda, Tsukada and
  Sakagami}]{pan2008reward}
\bibinfo{author}{Pan, X.}, \bibinfo{author}{Sawa, K.}, \bibinfo{author}{Tsuda,
  I.}, \bibinfo{author}{Tsukada, M.}, \bibinfo{author}{Sakagami, M.},
  \bibinfo{year}{2008}.
\newblock \bibinfo{title}{Reward prediction based on stimulus categorization in
  primate lateral prefrontal cortex}.
\newblock \bibinfo{journal}{Nature neuroscience} \bibinfo{volume}{11},
  \bibinfo{pages}{703--712}.
\bibitem[{Pandya and Seltzer(1982)}]{pandya1982intrinsic}
\bibinfo{author}{Pandya, D.N.}, \bibinfo{author}{Seltzer, B.},
  \bibinfo{year}{1982}.
\newblock \bibinfo{title}{Intrinsic connections and architectonics of posterior
  parietal cortex in the rhesus monkey}.
\newblock \bibinfo{journal}{Journal of Comparative Neurology}
  \bibinfo{volume}{204}, \bibinfo{pages}{196--210}.
\bibitem[{Parisi et~al.(2019)Parisi, Kemker, Part, Kanan and
  Wermter}]{parisi2019continual}
\bibinfo{author}{Parisi, G.I.}, \bibinfo{author}{Kemker, R.},
  \bibinfo{author}{Part, J.L.}, \bibinfo{author}{Kanan, C.},
  \bibinfo{author}{Wermter, S.}, \bibinfo{year}{2019}.
\newblock \bibinfo{title}{Continual lifelong learning with neural networks: A
  review}.
\newblock \bibinfo{journal}{Neural Networks} \bibinfo{volume}{113},
  \bibinfo{pages}{54--71}.
\bibitem[{Pasquereau et~al.(2007)Pasquereau, Nadjar, Arkadir, Bezard,
  Goillandeau, Bioulac, Gross and Boraud}]{Pasquereau2007}
\bibinfo{author}{Pasquereau, B.}, \bibinfo{author}{Nadjar, A.},
  \bibinfo{author}{Arkadir, D.}, \bibinfo{author}{Bezard, E.},
  \bibinfo{author}{Goillandeau, M.}, \bibinfo{author}{Bioulac, B.},
  \bibinfo{author}{Gross, C.E.}, \bibinfo{author}{Boraud, T.},
  \bibinfo{year}{2007}.
\newblock \bibinfo{title}{Shaping of motor responses by incentive values
  through the basal ganglia}.
\newblock \bibinfo{journal}{J Neurosci} \bibinfo{volume}{27},
  \bibinfo{pages}{1176--83}.
\newblock \URLprefix \url{http://www.ncbi.nlm.nih.gov/pubmed/17267573},
  \DOIprefix\doi{10.1523/JNEUROSCI.3745-06.2007}.
\bibitem[{Poldrack and Gorgolewski(2017)}]{Poldrack2017-ni}
\bibinfo{author}{Poldrack, R.A.}, \bibinfo{author}{Gorgolewski, K.J.},
  \bibinfo{year}{2017}.
\newblock \bibinfo{title}{{OpenfMRI}: Open sharing of task {fMRI} data}.
\newblock \bibinfo{journal}{Neuroimage} \bibinfo{volume}{144},
  \bibinfo{pages}{259--261}.
\bibitem[{Pradeep et~al.(2013)Pradeep, Knight and Gurumoorthy}]{Pradeep2013-gb}
\bibinfo{author}{Pradeep, A.}, \bibinfo{author}{Knight, R.T.},
  \bibinfo{author}{Gurumoorthy, R.}, \bibinfo{year}{2013}.
\newblock \bibinfo{title}{Neuro-informatics repository system}.
\bibitem[{Rabiner and Juang(1986)}]{rabiner1986introduction}
\bibinfo{author}{Rabiner, L.}, \bibinfo{author}{Juang, B.},
  \bibinfo{year}{1986}.
\newblock \bibinfo{title}{An introduction to hidden markov models}.
\newblock \bibinfo{journal}{ieee assp magazine} \bibinfo{volume}{3},
  \bibinfo{pages}{4--16}.
\bibitem[{Rand et~al.(2012)Rand, Greene and Nowak}]{rand2012spontaneous}
\bibinfo{author}{Rand, D.G.}, \bibinfo{author}{Greene, J.D.},
  \bibinfo{author}{Nowak, M.A.}, \bibinfo{year}{2012}.
\newblock \bibinfo{title}{Spontaneous giving and calculated greed}.
\newblock \bibinfo{journal}{Nature} \bibinfo{volume}{489},
  \bibinfo{pages}{427--430}.
\bibitem[{Rao and Ballard(1999)}]{Rao1999-ym}
\bibinfo{author}{Rao, R.P.}, \bibinfo{author}{Ballard, D.H.},
  \bibinfo{year}{1999}.
\newblock \bibinfo{title}{Predictive coding in the visual cortex: a functional
  interpretation of some extra-classical receptive-field effects}.
\newblock \bibinfo{journal}{Nat. Neurosci.} \bibinfo{volume}{2},
  \bibinfo{pages}{79--87}.
\bibitem[{Reynolds et~al.(2001)Reynolds, Hyland and Wickens}]{Reynolds2001}
\bibinfo{author}{Reynolds, J.N.}, \bibinfo{author}{Hyland, B.I.},
  \bibinfo{author}{Wickens, J.R.}, \bibinfo{year}{2001}.
\newblock \bibinfo{title}{A cellular mechanism of reward-related learning}.
\newblock \bibinfo{journal}{Nature} \bibinfo{volume}{413},
  \bibinfo{pages}{67--70}.
\newblock \DOIprefix\doi{10.1038/35092560}.
\bibitem[{Rosenbloom(2011)}]{Rosenbloom2011}
\bibinfo{author}{Rosenbloom, P.S.}, \bibinfo{year}{2011}.
\newblock \bibinfo{title}{{Rethinking cognitive architecture via graphical
  models}}.
\newblock \bibinfo{journal}{Cognitive Systems Research} \bibinfo{volume}{12},
  \bibinfo{pages}{198--209}.
\newblock \URLprefix \url{http://dx.doi.org/10.1016/j.cogsys.2010.07.006},
  \DOIprefix\doi{10.1016/j.cogsys.2010.07.006}.
\bibitem[{Rosenbloom et~al.(2016)Rosenbloom, Demski and
  Ustun}]{rosenbloom2016sigma}
\bibinfo{author}{Rosenbloom, P.S.}, \bibinfo{author}{Demski, A.},
  \bibinfo{author}{Ustun, V.}, \bibinfo{year}{2016}.
\newblock \bibinfo{title}{The sigma cognitive architecture and system: Towards
  functionally elegant grand unification}.
\newblock \bibinfo{journal}{Journal of Artificial General Intelligence}
  \bibinfo{volume}{7}, \bibinfo{pages}{1--103}.
\bibitem[{Sagar et~al.(2016)Sagar, Seymour and Henderson}]{sagar2016}
\bibinfo{author}{Sagar, M.}, \bibinfo{author}{Seymour, M.},
  \bibinfo{author}{Henderson, A.}, \bibinfo{year}{2016}.
\newblock \bibinfo{title}{Creating connection with autonomous facial
  animation}.
\newblock \bibinfo{journal}{Communications of the ACM} \bibinfo{volume}{59},
  \bibinfo{pages}{82--91}.
\newblock \DOIprefix\doi{10.1145/2950041}.
\bibitem[{Sakagami and Pan(2007)}]{sakagami2007functional}
\bibinfo{author}{Sakagami, M.}, \bibinfo{author}{Pan, X.},
  \bibinfo{year}{2007}.
\newblock \bibinfo{title}{Functional role of the ventrolateral prefrontal
  cortex in decision making}.
\newblock \bibinfo{journal}{Current opinion in neurobiology}
  \bibinfo{volume}{17}, \bibinfo{pages}{228--233}.
\bibitem[{Samejima et~al.(2005)Samejima, Ueda, Doya and Kimura}]{Samejima2005}
\bibinfo{author}{Samejima, K.}, \bibinfo{author}{Ueda, Y.},
  \bibinfo{author}{Doya, K.}, \bibinfo{author}{Kimura, M.},
  \bibinfo{year}{2005}.
\newblock \bibinfo{title}{Representation of action-specific reward values in
  the striatum}.
\newblock \bibinfo{journal}{Science} \bibinfo{volume}{310},
  \bibinfo{pages}{1337--40}.
\newblock \DOIprefix\doi{10.1126/science.1115270}.
\bibitem[{Samsonovich et~al.(2016)Samsonovich, Klimov and
  Rybina}]{Samsonovich2016-tn}
\bibinfo{editor}{Samsonovich, A.V.}, \bibinfo{editor}{Klimov, V.V.},
  \bibinfo{editor}{Rybina, G.V.} (Eds.), \bibinfo{year}{2016}.
\newblock \bibinfo{title}{Biologically Inspired Cognitive Architectures
  ({BICA}) for Young Scientists : Proceedings of the First International Early
  Research Career Enhancement School ({FIERCES} 2016)}.
\newblock \bibinfo{publisher}{Springer, Cham}.
\bibitem[{Sasaki et~al.(2020)Sasaki, Yamakawa and Arakawa}]{Sasaki2020-zp}
\bibinfo{author}{Sasaki, M.}, \bibinfo{author}{Yamakawa, H.},
  \bibinfo{author}{Arakawa, N.}, \bibinfo{year}{2020}.
\newblock \bibinfo{title}{Construction of a whole brain reference architecture
  ({WBRA})}, in: \bibinfo{booktitle}{International Symposium on Artificial
  Intelligence and Brain Science}, p.~\bibinfo{pages}{31}.
\bibitem[{Sato and Kameya(1997)}]{sato1997prism}
\bibinfo{author}{Sato, T.}, \bibinfo{author}{Kameya, Y.}, \bibinfo{year}{1997}.
\newblock \bibinfo{title}{Prism: a language for symbolic-statistical modeling},
  in: \bibinfo{booktitle}{IJCAI}, pp. \bibinfo{pages}{1330--1339}.
\bibitem[{Savelli et~al.(2008)Savelli, Yoganarasimha and
  Knierim}]{Savelli2008-dl}
\bibinfo{author}{Savelli, F.}, \bibinfo{author}{Yoganarasimha, D.},
  \bibinfo{author}{Knierim, J.J.}, \bibinfo{year}{2008}.
\newblock \bibinfo{title}{Influence of boundary removal on the spatial
  representations of the medial entorhinal cortex}.
\newblock \bibinfo{journal}{Hippocampus} \bibinfo{volume}{18},
  \bibinfo{pages}{1270--1282}.
\bibitem[{Schmidhuber(1990)}]{schmidhuber1990making}
\bibinfo{author}{Schmidhuber, J.}, \bibinfo{year}{1990}.
\newblock \bibinfo{title}{Making the world differentiable: On using
  self-supervised fully recurrent n eu al networks for dynamic reinforcement
  learning and planning in non-stationary environm nts} .
\bibitem[{Schultz et~al.(1997)Schultz, Dayan and Montague}]{Schultz1997}
\bibinfo{author}{Schultz, W.}, \bibinfo{author}{Dayan, P.},
  \bibinfo{author}{Montague, P.R.}, \bibinfo{year}{1997}.
\newblock \bibinfo{title}{A neural substrate of prediction and reward}.
\newblock \bibinfo{journal}{Science} \bibinfo{volume}{275},
  \bibinfo{pages}{1593--1599}.
\newblock \DOIprefix\doi{10.1126/science.275.5306.1593}.
\bibitem[{Seitz et~al.(2009)Seitz, Kim and Watanabe}]{Seitz2009}
\bibinfo{author}{Seitz, A.R.}, \bibinfo{author}{Kim, D.},
  \bibinfo{author}{Watanabe, T.}, \bibinfo{year}{2009}.
\newblock \bibinfo{title}{Rewards evoke learning of unconsciously processed
  visual stimuli in adult humans}.
\newblock \bibinfo{journal}{Neuron} \bibinfo{volume}{61},
  \bibinfo{pages}{700--7}.
\newblock \DOIprefix\doi{10.1016/j.neuron.2009.01.016}.
\bibitem[{Solstad et~al.(2008)Solstad, Boccara, Kropff, Moser and
  Moser}]{Solstad2008-ko}
\bibinfo{author}{Solstad, T.}, \bibinfo{author}{Boccara, C.N.},
  \bibinfo{author}{Kropff, E.}, \bibinfo{author}{Moser, M.B.},
  \bibinfo{author}{Moser, E.I.}, \bibinfo{year}{2008}.
\newblock \bibinfo{title}{Representation of geometric borders in the entorhinal
  cortex}.
\bibitem[{Srinivas et~al.(2020)Srinivas, Laskin and Abbeel}]{srinivas2020curl}
\bibinfo{author}{Srinivas, A.}, \bibinfo{author}{Laskin, M.},
  \bibinfo{author}{Abbeel, P.}, \bibinfo{year}{2020}.
\newblock \bibinfo{title}{Curl: Contrastive unsupervised representations for
  reinforcement learning}.
\newblock \bibinfo{journal}{arXiv preprint arXiv:2004.04136} .
\bibitem[{Sutton and Barto(2018)}]{sutton2018reinforcement}
\bibinfo{author}{Sutton, R.S.}, \bibinfo{author}{Barto, A.G.},
  \bibinfo{year}{2018}.
\newblock \bibinfo{title}{Reinforcement learning: An introduction}.
\newblock \bibinfo{publisher}{MIT press}.
\bibitem[{Suzuki et~al.(2021)Suzuki, Kaneko and Matsuo}]{suzuki2021pixyz}
\bibinfo{author}{Suzuki, M.}, \bibinfo{author}{Kaneko, T.},
  \bibinfo{author}{Matsuo, Y.}, \bibinfo{year}{2021}.
\newblock \bibinfo{title}{Pixyz: a library for developing deep generative
  models}.
\newblock \href{http://arxiv.org/abs/2107.13109}{{\tt arXiv:2107.13109}}.
\bibitem[{Suzuki et~al.(2016)Suzuki, Nakayama and Matsuo}]{suzuki2016joint}
\bibinfo{author}{Suzuki, M.}, \bibinfo{author}{Nakayama, K.},
  \bibinfo{author}{Matsuo, Y.}, \bibinfo{year}{2016}.
\newblock \bibinfo{title}{Joint multimodal learning with deep generative
  models}.
\newblock \bibinfo{journal}{arXiv preprint arXiv:1611.01891} .
\bibitem[{Takahashi et~al.(2015)Takahashi, Itaya, Nakamura, Koizumi, Arakawa,
  Tomita and Yamakawa}]{Takahashi2015-fd}
\bibinfo{author}{Takahashi, K.}, \bibinfo{author}{Itaya, K.},
  \bibinfo{author}{Nakamura, M.}, \bibinfo{author}{Koizumi, M.},
  \bibinfo{author}{Arakawa, N.}, \bibinfo{author}{Tomita, M.},
  \bibinfo{author}{Yamakawa, H.}, \bibinfo{year}{2015}.
\newblock \bibinfo{title}{A generic software platform for brain-inspired
  cognitive computing}.
\newblock \bibinfo{journal}{Procedia Comput. Sci.} \bibinfo{volume}{71},
  \bibinfo{pages}{31--37}.
\bibitem[{Tanaka et~al.(2015)Tanaka, Pan, Oguchi, Taylor and
  Sakagami}]{tanaka2015dissociable}
\bibinfo{author}{Tanaka, S.}, \bibinfo{author}{Pan, X.},
  \bibinfo{author}{Oguchi, M.}, \bibinfo{author}{Taylor, J.E.},
  \bibinfo{author}{Sakagami, M.}, \bibinfo{year}{2015}.
\newblock \bibinfo{title}{Dissociable functions of reward inference in the
  lateral prefrontal cortex and the striatum}.
\newblock \bibinfo{journal}{Frontiers in psychology} \bibinfo{volume}{6},
  \bibinfo{pages}{995}.
\bibitem[{Tanaka et~al.(2018)Tanaka, Tanaka, Kondo, Terada, Kawaguchi and
  Matsuzaki}]{Tanaka2018}
\bibinfo{author}{Tanaka, Y.H.}, \bibinfo{author}{Tanaka, Y.R.},
  \bibinfo{author}{Kondo, M.}, \bibinfo{author}{Terada, S.I.},
  \bibinfo{author}{Kawaguchi, Y.}, \bibinfo{author}{Matsuzaki, M.},
  \bibinfo{year}{2018}.
\newblock \bibinfo{title}{Thalamocortical axonal activity in motor cortex
  exhibits layer-specific dynamics during motor learning}.
\newblock \bibinfo{journal}{Neuron} \bibinfo{volume}{100},
  \bibinfo{pages}{244--258 e12}.
\newblock \DOIprefix\doi{10.1016/j.neuron.2018.08.016}.
\bibitem[{Tanevska et~al.(2019)Tanevska, Rea, Sandini, Ca{\~n}amero and
  Sciutti}]{tanevska2019cognitive}
\bibinfo{author}{Tanevska, A.}, \bibinfo{author}{Rea, F.},
  \bibinfo{author}{Sandini, G.}, \bibinfo{author}{Ca{\~n}amero, L.},
  \bibinfo{author}{Sciutti, A.}, \bibinfo{year}{2019}.
\newblock \bibinfo{title}{A cognitive architecture for socially adaptable
  robots}, in: \bibinfo{booktitle}{2019 Joint IEEE 9th International Conference
  on Development and Learning and Epigenetic Robotics (ICDL-EpiRob)},
  \bibinfo{organization}{IEEE}. pp. \bibinfo{pages}{195--200}.
\bibitem[{Tangiuchi et~al.(2019)Tangiuchi, Mochihashi, Nagai, Uchida, Inoue,
  Kobayashi, Nakamura, Hagiwara, Iwahashi and Inamura}]{Taniguchi2019langrobo}
\bibinfo{author}{Tangiuchi, T.}, \bibinfo{author}{Mochihashi, D.},
  \bibinfo{author}{Nagai, T.}, \bibinfo{author}{Uchida, S.},
  \bibinfo{author}{Inoue, N.}, \bibinfo{author}{Kobayashi, I.},
  \bibinfo{author}{Nakamura, T.}, \bibinfo{author}{Hagiwara, Y.},
  \bibinfo{author}{Iwahashi, N.}, \bibinfo{author}{Inamura, T.},
  \bibinfo{year}{2019}.
\newblock \bibinfo{title}{Survey on frontiers of language and robotics}.
\newblock \bibinfo{journal}{Advanced Robotics} \bibinfo{volume}{33},
  \bibinfo{pages}{700--730}.
\newblock \URLprefix \url{https://doi.org/10.1080/01691864.2019.1632223},
  \DOIprefix\doi{10.1080/01691864.2019.1632223},
  \href{http://arxiv.org/abs/https://doi.org/10.1080/01691864.2019.1632223}{{\tt
  arXiv:https://doi.org/10.1080/01691864.2019.1632223}}.
\bibitem[{Taniguchi et~al.(2021a)Taniguchi, Fukawa and
  Yamakawa}]{Taniguchi2021hpf-pgm}
\bibinfo{author}{Taniguchi, A.}, \bibinfo{author}{Fukawa, A.},
  \bibinfo{author}{Yamakawa, H.}, \bibinfo{year}{2021}a.
\newblock \bibinfo{title}{{Hippocampal formation-inspired probabilistic
  generative model}}.
\newblock \bibinfo{journal}{arXiv} .
\bibitem[{Taniguchi et~al.(2017)Taniguchi, Hagiwara, Taniguchi and
  Inamura}]{taniguchi2017online}
\bibinfo{author}{Taniguchi, A.}, \bibinfo{author}{Hagiwara, Y.},
  \bibinfo{author}{Taniguchi, T.}, \bibinfo{author}{Inamura, T.},
  \bibinfo{year}{2017}.
\newblock \bibinfo{title}{Online spatial concept and lexical acquisition with
  simultaneous localization and mapping}, in: \bibinfo{booktitle}{2017 IEEE/RSJ
  International Conference on Intelligent Robots and Systems (IROS)},
  \bibinfo{organization}{IEEE}. pp. \bibinfo{pages}{811--818}.
\bibitem[{Taniguchi et~al.(2020a)Taniguchi, Hagiwara, Taniguchi and
  Inamura}]{taniguchi2020improved}
\bibinfo{author}{Taniguchi, A.}, \bibinfo{author}{Hagiwara, Y.},
  \bibinfo{author}{Taniguchi, T.}, \bibinfo{author}{Inamura, T.},
  \bibinfo{year}{2020}a.
\newblock \bibinfo{title}{Improved and scalable online learning of spatial
  concepts and language models with mapping}.
\newblock \bibinfo{journal}{Autonomous Robots} , \bibinfo{pages}{1--20}.
\bibitem[{Taniguchi et~al.(2020b)Taniguchi, Hagiwara, Taniguchi and
  Inamura}]{ataniguchi2020spconavi}
\bibinfo{author}{Taniguchi, A.}, \bibinfo{author}{Hagiwara, Y.},
  \bibinfo{author}{Taniguchi, T.}, \bibinfo{author}{Inamura, T.},
  \bibinfo{year}{2020}b.
\newblock \bibinfo{title}{{Spatial Concept-Based Navigation with Human Speech
  Instructions via Probabilistic Inference on Bayesian Generative Model}}.
\newblock \bibinfo{journal}{Advanced Robotics} \bibinfo{volume}{34},
  \bibinfo{pages}{1213--1228}.
\newblock \URLprefix
  \url{https://www.tandfonline.com/doi/full/10.1080/01691864.2020.1817777},
  \DOIprefix\doi{10.1080/01691864.2020.1817777}.
\bibitem[{Taniguchi et~al.(2021b)Taniguchi, Isobe, Hafi, Hagiwara and
  Taniguchi}]{ataniguchi2020TidyUpHere}
\bibinfo{author}{Taniguchi, A.}, \bibinfo{author}{Isobe, S.},
  \bibinfo{author}{Hafi, L.E.L.}, \bibinfo{author}{Hagiwara, Y.},
  \bibinfo{author}{Taniguchi, T.}, \bibinfo{year}{2021}b.
\newblock \bibinfo{title}{{Autonomous Planning Based on Spatial Concepts to
  Tidy Up Home Environments with Service Robots}}.
\newblock \bibinfo{journal}{Advanced Robotics}
  \DOIprefix\doi{10.1080/01691864.2021.1890212}.
\bibitem[{Taniguchi et~al.(2016a)Taniguchi, Taniguchi and
  Inamura}]{taniguchi2016spatial}
\bibinfo{author}{Taniguchi, A.}, \bibinfo{author}{Taniguchi, T.},
  \bibinfo{author}{Inamura, T.}, \bibinfo{year}{2016}a.
\newblock \bibinfo{title}{Spatial concept acquisition for a mobile robot that
  integrates self-localization and unsupervised word discovery from spoken
  sentences}.
\newblock \bibinfo{journal}{IEEE Transactions on Cognitive and Developmental
  Systems} \bibinfo{volume}{8}, \bibinfo{pages}{285--297}.
\bibitem[{Taniguchi et~al.(2018a)Taniguchi, Taniguchi and
  Inamura}]{taniguchi2018unsupervised}
\bibinfo{author}{Taniguchi, A.}, \bibinfo{author}{Taniguchi, T.},
  \bibinfo{author}{Inamura, T.}, \bibinfo{year}{2018}a.
\newblock \bibinfo{title}{Unsupervised spatial lexical acquisition by updating
  a language model with place clues}.
\newblock \bibinfo{journal}{Robotics and Autonomous Systems}
  \bibinfo{volume}{99}, \bibinfo{pages}{166--180}.
\bibitem[{Taniguchi et~al.(2016b)Taniguchi, Nagai, Nakamura, Iwahashi, Ogata
  and Asoh}]{Taniguchi2016symbol}
\bibinfo{author}{Taniguchi, T.}, \bibinfo{author}{Nagai, T.},
  \bibinfo{author}{Nakamura, T.}, \bibinfo{author}{Iwahashi, N.},
  \bibinfo{author}{Ogata, T.}, \bibinfo{author}{Asoh, H.},
  \bibinfo{year}{2016}b.
\newblock \bibinfo{title}{Symbol emergence in robotics: a survey}.
\newblock \bibinfo{journal}{Advanced Robotics} \bibinfo{volume}{30},
  \bibinfo{pages}{706--728}.
\bibitem[{Taniguchi et~al.(2016c)Taniguchi, Nagasaka and
  Nakashima}]{taniguchi2016nonparametric}
\bibinfo{author}{Taniguchi, T.}, \bibinfo{author}{Nagasaka, S.},
  \bibinfo{author}{Nakashima, R.}, \bibinfo{year}{2016}c.
\newblock \bibinfo{title}{Nonparametric bayesian double articulation analyzer
  for direct language acquisition from continuous speech signals}.
\newblock \bibinfo{journal}{IEEE Transactions on Cognitive and Developmental
  Systems} \bibinfo{volume}{8}, \bibinfo{pages}{171--185}.
\bibitem[{Taniguchi et~al.(2020c)Taniguchi, Nakamura, Suzuki, Kuniyasu,
  Hayashi, Taniguchi, Horii and Nagai}]{taniguchi2020neuro}
\bibinfo{author}{Taniguchi, T.}, \bibinfo{author}{Nakamura, T.},
  \bibinfo{author}{Suzuki, M.}, \bibinfo{author}{Kuniyasu, R.},
  \bibinfo{author}{Hayashi, K.}, \bibinfo{author}{Taniguchi, A.},
  \bibinfo{author}{Horii, T.}, \bibinfo{author}{Nagai, T.},
  \bibinfo{year}{2020}c.
\newblock \bibinfo{title}{Neuro-serket: development of integrative cognitive
  system through the composition of deep probabilistic generative models}.
\newblock \bibinfo{journal}{New Generation Computing} , \bibinfo{pages}{1--26}.
\bibitem[{Taniguchi et~al.(2018b)Taniguchi, Ugur, Hoffmann, Jamone, Nagai,
  Rosman, Matsuka, Iwahashi, Oztop, Piater et~al.}]{Taniguchi2018symbol}
\bibinfo{author}{Taniguchi, T.}, \bibinfo{author}{Ugur, E.},
  \bibinfo{author}{Hoffmann, M.}, \bibinfo{author}{Jamone, L.},
  \bibinfo{author}{Nagai, T.}, \bibinfo{author}{Rosman, B.},
  \bibinfo{author}{Matsuka, T.}, \bibinfo{author}{Iwahashi, N.},
  \bibinfo{author}{Oztop, E.}, \bibinfo{author}{Piater, J.}, et~al.,
  \bibinfo{year}{2018}b.
\newblock \bibinfo{title}{Symbol emergence in cognitive developmental systems:
  a survey}.
\newblock \bibinfo{journal}{IEEE Transactions on Cognitive and Developmental
  Systems} .
\bibitem[{Taube et~al.(1990a)Taube, Muller and Ranck}]{Taube1990-vm}
\bibinfo{author}{Taube, J.S.}, \bibinfo{author}{Muller, R.U.},
  \bibinfo{author}{Ranck, Jr, J.B.}, \bibinfo{year}{1990}a.
\newblock \bibinfo{title}{Head-direction cells recorded from the postsubiculum
  in freely moving rats. i. description and quantitative analysis}.
\newblock \bibinfo{journal}{J. Neurosci.} \bibinfo{volume}{10},
  \bibinfo{pages}{420--435}.
\bibitem[{Taube et~al.(1990b)Taube, Muller and Ranck}]{Taube1990-oi}
\bibinfo{author}{Taube, J.S.}, \bibinfo{author}{Muller, R.U.},
  \bibinfo{author}{Ranck, Jr, J.B.}, \bibinfo{year}{1990}b.
\newblock \bibinfo{title}{Head-direction cells recorded from the postsubiculum
  in freely moving rats. {II}. effects of environmental manipulations}.
\newblock \bibinfo{journal}{J. Neurosci.} \bibinfo{volume}{10},
  \bibinfo{pages}{436--447}.
\bibitem[{Thrun et~al.(2005)Thrun, Burgard and Fox}]{thrun2005probabilistic}
\bibinfo{author}{Thrun, S.}, \bibinfo{author}{Burgard, W.},
  \bibinfo{author}{Fox, D.}, \bibinfo{year}{2005}.
\newblock \bibinfo{title}{{Probabilistic Robotics}}.
\newblock \bibinfo{publisher}{MIT Press}.
\bibitem[{Todorov(2008)}]{Todorov2008}
\bibinfo{author}{Todorov, E.}, \bibinfo{year}{2008}.
\newblock \bibinfo{title}{General duality between optimal control and
  estimation}, in: \bibinfo{booktitle}{The 47th IEEE Conference on Decision and
  Control}, pp. \bibinfo{pages}{4286 -- 4292}.
\bibitem[{Tolman(1948)}]{Tolman1948}
\bibinfo{author}{Tolman, E.C.}, \bibinfo{year}{1948}.
\newblock \bibinfo{title}{{Cognitive maps in rats and men}}.
\newblock \bibinfo{journal}{Psychological Review} \bibinfo{volume}{55},
  \bibinfo{pages}{189--208}.
\newblock \DOIprefix\doi{10.1037/h0061626}.
\bibitem[{Tran et~al.(2017)Tran, Hoffman, Saurous, Brevdo, Murphy and
  Blei}]{tran2017deep}
\bibinfo{author}{Tran, D.}, \bibinfo{author}{Hoffman, M.D.},
  \bibinfo{author}{Saurous, R.A.}, \bibinfo{author}{Brevdo, E.},
  \bibinfo{author}{Murphy, K.}, \bibinfo{author}{Blei, D.M.},
  \bibinfo{year}{2017}.
\newblock \bibinfo{title}{Deep probabilistic programming}.
\newblock \bibinfo{journal}{arXiv preprint arXiv:1701.03757} .
\bibitem[{Uchiyama et~al.(2017)Uchiyama, Ikeda and
  Taketomi}]{ref:taketomi2017visual}
\bibinfo{author}{Uchiyama, H.}, \bibinfo{author}{Ikeda, S.},
  \bibinfo{author}{Taketomi, T.}, \bibinfo{year}{2017}.
\newblock \bibinfo{title}{Visual slam algorithms: a survey from 2010 to 2016}.
\newblock \bibinfo{journal}{IPSJ Transactions on Computer Vision and
  Applications} \bibinfo{volume}{9}, \bibinfo{pages}{16}.
\bibitem[{Ungerleider et~al.(1989)Ungerleider, Gaffan and
  Pelak}]{ungerleider1989projections}
\bibinfo{author}{Ungerleider, L.}, \bibinfo{author}{Gaffan, D.},
  \bibinfo{author}{Pelak, V.}, \bibinfo{year}{1989}.
\newblock \bibinfo{title}{Projections from inferior temporal cortex to
  prefrontal cortex via the uncinate fascicle in rhesus monkeys}.
\newblock \bibinfo{journal}{Experimental brain research} \bibinfo{volume}{76},
  \bibinfo{pages}{473--484}.
\bibitem[{Ungerleider(1982)}]{ungerleider1982two}
\bibinfo{author}{Ungerleider, L.G.}, \bibinfo{year}{1982}.
\newblock \bibinfo{title}{Two cortical visual systems}.
\newblock \bibinfo{journal}{Analysis of visual behavior} ,
  \bibinfo{pages}{549--586}.
\bibitem[{Vaswani et~al.(2017)Vaswani, Shazeer, Parmar, Uszkoreit, Jones,
  Gomez, Kaiser and Polosukhin}]{vaswani2017attention}
\bibinfo{author}{Vaswani, A.}, \bibinfo{author}{Shazeer, N.},
  \bibinfo{author}{Parmar, N.}, \bibinfo{author}{Uszkoreit, J.},
  \bibinfo{author}{Jones, L.}, \bibinfo{author}{Gomez, A.N.},
  \bibinfo{author}{Kaiser, {\L}.}, \bibinfo{author}{Polosukhin, I.},
  \bibinfo{year}{2017}.
\newblock \bibinfo{title}{Attention is all you need}, in:
  \bibinfo{booktitle}{Advances in neural information processing systems}, pp.
  \bibinfo{pages}{5998--6008}.
\bibitem[{Vernon et~al.(2007)Vernon, Metta and Sandini}]{Vernon2007}
\bibinfo{author}{Vernon, D.}, \bibinfo{author}{Metta, G.},
  \bibinfo{author}{Sandini, G.}, \bibinfo{year}{2007}.
\newblock \bibinfo{title}{{The iCub cognitive architecture: Interactive
  development in a humanoid robot}}.
\newblock \bibinfo{journal}{2007 IEEE 6th International Conference on
  Development and Learning, ICDL} ,
  \bibinfo{pages}{122--127}\DOIprefix\doi{10.1109/DEVLRN.2007.4354038}.
\bibitem[{Vernon et~al.(2016)Vernon, {Von Hofsten} and Fadiga}]{Vernon2016}
\bibinfo{author}{Vernon, D.}, \bibinfo{author}{{Von Hofsten}, C.},
  \bibinfo{author}{Fadiga, L.}, \bibinfo{year}{2016}.
\newblock \bibinfo{title}{{Desiderata for developmental cognitive
  architectures}}.
\newblock \bibinfo{journal}{Biologically Inspired Cognitive Architectures}
  \bibinfo{volume}{18}, \bibinfo{pages}{116--127}.
\newblock \URLprefix \url{http://dx.doi.org/10.1016/j.bica.2016.10.004},
  \DOIprefix\doi{10.1016/j.bica.2016.10.004}.
\bibitem[{Von~Helmholtz(1867)}]{von1867treatise}
\bibinfo{author}{Von~Helmholtz, H.}, \bibinfo{year}{1867}.
\newblock \bibinfo{title}{Treatise on physiological optics vol. iii} .
\bibitem[{Voorn et~al.(2004)Voorn, Vanderschuren, Groenewegen, Robbins and
  Pennartz}]{Voorn2004}
\bibinfo{author}{Voorn, P.}, \bibinfo{author}{Vanderschuren, L.J.},
  \bibinfo{author}{Groenewegen, H.J.}, \bibinfo{author}{Robbins, T.W.},
  \bibinfo{author}{Pennartz, C.M.}, \bibinfo{year}{2004}.
\newblock \bibinfo{title}{Putting a spin on the dorsal-ventral divide of the
  striatum}.
\newblock \bibinfo{journal}{Trends Neurosci} \bibinfo{volume}{27},
  \bibinfo{pages}{468--74}.
\newblock \DOIprefix\doi{10.1016/j.tins.2004.06.006}.
\bibitem[{Wang et~al.(2020)Wang, Elfwing and Uchibe}]{Wang2020}
\bibinfo{author}{Wang, J.}, \bibinfo{author}{Elfwing, S.},
  \bibinfo{author}{Uchibe, E.}, \bibinfo{year}{2020}.
\newblock \bibinfo{title}{Modular deep reinforcement learning from reward and
  punishment for robot navigation}.
\newblock \bibinfo{journal}{Neural Networks}
  \DOIprefix\doi{10.1016/j.neunet.2020.12.001}.
\bibitem[{Whitlock et~al.(2008)Whitlock, Sutherland, Witter, Moser and
  Moser}]{Whitlock2008-zq}
\bibinfo{author}{Whitlock, J.R.}, \bibinfo{author}{Sutherland, R.J.},
  \bibinfo{author}{Witter, M.P.}, \bibinfo{author}{Moser, M.B.},
  \bibinfo{author}{Moser, E.I.}, \bibinfo{year}{2008}.
\newblock \bibinfo{title}{Navigating from hippocampus to parietal cortex}.
\newblock \bibinfo{journal}{Proc. Natl. Acad. Sci. U. S. A.}
  \bibinfo{volume}{105}, \bibinfo{pages}{14755--14762}.
\bibitem[{Wilber et~al.(2014)Wilber, Clark, Demecha, Mesina, Vos and
  McNaughton}]{Wilber2014-ob}
\bibinfo{author}{Wilber, A.A.}, \bibinfo{author}{Clark, B.J.},
  \bibinfo{author}{Demecha, A.J.}, \bibinfo{author}{Mesina, L.},
  \bibinfo{author}{Vos, J.M.}, \bibinfo{author}{McNaughton, B.L.},
  \bibinfo{year}{2014}.
\newblock \bibinfo{title}{Cortical connectivity maps reveal anatomically
  distinct areas in the parietal cortex of the rat}.
\newblock \bibinfo{journal}{Front. Neural Circuits} \bibinfo{volume}{8},
  \bibinfo{pages}{146}.
\bibitem[{Wilber et~al.(2017)Wilber, Skelin, Wu and McNaughton}]{Wilber2017-dp}
\bibinfo{author}{Wilber, A.A.}, \bibinfo{author}{Skelin, I.},
  \bibinfo{author}{Wu, W.}, \bibinfo{author}{McNaughton, B.L.},
  \bibinfo{year}{2017}.
\newblock \bibinfo{title}{Laminar organization of encoding and memory
  reactivation in the parietal cortex}.
\newblock \bibinfo{journal}{Neuron} \bibinfo{volume}{95},
  \bibinfo{pages}{1406--1419.e5}.
\bibitem[{Wolpert et~al.(2003)Wolpert, Doya and Kawato}]{wolpert2003unifying}
\bibinfo{author}{Wolpert, D.M.}, \bibinfo{author}{Doya, K.},
  \bibinfo{author}{Kawato, M.}, \bibinfo{year}{2003}.
\newblock \bibinfo{title}{A unifying computational framework for motor control
  and social interaction}.
\newblock \bibinfo{journal}{Philosophical Transactions of the Royal Society of
  London. Series B: Biological Sciences} \bibinfo{volume}{358},
  \bibinfo{pages}{593--602}.
\bibitem[{Wolpert et~al.(1998)Wolpert, Miall and Kawato}]{Wolpert1998}
\bibinfo{author}{Wolpert, D.M.}, \bibinfo{author}{Miall, R.C.},
  \bibinfo{author}{Kawato, M.}, \bibinfo{year}{1998}.
\newblock \bibinfo{title}{Internal models in the cerebellum}.
\newblock \bibinfo{journal}{Trends in Cognitive Sciences} \bibinfo{volume}{2},
  \bibinfo{pages}{338--347}.
\bibitem[{Yagishita et~al.(2014)Yagishita, Hayashi-Takagi, Ellis-Davies,
  Urakubo, Ishii and Kasai}]{Yagishita2014}
\bibinfo{author}{Yagishita, S.}, \bibinfo{author}{Hayashi-Takagi, A.},
  \bibinfo{author}{Ellis-Davies, G.C.}, \bibinfo{author}{Urakubo, H.},
  \bibinfo{author}{Ishii, S.}, \bibinfo{author}{Kasai, H.},
  \bibinfo{year}{2014}.
\newblock \bibinfo{title}{A critical time window for dopamine actions on the
  structural plasticity of dendritic spines}.
\newblock \bibinfo{journal}{Science} \bibinfo{volume}{345},
  \bibinfo{pages}{1616--20}.
\newblock \DOIprefix\doi{10.1126/science.1255514}.
\bibitem[{Yamagishi et~al.(2017)Yamagishi, Matsumoto, Kiyonari, Takagishi, Li,
  Kanai and Sakagami}]{yamagishi2017response}
\bibinfo{author}{Yamagishi, T.}, \bibinfo{author}{Matsumoto, Y.},
  \bibinfo{author}{Kiyonari, T.}, \bibinfo{author}{Takagishi, H.},
  \bibinfo{author}{Li, Y.}, \bibinfo{author}{Kanai, R.},
  \bibinfo{author}{Sakagami, M.}, \bibinfo{year}{2017}.
\newblock \bibinfo{title}{Response time in economic games reflects different
  types of decision conflict for prosocial and proself individuals}.
\newblock \bibinfo{journal}{Proceedings of the National Academy of Sciences}
  \bibinfo{volume}{114}, \bibinfo{pages}{6394--6399}.
\bibitem[{Yamakawa(2020)}]{Yamakawa2020-xa}
\bibinfo{author}{Yamakawa, H.}, \bibinfo{year}{2020}.
\newblock \bibinfo{title}{Revealing the computational meaning of neocortical
  interarea signals}.
\newblock \bibinfo{journal}{Front. Comput. Neurosci.} \bibinfo{volume}{14},
  \bibinfo{pages}{74}.
\bibitem[{Yamakawa(2021)}]{Yamakawa2021-qa}
\bibinfo{author}{Yamakawa, H.}, \bibinfo{year}{2021}.
\newblock \bibinfo{title}{The whole brain architecture approach: Accelerating
  the development of artificial general intelligence by referring to the
  brain}.
\newblock \bibinfo{journal}{Neural Netw.} .
\bibitem[{Yoshizawa et~al.(2018)Yoshizawa, Ito and Doya}]{Yoshizawa2018}
\bibinfo{author}{Yoshizawa, T.}, \bibinfo{author}{Ito, M.},
  \bibinfo{author}{Doya, K.}, \bibinfo{year}{2018}.
\newblock \bibinfo{title}{Reward-predictive neural activities in striatal
  striosome compartments}.
\newblock \bibinfo{journal}{eNeuro} \bibinfo{volume}{5},
  \bibinfo{pages}{e0367--17.2018}.
\newblock \DOIprefix\doi{10.1523/ENEURO.0367-17.2018}.
\bibitem[{{Zhang} et~al.(2017){Zhang}, {Chan} and {Jaitly}}]{verydeep}
\bibinfo{author}{{Zhang}, Y.}, \bibinfo{author}{{Chan}, W.},
  \bibinfo{author}{{Jaitly}, N.}, \bibinfo{year}{2017}.
\newblock \bibinfo{title}{Very deep convolutional networks for end-to-end
  speech recognition}, in: \bibinfo{booktitle}{2017 IEEE International
  Conference on Acoustics, Speech and Signal Processing (ICASSP)}, pp.
  \bibinfo{pages}{4845--4849}.
\newblock \DOIprefix\doi{10.1109/ICASSP.2017.7953077}.

\end{thebibliography}







\end{document}